%% file: neurips_2025.tex
\documentclass{article}

     \usepackage[preprint,nonatbib]{neurips_2025}

\usepackage[utf8]{inputenc} 
\usepackage[T1]{fontenc}    
\usepackage{hyperref}       
\usepackage{url}            
\usepackage{booktabs}       
\usepackage{amsfonts}       
\usepackage{nicefrac}       
\usepackage{microtype}      
\usepackage{xcolor}         
\usepackage{graphicx}
\usepackage{makecell}
\usepackage{tcolorbox}
\usepackage{enumitem}
\usepackage{multirow}
\usepackage{amsmath}
\usepackage{todonotes}
\usepackage{subcaption}
\usepackage[numbers]{natbib}


\title{Learning-Time Encoding Shapes Unlearning in LLMs}

\author{%
  Ruihan Wu\thanks{Equal contribution} \\
  UC, San Diego\\
  \texttt{ruw076@ucsd.edu} \\
  \And
  Konstantin Garov$^{*}$ \\
  UC, San Diego\\
  \texttt{kgarov@ucsd.edu} \\
  \And
  Kamalika Chaudhuri \\
  UC, San Diego\\
  \texttt{kamalika@ucsd.edu} \\
}

\begin{document}

\maketitle

\begin{abstract}
As large language models (LLMs) are increasingly deployed in the real world, the ability to ``unlearn'', or remove specific pieces of knowledge post hoc, has become essential for a variety of reasons ranging from privacy regulations to correcting outdated or harmful content. Prior work has proposed unlearning benchmarks and algorithms, and has typically assumed that the training process and the target model are fixed. In this work, we empirically investigate how learning-time choices in knowledge encoding impact the effectiveness of unlearning factual knowledge. Our experiments reveal two key findings: (1) learning with paraphrased descriptions improves unlearning performance and (2) unlearning individual piece of knowledge from a chunk of text is challenging. Our results suggest that learning-time knowledge encoding may play a central role in enabling reliable post-hoc unlearning. \footnote{The code is publicly released at \url{https://github.com/wrh14/learning_time_shapes_unlearning}.}

%

\end{abstract}

\input{sections/intro}
\input{sections/settings_new}
\input{sections/dataset}
\input{sections/mem_vs_gen}
\input{sections/group_ft}
\input{sections/related_work}
\input{sections/conclusion}

\newpage
\bibliographystyle{unsrt}
\bibliography{references}

\newpage
\appendix
\input{sections/appendix}


\end{document}

%% file: sections/intro.tex
\section{Introduction}

Large Language Models (LLMs) acquire vast amounts of factual knowledge through large-scale pretraining as well as subsequent fine-tuning. 
As they are increasingly deployed in real applications, there is an increasing need for ``unlearning'' certain information in an efficient post-hoc way~\citep{bourtoule2021machine, liu2025rethinking} from pre-trained or the fine-tuned models.
This need arises for several reasons. 
One is compliance with privacy regulations such as the GDPR's "Right to be Forgotten"~\citep{gdpr2016} -- for example, when a user requests that personal data used during training be removed. 
Other motivations include addressing copyright violations~\citep{eldan2023s, dou2024avoiding, vyas2023provable}, removing unsafe or harmful content (such as instructions for building weapons)~\citep{yao2024large, li2024wmdp}, and removing personal and sensitive information~\citep{jang2022knowledge, wu2023depn, barrett2023identifying}. These diverse motivations often align with slightly different objectives for the unlearning process.

One common goal of unlearning in LLMs is to make specific factual knowledge non-extractable, which means that prevent the model from generating it in response to relevant prompts \citep{jang2022knowledge, si2023knowledge, guo2024mechanistic, tian2024forget, choi2024snap, yuan2025towards, wu2024evaluating, patilcan}, and at the same time retain the remaining knowledge.
Prior work has primarily focused on benchmarks~\citep{mainitofu, shi2024muse, yao2024machine, jin2024rwku} and developing algorithms~\citep{ilharco2022editing, si2023knowledge,zhangnegative, yu2023unlearning, wu2023depn, jia2025wagle, eldan2023s, patilcan}, and typically assume that both the trained model and the unlearning targets are fixed. 
The central goal in these studies is to improve the effectiveness of the unlearning method itself.
However, a crucial factor is often overlooked: the way a model is trained -- including how knowledge is encoded in the training data -- may significantly influence how challenging it is to later unlearn that knowledge.
This raises a fundamental question:
$$\textbf{Does learning-time knowledge encoding affect knowledge unlearning?}$$
By varying  how knowledge is encoded as textual data while keeping the set of factual knowledge constant, we aim to understand how learning-time encoding shapes unlearning.

To ensure fair comparison, we investigate this question through controlled experiments. For this purpose, we extend two existing unlearning datasets -- Eval-DU~\citep{wu2024evaluating} and TOFU~\cite{mainitofu} -- resulting in \emph{Eval-DU+} and \emph{TOFU+}. Both datasets involve synthetic biographies of ``made-up'' characters that are unlikely to occur in the pre-training corpus; 
this allows us to control the exact textual encodings observed by the LLM during training.
We fine-tune LLMs on identical sets of factual knowledge, varying only the knowledge textual encoding. After fine-tuning, we attempt to unlearn specific pieces of knowledge and analyze the differences in the unlearning across different types of encoding.
Notably, our study focuses on unlearning from fine-tuned models, a common scenario where sensitive content or private user data could be introduced\footnote{We also include experiments with causal language modeling, same as the pre-training objective, and multiple LLM architectures, which may offer indirect evidence toward generalization to pretrained models. However, due to the lack of visibility into pretraining data of the existing publicly pre-trained models and limited computational resources for pretraining from scratch on a sufficiently large controlled corpus, we leave formal validation of this generalization to future work.}.


Using the constructed testbed, we first empirically study \textbf{the effects of paraphrased texts on knowledge unlearning.}
We compare two fine-tuning setups: one in which each knowledge piece is encoded by a single description, and another in which each piece is associated with multiple paraphrased descriptions. 
We observe that learning with multiple paraphrased descriptions improves {\em{unlearning}} effectiveness. It helps remove memorization of the original training texts and reduces the model’s ability to extract the target knowledge when prompted with unseen paraphrased inputs.

Second, we aim to \textbf{empirically understand the behavior of unlearning knowledge embedded within chunks of text}.
The finetuning set consists of chunks of text, where each chunk summarizes multiple pieces of knowledge.
We observe that unlearning individual knowledge pieces becomes significantly more challenging when the target knowledge are entangled with retained content within the same chunk.
Motivated by this, we further formulate and empirically validate two hypotheses:
(1) unlearning is more effective when the unlearning split aligns with the chunk boundaries in the training data; and
(2) the isolation of forget from retain knowledge within the same chunk of text makes unlearning easier.


Our empirical results suggest two practical strategies: \textbf{paraphrasing}, that is using multiple paraphrased descriptions of knowledge during fine-tuning,  and \textbf{separating}, 
that is structuring the training data to avoid text entanglement along potential unlearn and retain splits in the future unlearning.
Both of them can improve the post-hoc efficiency of unlearning for large language models.


%% file: sections/settings_new.tex
\section{Problem Set-up}
\label{sec:problem_setup}

\subsection{Problem formulation}
\label{sec:problem_form}

A single piece of knowledge can be encoded in training data in different ways.
Prior work~\citep{allen2024physics, allenphysics} suggests that learning with different encodings influences both the model’s memorization of training instances and its ability to extract the underlying knowledge when prompted with alternative phrasings.
Meanwhile, unlearning factual knowledge aims to remove both the memorized content and the model’s ability to extract knowledge from unseen prompts. This raises a natural question: Does the difficulty of unlearning a piece of knowledge $k$ vary depending on how $k$ was encoded during training?
In this paper, we investigate two concrete problem settings to answer this question.

\textbf{Problem I: The effect of text paraphrasing on unlearning.} 
Prior studies have shown that paraphrased representations can lead LLMs to internalize knowledge more robustly~\citep{allenphysics}, improving generalization to unseen prompts.
This raises the question:
$$
\textbf{Do paraphrased encodings of knowledge during training help post-hoc unlearning?}
$$

%
%

\begin{figure*}
\includegraphics[width=\textwidth]{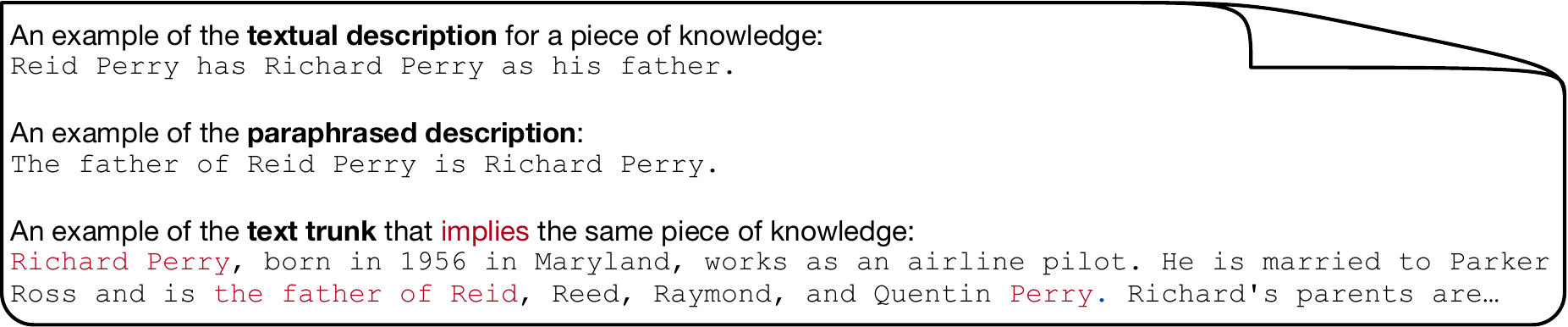}	
\vspace{-3ex}
\caption{Examples of different textual descriptions for the same piece of knowledge in Eval-DU+.}
\vspace{-2ex}
\label{fig:k_example}
\end{figure*}

We consider two fine-tuning datasets, \texttt{FT-Single} and \texttt{FT-Mul} both encoding the same knowledge set.
In \texttt{FT-Single}, each knowledge piece is represented by a single description and in \texttt{FT-Mul} by multiple paraphrased descriptions; see examples in Figure~\ref{fig:k_example}).
We fine-tune LLMs on each dataset and compare unlearning performance for different unlearning methods.

\textbf{Problem II: the unlearning from text chunks.} 
In natural datasets, knowledge is often embedded in larger text alongside multiple other knowledge pieces, such as the paragraphs from Wikipedia. Unlearn requests may apply only to a subset of the knowledge within a chunk, while the rest of the content is to be preserved.
For instance, in a biography of a public figure, personal details may need to be unlearned, while professional accomplishments should remain intact.
This raises the question:
$$
\textbf{How effective is unlearning a subset of the knowledge within text chunks?}
$$

To explore this, we construct a fine-tuning dataset \texttt{FT-Mul-Chunk}, where each knowledge piece is implicitly embedded across multiple paraphrased text chunks.
Each chunk may also include other knowledge pieces that will not be targeted for removal; see examples in Figure~\ref{fig:k_example}.

\paragraph{Rationale for focusing on fine-tuning.}
Our study focuses on unlearning from fine-tuned models, an important use-case in which sensitive or private user data is often introduced during customization for downstream tasks.
It also allows precise control over the knowledge space.
While the study targets fine-tuning, we include causal language modeling (the same training objective as pre-training) and multiple LLM architectures, which may offer indirect evidence toward generalization to pretrained models.
A formal investigation of the novel unlearning problem proposed in this work within the pre-training setting remains an important direction. 
However, we leave this to future work due to limited transparency in data of the existing pre-trained models and the high computational cost of pretraining a model from scratch on a sufficiently large and controlled corpus.



\subsection{Unlearning methods} 
\label{sec:unlearning_setup}
In this subsection, we introduce the unlearning methods we evaluate in our empirical analysis. Suppose an LLM is already trained on a represention of a knowledge base $K$. The objective of \textit{factual knowledge unlearning} is that a subset $K_{\text{ul}}$ of $K$ becomes no longer extractable from the LLM while preserving the model's utility. We consider three choices for the textual encoding $D_{\text{ul}}$ of $K_{\text{ul}}$ and two unlearning algorithms.

\paragraph{Knowledge textual encodings for unlearning.} 
A common approach for defining the unlearning dataset $D_{\rm ul}$ is to identify the exact data points used during fine-tuning to represent $K_{\rm ul}$. This aligns with GDPR’s original motivation of removing the influence of specific records. However, 
this is not always applicable in factual knowledge unlearning: 
rather than the exact samples from the training data, the unlearning requests are formulated only based on the target knowledge.
First, identifying the samples among the fine-tuning texts representing $K_{\rm ul}$ may be infeasible. More importantly, there may be no single data point that encodes only the target knowledge, making it difficult to remove it without affecting other knowledge.
Alternatively,  $D_{\rm ul}$ can be constructed by generating textual representations for the target knowledge at unlearning time. We consider the following three options: 


\begin{enumerate}[leftmargin=*, nosep]
	\item \texttt{UL-Exact}~\citep{mainitofu, eldan2023s, shi2024muse}: $D_{\text{ul}}$ consists of the exact texts used to represent $k$ during fine-tuning. For models fine-tuned on \texttt{FT-Single} or \texttt{FT-Mul}, we directly reuse the descriptions in the fine-tuning dataset. For models trained on \texttt{FT-Mul-Chunk}, we pick text chunks from the fine-tuning set that implicitly encode $k$, though these chunks may also include other non-targeted knowledge.
	\item \texttt{UL-Single}~\citep{patilcan}: For every target knowledge piece $k$, $D_{\text{ul}}$ includes one textual description of $k$ that differs from the description used in fine-tuning.
	\item \texttt{UL-Mul}~\citep{patilcan}: For every target knowledge piece $k$, $D_{\text{ul}}$ includes multiple paraphrased descriptions of $k$ not used in fine-tuning, offering diverse yet unseen ways of expressing the same knowledge.
\end{enumerate}

\paragraph{Unlearning algorithms.} We experiment with two representative unlearning algorithms that are also evaluated in previous benchmarks~\citep{mainitofu, shi2024muse, wu2024evaluating}: \textbf{gradient ascent (GA)}~\citep{jang2022knowledge} and \textbf{task vector (TV)}~\citep{ilharco2022editing, zhang2023composing}. 
\textbf{GA} removes knowledge by ascending the loss on the unlearning dataset \( D_{\rm ul} \), updating parameters $\theta$ in the LLM $\pi_{\theta}$ over \( T \) steps as $\theta_{t+1} := \theta_t + \eta_t \cdot \nabla_{\theta} \mathbb{E}_{D_{\mathrm{ul}}}[\ell(\pi_{\theta_t}, x)].$
The trade-off between unlearning and utility preservation is controlled by the number of steps \( t \): more steps generally yield stronger unlearning but risk greater utility loss. \textbf{TV} computes a parameter difference vector between the original model \( \theta_{\mathrm{original}} \) and a model \( \theta_{\mathrm{overfit}} \) trained to overfit \( D_{\mathrm{ul}} \). The final model is then defined as $\theta_{\mathrm{unlearn}} = \theta_{\mathrm{original}} - \alpha(\theta_{\mathrm{overfit}} - \theta_{\mathrm{original}}),$ where the scaling factor \( \alpha \) controls the strength of unlearning.
We also discuss other existing unlearning algorithms in the related work.

\subsection{Unlearning evalutions}
\textbf{Two types unlearning-retain trade-off.} Similar to existing unlearning benchmarks~\citep{mainitofu, shi2024muse, wu2024evaluating}, we evaluate unlearning effectiveness through the trade-off between forgetting the target knowledge and retaining the non-target (retain) knowledge. Let $e(\mathrm{LLM}, x_{k})\in [0, 1]$ be a knowledge score measuring the degree to which the model retains knowledge $k$ based on a description $x_{k}$. Given a target and a retain knowledge sets $K_{\rm ul}$ and $K_{\rm rt}$, we define the average knowledge scores:

$$
\text{Unlearn Score: }\frac{1}{|K_{\rm ul}|}\sum_{k\in K_{\rm ul}}e(\mathrm{LLM}, x_{k}),\ \ \ \text{Retain Score: }\frac{1}{|K_{\rm rt}|}\sum_{k\in K_{\rm rt}}e(\mathrm{LLM}, x_{k})
$$
The goal of unlearning is to minimize the unlearning score (i.e., forget target knowledge) while maximizing the retain score (i.e., preserve non-target knowledge). In particular, we consider two evaluation modes based on how $x_k$ is defined:
\begin{enumerate}[leftmargin=*, nosep]
\item \textit{Memorization trade-off}: $x_{k}$ is the text description of $k$ appearing in fine-tuning dataset. This evaluates the model’s memorization of the texts used during the model's training.
\item \textit{Extraction trade-off}: $x_{k}$ is  a paraphrased description of $k$ not used during fine-tuning or unlearning. This evaluates the model’s ability to extract knowledge beyond its memorized description.
\end{enumerate}

\begin{figure*}
\centering
\includegraphics[width=0.9\textwidth]{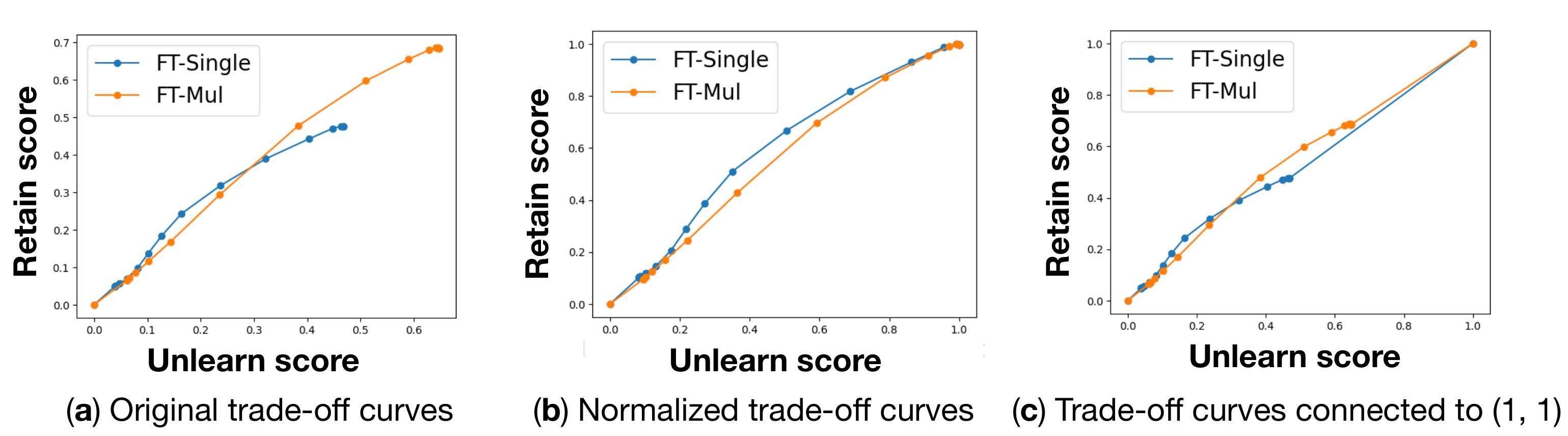}	
\vspace{-2ex}
\caption{Illustrations for Norm-AUC and AUC.
\textbf{(a)} shows the vanilla extraction trade-off curves comparison when the models are fine-tuned by \texttt{FT-Single} and \texttt{FT-Mul} and the unlearning methods are fixed; \textbf{(b)} and \textbf{(c)} show the curves for calculating Norm-AUC and AUC given the curves in (a).}
\vspace{-2ex}
\label{fig:curve_example}
\end{figure*}

\paragraph{Quantitative metrics for evaluating the trade-off: Norm-AUC and AUC.}
To evaluate the unlearn-retain trade-off for an unlearning method, we vary the parameter controlling the trade-off (e.g. $t$ in GA and $\alpha$ in TV) across a list of pre-defined values. For each parameter value we obtain a model checkpoint, whose unlearn and retain scores we compute. These scores are plotted to form a trade-off curve (Figure~\ref{fig:curve_example}), where curves closer to the top-left indicate a more favorable trade-off.


When comparing different fine-tuning strategies under a fixed unlearning configuration (i.e., using the same unlearning data and algorithm), the trade-off curves may start at different points due to the different fine-tuned models. For instance, models fine-tuned with \texttt{FT-Mul} typically achieve higher initial knowledge scores. To account for this we define the \textbf{Norm-AUC ($\uparrow$)}. This metric first normalizes all knowledge scores by their value in the original fine-tuned model and then computes the area under the normalized curve (Figure~\ref{fig:curve_example}, middle). A higher Norm-AUC indicates a more efficient unlearning and a Norm-AUC of 0.5 implies that unlearn and retain scores are decreasing at the same rate. In addition, we also report the absolute \textbf{AUC ($\uparrow$)}. For fairness before computing AUC, we align all curves to start from the same reference point \((1, 1)\)  (Figure~\ref{fig:curve_example}, right). Together, the two metrics provide complementary insights: Norm-AUC highlights the relative efficiency of unlearning, while AUC captures the absolute level of retained knowledge at different unlearning stages.



%

%% file: sections/dataset.tex
\section{More Experimental Set-ups}
\label{sec:exp_setup}
In addition to the fine-tuning, unlearning, and evaluation setups introduced in the previous section, we now describe the dataset and model configurations used in our experiments. Table~\ref{tab:sum_setup} provides a summary of all experimental settings.

\begin{table*}[!t]
\caption{Summary of experimental setups used in this paper. Our study focuses on the effects of fine-tuning on unlearning; the remaining configurations define the framework for evaluation.}
\label{tab:sum_setup}
\centering
\resizebox{\textwidth}{!}{%
\begin{tabular}{c|cc|cc|c|c}
\toprule
	\multirow{2}{*}{Fine-tuning} & \multicolumn{2}{c|}{Unlearning} & \multicolumn{2}{c|}{Unlearn-Retain Evaluation} & \multirow{2}{*}{Dataset} & \multirow{2}{*}{Model} \\
	& Unlearning Data & Unlearning Algo. & Type of Trade-off & Quantative Metric & &  \\
	\midrule
	\makecell{\texttt{FT-Single} \\ \texttt{FT-Mul} \\ \texttt{FT-Mul-Chunk}} & 
	\makecell{\texttt{UL-Exact} \\ \texttt{UL-Single} \\  \texttt{UL-Mul}} & 
	\makecell{Gradieng Ascent \\ Task Vector} & 
	\makecell{Memorization \\ Extraction} & 
	\makecell{Norm-AUC \\ AUC} & 
	\makecell{Eval-DU+ \\ TOFU+} &
	\makecell{Llama2-7B \\ Llama3-8B \\ Gemma2-2B}
	\\
	\bottomrule
\end{tabular}
}
\end{table*}
\subsection{Dataset preparation -- Eval-DU+ and TOFU+}
\label{sec:dataset}
\paragraph{Dataset augmentations.} 
In order to systematically study how learning-time knowledge encodings affect unlearning, we construct two datasets designed to support controlled experiments. Specifically, we augment two existing unlearning datasets — Eval-DU~\citep{wu2024evaluating} and TOFU~\citep{mainitofu} — to form \textbf{Eval-DU+} and \textbf{TOFU+}.
The original datasets offer two properties that allow us to construct the augmented datasets: structured knowledge spaces and initial textual descriptions for each piece of knowledge.

We begin by defining the atomic knowledge pieces and their partitioning, which are later used to generate text chunks in both datasets.
\textbf{In Eval-DU},  each knowledge piece is a factual triple (subject, relation, object), such as family relationships or biographical details like birth year, birthplace, and occupation. The dataset contains 862 such facts involving 100 fictitious individuals. We group the knowledge pieces by subject to form 100 sets of facts.
\textbf{TOFU}  is a question-answering (QA) dataset about fictitious authors. It includes 200 authors, each associated with 20 QA pairs. We treat each QA pair as representing one atomic fact and partition the knowledge by author, yielding 200 partitions.

We extend both datasets by generating the additional data required for our experimental setup
\footnote{All generations are performed using ChatGPT-4o~\citep{achiam2023gpt}. See Appendix for generation prompts and examples.}:
(1) multiple paraphrased descriptions for each individual knowledge piece, and
(2) multiple paraphrased text chunks for each partition of the knowledge set.
Figure~\ref{fig:k_example} shows the data examples in Eval-DU+.
In TOFU+, each text chunk is a synthesized QA pair that consolidates the content of all 20 original QA pairs in a partition (i.e., all facts for a given author).
Examples of these QA pairs are provided in the appendix.

Notably, in Eval-DU+, both the knowledge descriptions and the corresponding text chunks are presented in a narrative format, while in TOFU+, both follow a question–answer (QA) format. In addition, the two datasets span distinct knowledge domains: Eval-DU+ focuses on relational and biographical facts, whereas TOFU+ centers on fictional author profiles.
\textbf{By constructing Eval-DU+ and TOFU+ and conducting experiments across these two domains and representational formats, we establish a robust testbed for analyzing how learning-time knowledge encodings influence the unlearning.}

\paragraph{Unlearn–retain split in Eval-DU+ and TOFU+.}
In Eval-DU+, we randomly select 100 out of 862 knowledge pieces as the unlearn split, with the remaining pieces forming the retain split.
In TOFU+, we adopt the original unlearn–retain split: 40 knowledge pieces associated with 2 out of the 200 authors form the unlearn split, while the knowledge associated with the remaining authors constitutes the retain set. 
\textbf{Importantly, there are structural differences in the distribution of unlearning targets.}
In Eval-DU+ the text chunk of an individual is likely to contain both unlearn and retain knowledge. In contrast, in TOFU+ all unlearn knowledge is concentrated on two authors, meaning that text chunks are either fully targeted for unlearning or fully in the retain split. 
\textbf{This leads to key empirical differences discussed in later sections.}


\paragraph{Knowledge score function $e$.}
We use a knowledge score \( e(\mathrm{LLM}, x_k) \) to measure how well an LLM retains a knowledge piece \( k \) when presented with its textual representation \( x_k \). 
This score forms the basis of our unlearn–retain trade-off evaluations, as defined in Section~\ref{sec:problem_setup}.

In TOFU+, where \( x_k \) is a QA pair, we adopt the ``Probability'' metric from the original TOFU benchmark: given a question embedded in a prompt template, the score is the likelihood the LLM assigns to generating the reference answer.
In Eval-DU+, where \( x_k \) is a sentence encoding a knowledge triple \( (s, r, o) \), we identify the words or phrases corresponding to the subject, relation, and object. We then compute the conditional probability of the final token (e.g., the object) given the preceding tokens in the sentence. This score reflects how well the model can extract a missing element of the triple when the other two are provided in context.
For simplicity, we refer to both of these scoring methods as \textit{probability} throughout the paper.

\subsection{Model set-ups}
We now describe the model setup. Implementation details including hyperparameters for fine-tuning and unlearning specific to each model are provided in the appendix.

\paragraph{Datasets and models.}
Our experiments involve three large language models: Llama2-7B~\citep{touvron2023llama}, Llama3-8B~\citep{grattafiori2024llama}, and Gemma2-2B~\citep{team2024gemma}. We evaluate four combinations of models and datasets: (Llama2-7B, Eval-DU+), (Llama3-8B, Eval-DU+), (Gemma2-2B, Eval-DU+), and (Llama2-7B, TOFU+). We expect our findings to remain consistent across two datasets and multiple model families, supporting broader generalization to unseen models and datasets.

\paragraph{Model finetuning set-up.}
The number of paraphrases is 3 in both \texttt{FT-Mul} and \texttt{FT-Mul-Chunk}.
Fine-tuning prodcures all start from the public pre-trained models.
For Eval-DU+, we perform fine-tuning with Causal Language-Modeling (same objective as the pre-training~\citep{radford2018improving}), which minimizes the next-token prediction loss over all tokens in each training example.
In contrast, TOFU+ is structured in a QA format, so we adopt supervised fine-tuning~\citep{radford2018improving, ouyang2022training}: each QA pair is placed in a predefined QA template, and the objective is to minimize the loss only over the answer tokens.
We use the Adam optimizer for all fine-tuning experiments and update all model parameters during fine-tuning.
While ensuring that each model achieves a near-perfect fit on its fine-tuning data, we additionally evaluate general utility on standard LLM benchmarks to confirm that the models retain broad capabilities after fine-tuning.
Please check the general benchmark performance in the appendix.


%% file: sections/mem_vs_gen.tex
\section{Experiment Results}
\label{sec:exp_result}
\subsection{Empirical Study I: Effects of Paraphrased Texts on Knowledge Unlearning}
\label{sec:result_paraphrase}
\begin{table}[!t]
\caption{Average knowledge scores among finetuning set (FT Probs.) or unseen test set (Test Probs.).}
\label{tab:ft_utility}
\resizebox{\textwidth}{!}{%
	\begin{tabular}{c|cc|cc|cc|cc}
	\toprule
	& \multicolumn{2}{c|}{Llama2-7B, Eval-DU+} & \multicolumn{2}{c|}{Llama3-8B, Eval-DU+} & \multicolumn{2}{c|}{Gemma2-2B, Eval-DU+} & \multicolumn{2}{c}{Llama2-7B, TOFU+}\\
	\midrule
	& FT Probs. & Test Probs. & FT Probs. & Test Probs. & FT Probs. & Test Probs. & FT Probs. & Test Probs.\\
	\midrule
	\texttt{FT-Single} & 0.95 & 0.47 & 0.97 & 0.44 & 0.97 & 0.39 & 0.99 & 0.12  \\
	\texttt{FT-Mul} & 0.92 & 0.68 & 0.95 & 0.64 & 0.95	& 0.61 & 0.99 & 0.16  \\
	\texttt{FT-Mul-Chunk}\footnotemark & - & 0.46 & - & 0.43 & - & 0.40 & - & 0.13 \\
	\bottomrule	
	\end{tabular}
}
\end{table}
\footnotetext{FT Probs. is not applicable to \texttt{FT-Mul-Chunk} because the probability-based knowledge score is defined with respect to a single textual description of a knowledge piece $k$. In \texttt{FT-Mul-Chunk}, the related words of $k$ can be distributed and shared with other knowledge in the same text-chunk.}

In this section, we study how paraphrased descriptions in the fine-tuning dataset affect the difficulty of unlearning.
For four combinations of datasets and pre-trained LLMs, we fine-tune models using two training sets: \texttt{FT-Single}, where each knowledge piece is encoded with a single description, and \texttt{FT-Mul}, where each is encoded with multiple paraphrased descriptions.
Table~\ref{tab:ft_utility} reports the performance of these fine-tuned models on both fine-tuning and test sets.

To compare their unlearning behaviors, we apply six unlearning configurations (2 unlearning algorithms $\times$ 3 choices of unlearning data) across the four dataset–LLM combinations.
The results for the extraction trade-off are presented in Table~\ref{tab:single_mul_extraction}; results for the memorization trade-off, which exhibit similar trends, are included in the appendix.

\textbf{Main observation: fine-tuning with paraphrased descriptions (\texttt{FT-Mul}) consistently leads to more effective unlearning.}
We can observe this advantage across datasets, model types, unlearning methods, and evaluation metric.
Out of 48 total comparisons when evaluated with the \emph{extraction} trade-off, \texttt{FT-Mul} outperforms (or matches) \texttt{FT-Single} in 44 cases.
Similarly, \texttt{FT-Mul} outperforms (or matches) \texttt{FT-Single} in 39/48 cases when evaluated with the \emph{memorization} trade-off.

\textbf{Practical strategy I: paraphrasing.} Incorporate multiple paraphrased descriptions of each knowledge piece during fine-tuning -- or simply, augment the fine-tuning set through the addition of paraphrases. As suggested by our results, this would improve the effectiveness of unlearning by enhancing the model's ability to forget targeted information while preserving unrelated content.

\begin{table}[!t]
\caption{\texttt{FT-Single} versus \texttt{FT-Mul}: Norm-AUC ($\uparrow$) / AUC ($\uparrow$) of the extraction trade-off. We \textbf{bold} the better score between \texttt{FT-Mul} and \texttt{FT-Single}. Across all comparisons, we observe that \texttt{FT-Mul} outperforms or matches \texttt{FT-Single} in 44 out of 48 cases.
}
\label{tab:single_mul_extraction}
\resizebox{\textwidth}{!}{%
\begin{tabular}{c|c|ccc|ccc}
\toprule
 Model, & \multirow{2}{*}{FT Choices}                 & \multicolumn{3}{c|}{GA}                                   & \multicolumn{3}{c}{TV}                                   \\
      Dataset     &           & \texttt{UL-Exact} & \texttt{UL-Single} & \texttt{UL-Mul} & \texttt{UL-Exact} & \texttt{UL-Single} & \texttt{UL-Mul} \\
\midrule
Llama2-7B, & \texttt{FT-Single}    & \textbf{0.59} / 0.52 & 0.62 / 0.53 & 0.63 / 0.53 & 0.57 / 0.52 & \textbf{0.62} / 0.53 & 0.69 / 0.55 \\
Eval-DU+ & \texttt{FT-Mul}       & 0.55 / \textbf{0.54} & \textbf{0.63} / \textbf{0.58} & \textbf{0.64} / \textbf{0.58} & \textbf{0.65} / \textbf{0.59} & \textbf{0.62} / \textbf{0.57} & \textbf{0.72} / \textbf{0.62} \\
\midrule
Llama3-8B, & \texttt{FT-Single}    &  0.55 / 0.52  &   0.60 /  0.53  &  \textbf{0.62} / 0.54 &  0.62 / 0.54  & \textbf{0.63} / 0.54 & \textbf{0.68} / 0.55    \\
Eval-DU+ & \texttt{FT-Mul}   & \textbf{0.62} / \textbf{0.58}   &  \textbf{0.61} / \textbf{0.57}   &  0.60 / \textbf{0.57} & \textbf{0.68} / \textbf{0.59}  & 0.59 / \textbf{0.56}    &   0.66 / \textbf{0.59}  \\
\midrule
Gemma2-2B,& \texttt{FT-Single}    &         0.52 / 0.52 &  0.57 / 0.53   &  0.61 / 0.54    &   0.60 / 0.54  &  0.59 / 0.53 &   0.66 / 0.53                        \\
 Eval-DU+ & \texttt{FT-Mul}       &   \textbf{0.60} / \textbf{0.55}  &  \textbf{0.63} / \textbf{0.56}  &  \textbf{0.65} / \textbf{0.57}  &  \textbf{0.70} / \textbf{0.58} &   \textbf{0.65} / \textbf{0.56}  & \textbf{0.67}  / \textbf{0.57}               \\
\midrule
Llama2-7B, & \texttt{FT-Single}    &  \textbf{0.65} / 0.50 & 0.59 / 0.50 & 0.59 / 0.50 & 0.54 / 0.50 & 0.59 / 0.50 & 0.59 / 0.50\\
TOFU+ & \texttt{FT-Mul}       &  0.62 / \textbf{0.51} & \textbf{0.63} / \textbf{0.51} & \textbf{0.64} / \textbf{0.51} & \textbf{0.58} / \textbf{0.51} & \textbf{0.64} / \textbf{0.51} & \textbf{0.65} / \textbf{0.51} \\
\bottomrule
\end{tabular}
}
\end{table}

%% file: sections/group_ft.tex
\subsection{Empirical Study II: Understanding the Unlearning from Text Chunks}
\label{sec:result_chunk}



\begin{table}[!t]
\centering
\caption{Norm-AUC ($\uparrow$) of the extraction trade-off when the finetuning is done by \texttt{FT-Mul-Chunk}. The most Norm-AUC values are close to 0.5 when unlearning with Eval-DU+, indicating limited effectiveness in unlearning. In contrast, with TOFU+, the Norm-AUC values are generally higher.}
\label{tab:ft_mul_chunk}
\resizebox{\textwidth}{!}{%
\begin{tabular}{c|ccc|ccc}
\toprule
\multirow{2}{*}{Model \& Dataset} & \multicolumn{3}{c|}{GA}                                   & \multicolumn{3}{c}{TV}                                   \\
 & \texttt{UL-Exact} & \texttt{UL-Single} & \texttt{UL-Mul} & \texttt{UL-Exact} & \texttt{UL-Single} & \texttt{UL-Mul} \\
\midrule
Llama2-7B, Eval-DU+ & 0.53 & 0.50 & 0.49 & 0.57 & 0.55 & 0.56\\
\midrule
Llama3-8B, Eval-DU+ & 0.53   &  0.46   &  0.47  & 0.54  & 0.51   & 0.52  \\
\midrule
Gemma2-2B, Eval-DU+ & 0.54  &  0.47  & 0.45  &  0.61  &  0.48   &   0.52                             \\
\midrule
Llama2-7B, TOFU+ & 0.59 & 0.58 & 0.58 & 0.59 & 0.60 & 0.60\\
\bottomrule
\end{tabular}
}
\end{table}

In this section, we examine the task of unlearning knowledge embedded within larger text chunks. We use the \texttt{FT-Mul-Chunk} setup for fine-tuning across four combinations of datasets and pre-trained LLMs, and evaluate unlearning for six configurations of unlearning methods and data encodings.

\textbf{Observation: Unlearning individual knowledge pieces is more difficult when they are entangled with retained content in the same text chunk.}
Table~\ref{tab:ft_mul_chunk} reports Norm-AUC values for the extraction trade-off. We observe that for \texttt{Eval-DU+} (across all three pre-trained LLMs), most Norm-AUC values are close to 0.5 across the six unlearning configurations. This indicates that unlearning tends to remove both target and retained knowledge from the target LLM at a similar rate — suggesting that the unlearning process is largely ineffective in selectively removing the intended content. 
Particularly, models fine-tuned with \texttt{FT-Mul-Chunk} exhibit knowledge scores (test probs.) comparable to those fine-tuned with \texttt{FT-Single} (test probs.). 
Given this similarity, unlearning from \texttt{FT-Mul-Chunk} still consistently results in lower Norm-AUC scores than \texttt{FT-Single}. 

A plausible explanation lies in the entanglement of the descriptions of the target and the non-target knowledge within text chunks. 
This is supported by the comparison between Eval-DU+ and TOFU+. As shown in Table~\ref{tab:ft_mul_chunk}, Norm-AUC values are noticeably higher for \texttt{TOFU+}, suggesting more effective unlearning. The key distinction lies in how the unlearn–retain split is defined. In \texttt{Eval-DU+}, target and retained knowledge are incorporated within shared chunks (see the example in Figure~\ref{fig:k_example}) while TOFU+ organizes chunks so that each is either fully composed of unlearned knowledge or entirely retained. This structural alignment enables unlearning methods to act on self-contained units, thereby resulting in increased unlearning effectiveness. 
These results indicate that representational entanglement between unlearn and retain split can be a primary obstacle to selective unlearning.
This explanation further motivates the following \textbf{two hypotheses}.

\begin{table}[!t]
\centering
\caption{Norm-AUC ($\uparrow$) of the extraction trade-off when the finetuning is done by \texttt{FT-Mul-Chunk} and \textbf{the unlearn split is aligned with the partitions of text chunks}. We also report the {\color{green}difference} of Norm-AUC if compared with the results of the original unlearning split in Table~\ref{tab:ft_mul_chunk}. 
}

\label{tab:ft_mul_chunk_new_split}
\resizebox{\textwidth}{!}{%
\begin{tabular}{c|ccc|ccc}
\toprule
\multirow{2}{*}{Model \& Dataset} & \multicolumn{3}{c|}{GA}                                   & \multicolumn{3}{c}{TV}                                   \\
 & \texttt{UL-Exact} & \texttt{UL-Single} & \texttt{UL-Mul} & \texttt{UL-Exact} & \texttt{UL-Single} & \texttt{UL-Mul} \\
\midrule
Llama2-7B, Eval-DU+ & 0.66 ({\color{green}+0.13}) & 0.56 ({\color{green}+0.06}) & 0.55 ({\color{green}+0.06}) & 0.60 ({\color{green}+0.03}) & 0.57 ({\color{green}+0.02}) & 0.59 ({\color{green}+0.03}) \\
\midrule
Llama3-8B, Eval-DU+ & 0.57 ({\color{green}+0.04})   &  0.52 ({\color{green}+0.06}) &  0.53 ({\color{green}+0.06}) &  0.63 ({\color{green}+0.09}) &     0.54 ({\color{green}+0.03})      &    0.55 ({\color{green}+0.03})   \\
\midrule
Gemma2-2B, Eval-DU+ & 0.61 ({\color{green}+0.07})  &  0.55 ({\color{green}+0.08})  &   0.56 ({\color{green}+0.11})   &   0.63 ({\color{green}+0.02})  &   0.55 ({\color{green}+0.07})  &   0.56 ({\color{green}+0.04})  \\
\bottomrule
\end{tabular}
}

\end{table}


\begin{table}[!t]
\centering
\caption{Norm-AUC ($\uparrow$) of the extraction trade-off under \texttt{FT-Mul-Chunk-Iso}, where  \textbf{the text chunks are concatenations of the individual knowledge descriptions}. We also report the {\color{green}difference} of Norm-AUC compared with the results of the original \texttt{FT-Mul-Chunk} in Table~\ref{tab:ft_mul_chunk}. 
}
\label{tab:ft_mul_chunk_iso}
\resizebox{0.75\textwidth}{!}{%
\begin{tabular}{c|cc|cc}
\toprule
\multirow{2}{*}{Dataset \& Model} & \multicolumn{2}{c|}{GA}                                   & \multicolumn{2}{c}{TV}                                   \\
& \texttt{UL-Single} & \texttt{UL-Mul} & \texttt{UL-Single} & \texttt{UL-Mul} \\
\midrule
Llama2-7B, Eval-DU+ & 0.61 ({\color{green}+0.11}) & 0.60 ({\color{green}+0.11}) & 0.64 ({\color{green}+0.09}) & 0.69 ({\color{green}+0.13}) \\
\midrule
Llama3-8B, Eval-DU+ & 0.59 ({\color{green}+0.13})  &  0.58 ({\color{green}+0.11})   &   0.61 ({\color{green}+0.10})  &   0.64 ({\color{green}+0.12}) \\
\midrule
Gemma2-2B, Eval-DU+ & 0.57  ({\color{green}+0.10})   &  0.57 ({\color{green}+0.14})  &  0.59 ({\color{green}+0.11})  &    0.61 ({\color{green}+0.09})   \\
\bottomrule
\end{tabular}
}
\end{table}

\textbf{Hypothesis 1: Unlearning is more effective when the unlearn split aligns with the chunk boundaries in the training data.}
To test this hypothesis, we construct a new unlearn split in Eval-DU+ that aligns more closely with how knowledge pieces are grouped within text chunks. 
In Eval-DU+, each chunk describes all facts related to a specific person. 
Therefore, we randomly select 12 people and include all knowledge pieces associated with them in the new unlearn split.\footnote{This split contains 102 knowledge pieces, closely matching the size 100 of the original unlearn split, thereby controlling for unlearn set size.}

We then evaluate the same six unlearning configurations on this new split using three LLMs. 
Table~\ref{tab:ft_mul_chunk_new_split} reports the corresponding Norm-AUC values for the extraction trade-off. 
Compared to the results in Table~\ref{tab:ft_mul_chunk}, we observe a consistent improvement in Norm-AUC, indicating more effective unlearning. 
These results support Hypothesis 1: unlearning is more effective when the unlearn split is aligned with the underlying structure of the text chunks.

\textbf{Hypothesis 2: Unlearning is more effective when the target knowledge is less entangled with the retain content within text chunks.}
We hypothesize that the difficulty of unlearning a single knowledge piece while preserving others in the same text chunk arises from entangled descriptions—that is, when unlearn and retain knowledge are interwoven within the same narrative. But what if the unlearn and retain pieces are presented in isolation within the chunk?

To explore this, we construct a new version of the fine-tuning data, denoted as \texttt{FT-Mul-Chunk-Iso}, where each text chunk is formed by simply concatenating independent sentence-level descriptions of the associated knowledge pieces. This ensures that each piece of knowledge is expressed separately, even when grouped in the same chunk. Below is an example:
\begin{tcolorbox}[
    colback=gray!5,
    colframe=black!75!black,
    fonttitle=\bfseries
]
\small\texttt{Parker Ross is the wife of Richard Perry. As a child, Reed Perry belongs to Richard Perry. Poppy Perry is Richard Perry's aunt...}
\end{tcolorbox}

We fine-tune LLMs using these disentangled chunks and evaluate unlearning effectiveness under the same six unlearning configurations, using the original unlearn split from \texttt{Eval-DU+}. Table~\ref{tab:ft_mul_chunk_iso} reports the corresponding Norm-AUC values for the extraction trade-off. Compared to the original \texttt{FT-Mul-Chunk} in Table~\ref{tab:ft_mul_chunk}, we observe consistent improvements in Norm-AUC. This means that unlearning is more effective when the unlearn and retain content are more clearly separated within the same text chunk.

\textbf{Practical strategy II: separating.} Design training data with unlearning in mind by identifying likely unlearning targets in advance (e.g., via a detector) and rewriting the corresponding data to separate potential target knowledge from retain content, either in standalone sentences or isolated text chunks.\footnote{This may seem paradoxical: if the unlearning target is known in advance, why not remove it before training? However, unlearning requests often arise after deployment, particularly when training data is collected from public sources. For instance, some celebrities may not want LLMs to retain family-related information from their Wikipedia pages, while others may prefer that it be preserved. Such preferences are difficult to anticipate at training time.}
This structural preparation can make post-hoc unlearning more effective, as supported by our empirical findings validating the two hypotheses.

%% file: sections/related_work.tex
\section{Related Work}
\label{sec:related_work}

\textbf{Machine unlearning for LLMs: algorithms.} Recently, machine unlearning for LLMS has emerged as an important area of research \cite{liu2025rethinking, si2023knowledge}. In this work, we focus on GA \cite{jang2022knowledge, barbulescu2024each} and TV (task vector) \cite{ilharco2022editing} methods. Other notable approaches include:
NPO  \cite{zhangnegative, bronec2025atyaephyra} which utilizes the DPO objective \cite{rafailov2023direct} treating the unlearn data as negative preference data, WHP uses a linear combination of the distributions induced by initial and a reinforced model as an unlearn model \cite{eldan2023s, liu2024revisiting}, UWC calibrates the post-unlearning parameters with the initial parameters to better preserve the model's utility \cite{wang2024unlearning}, GRU uses both the unlearning and retention gradients at each update step \cite{wang2024unlearning}. 
Regularizers are often employed to better preserve the model's utility. For example: augementing the unlearning objective with the retention gradient (GDR) \cite{mainitofu, zhangnegative, liu2022continual} and regularizing with the KL divergence on the retention set (KLR) \cite{mainitofu, zhangnegative}. Non-training based methods include: localization-informed unlearning \cite{li2024wmdp, meng2022locating, wu2023depn} which localize the components of the LLM related to the forget data and black-box in-context unlearning \cite{pawelczyk2023context}. Other recent promising approaches are \cite{jia2024soul, liu2024large, ji2024reversing, wang2024large, ishibashi2023knowledge, thaker2024guardrail, wang2025uipe, he2025deep}. 


\textbf{Machine unlearning for LLMs: evaluations.}
Evaluating the effectiveness machine unlearning method poses another challenge. As an example, \cite{eldan2023s} uses completion and question-answer probability-based scores, while \cite{lynch2024eight} proposes comparing the unlearned model and a model retrained on the retention data. UNCD uses Cognitive Diagnosis Modeling for fine-grained evaluation \cite{lang2025beyond}.  Besides TOFU (\cite{mainitofu}) and Eval-DU (\cite{wu2024evaluating}), several other benchmarks have been proposed to assess the effectiveness of unlearning in LLMs such as: WMDP - a dataset consisting of hazardous knowledge in multiple-choice format \cite{li2024wmdp} and RWKU for zero-shot
konwledge unlearning \cite{jin2024rwku}, MUSE proposes a comprehensive benchmark evaluating six desirable properties from the perspectives of both data owners and model deployers \cite{shi2024muse}, and PEBench for multimodal LLMs \cite{xu2025pebench}. Finally, \cite{thaker2024position} discusses the limitations of existing benchmarks. Beyond this it shows that entanglement of retain and unlearn data in test prompts decreases the evaluation score of an unlearned model.

%% file: sections/conclusion.tex
\section{Discussions and Conclusions}
\label{sec:conclusion}
\paragraph{Limitations and future work.} Although this paper focuses on the role of training data choices in unlearning, several other learning-time factors may also influence unlearning effectiveness. These include the model architecture (e.g., full-parameter tuning LoRA~\citep{hu2022lora}) and the learning algorithm (e.g., supervised fine-tuning vs. reinforcement learning~\citep{rafailov2023direct, lu2022quark}). A promising direction for future work is to systematically investigate how such factors impact the behavior and difficulty of unlearning.
Due to limited computational resources, our experiments are restricted to LLMs that undergo fine-tuning. While we believe the findings presented in this paper may generalize to the pretraining stage and to unlearning from pretrained models directly, validating this hypothesis remains an important avenue for future research when more resources are available.

\textbf{Conclusion.} In summary, this work takes an initial step toward understanding how learning-time knowledge encoding influences post-hoc unlearning in large language models. By isolating textual representation as the key variable and controlling for underlying factual content, we show that both paraphrasing diversity and data structure significantly impact unlearning effectiveness. Our empirical results reveal that using paraphrased representations and clearly separating the descriptions of knowledge in the unlearn and retain splits can greatly enhance the ability to remove targeted information while preserving unrelated content. These findings lay the groundwork for learning-time strategies that improve the adaptability and reliability of unlearning in LLMs.

%% file: sections/appendix.tex
The organization of this appendix is as below:
\begin{enumerate}
\item In Section	~\ref{sec:app_dataset}, we present the details of constructing benchmark datasets Eval-DU+ and TOFU+, including the detailed statistics of paraphrasing, the templates for generating the synthetic texts, and an illustration of calculating the knowledge score \textit{prbability} in Eval-DU+.
\item In Section~\ref{sec:app_exp_result}, we will present additional experimental results, including the performance of fine-tuned models on LLM general benchmarks, the unlearning results evaluated by memorization trade-off, and the full plots of trade-off curves used for calculating the Norm-AUC and AUC.
\item In Section~\ref{sec:app_exp_setup}, we will present the implementation details in our experiments, including the compute resources used in the epxeriments, the details of model fine-tuning, and the details of the unlearning.
\end{enumerate}
Our code for reproducing the results in the tables is anonymously released at \url{https://anonymous.4open.science/r/learning_time_shapes_unlearning-8F5A/README.md}.

\section{Details of Constructing Benchmark Datasets}
\label{sec:app_dataset}

\paragraph{Detailed statistics of paraphrasing.} We present the statistics of the paraphrasing and how they are used for learning, unlearning and evaluation in both datasets Eval-DU+ and TOFU+:
\begin{table*}[!h]
\resizebox{\textwidth}{!}{%
	\begin{tabular}{c|ccc|c}
	\toprule
		\multirow{2}{*}{Dataset} & \multicolumn{3}{c|}{\# paraphrasing for each $k$} & \multirow{2}{*}{\# paraphrasing in \texttt{FT-Mul-Chunk}} \\
		& \texttt{FT-Mul} & \texttt{UL-Mul} & Extraction Trade-Off \\ 
	\midrule
		Eval-DU+ & 3 & 3 & 3 & 3\\
		TOFU+ & 3 & 3 & 1 & 3\\
	\bottomrule
	\end{tabular}
}
\end{table*}

In \texttt{FT-Single} and \texttt{UL-Single}, the description of each $k$ is picked randomly from \texttt{FT-Mul} and \texttt{UL-Mul} respectively.
The texts used in \texttt{UL-Exact} and the memorization trade-off depend on the definition of fine-tuning texts by definition.

\paragraph{Templates for the prompt when generating the texts through ChatGPT-4o.} Here are the templates of how we generate the paraphrased descriptions for each knowledge piece given the initial texts provided by each original dataset and the paraphrased text chunks for each group of knowledge.\\ 
\begin{tcolorbox}[
    colback=gray!5,
    colframe=blue!75!black,
    title=Templates of generating the paraphrased descriptions for each knowledge piece,
    fonttitle=\bfseries
]
\textbf{Eval-DU+}\\
\texttt{Could you help rephrase the sentence \{Initial Text\} while keeping the word \{Objective Word\}? Please give me 8 variations.}\\

\textbf{TOFU+}\\
\texttt{Could you help rephrase both the question and the answer below?}
\texttt{Question: \{Intial Question\}}\\
\texttt{Answer: \{Intial Answer\}}\\
\texttt{Please give me 7 variations and list them as a sequence of QAs, formated by 1., 2., ..., 7.}
\end{tcolorbox}

\begin{tcolorbox}[
    colback=gray!5,
    colframe=blue!75!black,
    title=Templates of generating the paraphrased text chunks for each knowledge group,
    fonttitle=\bfseries
]
\textbf{Eval-DU+}\\
\texttt{Here are the family information and biographic information for \{Person Name\}. Could you summarize all information in one paragraph and give me 5 versions of them by shuffling the order of these information:}\\
\texttt{\{Text Description of the 1st Knowledge Piece\}}\\
\texttt{$\ldots$}\\
\texttt{Please list the versions by 1., 2., ...}\\

\textbf{TOFU+}\\
\texttt{Could you help summarize all information in the following 20 question-answering into one question-answer pair?}\\
\texttt{1.}\\
\texttt{Question: \{1st Question\}}\\
\texttt{Answer: \{1st Answer\}}\\
\texttt{$\ldots$}\\
\texttt{Please give me 3 variations and do not miss any information. Please response in the format}\\
\texttt{Variation 1:}\\
\texttt{Question 1:$\ldots$}\\
\texttt{Answer 1:$\ldots$}\\
\texttt{$\ldots$}
\end{tcolorbox}

After collecting the responses from ChatGPT-4o, we did some text extractions in order to get a organized list of target paraphrased texts.

\paragraph{Examples of QAs in TOFU+.} We have shown the text examples of Eval-DU+ in the main paper. Here are the examples after augmenting the TOFU\\
\begin{tcolorbox}[
    colback=gray!5,
    colframe=blue!75!black,
    fonttitle=\bfseries
]
\textbf{The original QA in TOFU} (Used in \texttt{FT-Single})

Q: \textit{Who is this celebrated LGBTQ+ author from Santiago, Chile known for their true crime genre work?}

A: \textit{The author in question is Jaime Vasquez, an esteemed LGBTQ+ writer who hails from Santiago, Chile and specializes in the true crime genre.}
\\

\textbf{The paraphrased QA} (Used in \texttt{FT-Mul}, \texttt{UL-Single}, \texttt{UL-Mul}, or extraction trade-off)

Q: \textit{Could you tell me about the celebrated LGBTQ+ author from Santiago, Chile who excels in the true crime genre?}

A: \textit{Jaime Vasquez is the celebrated author recognized within the LGBTQ+ community and beyond for their exceptional work in true crime, hailing from Santiago, Chile}
\\

\textbf{The big QA} (Used in \texttt{FT-Mul-Chunk})

Q: \textit{Who is Jaime Vasquez, and what is notable about his contributions to literature?}

A: \textit{{\color{blue}Jaime Vasquez is a celebrated LGBTQ+ author from Santiago, Chile}, born on February 25, 1958. With a father ... {\color{blue}he channels his passion for storytelling into the true crime genre}. His award-winning books, including ...}

\end{tcolorbox}

\paragraph{Examples of calculating probabilities in Eval-DU+.} 
In Eval-DU+, each knowledge piece has the structure tuple of (s, r, o). We are able to identify the keywords for s, r, or o in a given text description. For example, here is a text description for (\textit{Richard Perry}, \textit{father}, \textit{Reid Perry}) and we highlight the corresponding keywords.
$$
\textit{{\color{blue}Reid Perry} has {\color{blue}Richard Perry} as his {\color{blue}father}.}
$$
Then, we can calculate the likelihood of the keyword appearing the last in this sentence, which is \textit{father}, for a given LLM which modelizes the likelihood function $\pi_{\theta}$.

\section{Additional Results}
\label{sec:app_exp_result}
\paragraph{Performance of fine-tuned models on LLM general benchmarks.}
While ensuring that each model achieves a near-perfect fit on its fine-tuning data, we additionally evaluate general utility on three standard LLM benchmarks: \textit{MMLU}~\citep{hendrycks2021measuring} for multi-domain language understanding, \textit{PIQA}~\citep{bisk2020piqa} for commonsense reasoning, and \textit{RACE}~\citep{lai-etal-2017-race} for reading comprehension.
The results are presented in Table~\ref{tab:general_utility}. We observe that fine-tuning does not significantly degrade performance on these general tasks, confirming that the models retain broad capabilities.

\begin{table*}[!h]
\caption{Pretrained and finetuned LLMs on three general utility benchmarks.}
\label{tab:general_utility}
\resizebox{\textwidth}{!}{%
\begin{tabular}{c|ccc|ccc|ccc|ccc}
\toprule
LLM \& Dataset & \multicolumn{3}{c|}{Llama2-7B on Eval-DU+} & \multicolumn{3}{c|}{Llama3-8B on Eval-DU+}    & \multicolumn{3}{c|}{Gemma2-2B on Eval-DU+}     & \multicolumn{3}{c|}{Llama2-7B on TOFU+}     \\     
Metric & MMLU  & PIQA  & RACE & MMLU  & PIQA  & RACE & MMLU  & PIQA  & RACE & MMLU  & PIQA  & RACE  \\
\midrule
Pre-train  &   0.400    &   0.778     &   0.396       & 0.621 & 0.807 & 0.402 & 0.496 & 0.791 & 0.373 &   0.400    &   0.778     &   0.396   \\

\texttt{FT-Single}        & 0.383 & 0.775 & 0.398 & 0.612 & 0.801 & 0.386 & 0.496 & 0.798 & 0.380  & 0.335 & 0.758 & 0.398\\
\texttt{FT-Mul}          & 0.368 & 0.782 & 0.392  & 0.612 & 0.800 &	0.389 & 0.486 & 0.792 & 0.365 & 0.332 & 0.773 & 0.402\\
\texttt{FT-Mul-Trunk}         & 0.353 & 0.777 & 0.402              & 0.616 & 0.793 & 0.405   & 0.492 & 0.773 & 0.385    & 0.284 & 0.779 & 0.414\\
\bottomrule
\end{tabular}
}
\end{table*}

\paragraph{The unlearning results evaluated by  memorization trade-off.} Our main observation is that fine-tuning with paraphrased descriptions (\texttt{FT-Mul}) consistently leads to more effective unlearning when evaluated with the memorization trade-off.
We can observe this advantage across datasets, model types and unlearning methods:
out of 48 total comparisons \texttt{FT-Mul} outperforms (or matches) \texttt{FT-Single} in 39 cases when evaluated with the \emph{memorization} trade-off.

\begin{table}[!h]
\caption{\texttt{FT-Single} versus \texttt{FT-Mul} Norm-AUC ($\uparrow$) / AUC ($\uparrow$) of the memorization trade-off. We \textbf{bold} the better score between \texttt{FT-Mul} and \texttt{FT-Single}. Across all comparisons, we observe \texttt{FT-Mul} outperforms or matches \texttt{FT-Single} in 39 out of 48 cases.}
\label{tab:single_mul_memorization}
\resizebox{\textwidth}{!}{%
\begin{tabular}{c|c|ccc|ccc}
\toprule
Dataset, & \multirow{2}{*}{FT Choices}                 & \multicolumn{3}{c|}{GA}                                   & \multicolumn{3}{c}{TV}                                   \\
Model &                      & \texttt{UL-Exact} & \texttt{UL-Single} & \texttt{UL-Mul} & \texttt{UL-Exact} & \texttt{UL-Single} & \texttt{UL-Mul} \\
\midrule
Llama2-7B,  & \texttt{FT-Single}    & \textbf{0.66} / \textbf{0.64} & 0.54 / 0.54 & 0.56 / 0.55 & 0.64 / 0.63 & 0.58 / 0.57 & 0.65 / 0.63\\
Eval-DU+ & \texttt{FT-Mul}       &  0.60 / 0.59 & \textbf{0.59} / \textbf{0.59} & \textbf{0.58} / \textbf{0.58} & \textbf{0.68} / \textbf{0.66} & \textbf{0.60} / \textbf{0.59} & \textbf{0.69} / \textbf{0.67}\\
\midrule
Llama3-8B, & \texttt{FT-Single}    &   0.63 / 0.63  & 0.56 / 0.56  &  0.56 / 0.56  &  0.69 / 0.68 & 0.56 / 0.56 &  \textbf{0.63} / \textbf{0.63}  \\
Eval-DU+ & \texttt{FT-Mul}       &   \textbf{0.64} / \textbf{0.64}  &  \textbf{0.57} / \textbf{0.57} &  \textbf{0.58} / \textbf{0.58}  &  \textbf{0.72} / \textbf{0.70}  & \textbf{0.56} / \textbf{0.57}   &  \textbf{0.63} / 0.62  \\
\midrule
Gemma2-2B,  & \texttt{FT-Single}    &      0.60 / 0.60 & \textbf{0.63} / \textbf{0.62} & \textbf{0.63}/ \textbf{0.62} & 0.66 / 0.65 & \textbf{0.60} / \textbf{0.59} & 0.60 / 0.60                   \\
Eval-DU+ & \texttt{FT-Mul}       &   \textbf{0.62} / \textbf{0.61} & 0.62 / \textbf{0.62} & 0.60 / 0.59 & \textbf{0.75} / \textbf{0.73} & 0.59 / \textbf{0.59} & \textbf{0.63} / \textbf{0.62}      \\
\midrule
Llama2-7B, & \texttt{FT-Single}    &  \textbf{0.90} / \textbf{0.90} & 0.63 / 0.63 & 0.61 / 0.61 & 0.64 / 0.64 & 0.74 / 0.74 & 0.68 / 0.67\\
TOFU+ & \texttt{FT-Mul}       &  0.78 / 0.77 & \textbf{0.70} / \textbf{0.69} & \textbf{0.74} / \textbf{0.73} & \textbf{0.69} / \textbf{0.70} & \textbf{0.76} / \textbf{0.76} & \textbf{0.78} / \textbf{0.77}\\
\bottomrule
\end{tabular}
}
\end{table}

\paragraph{Full plots of trade-off curves.} For completion, we attach the full trade-off curves for calculating Norm-AUC and AUC. 

\begin{figure}[!t]
\centering
\begin{subfigure}[t]{0.32\textwidth}
    \includegraphics[width=\linewidth]{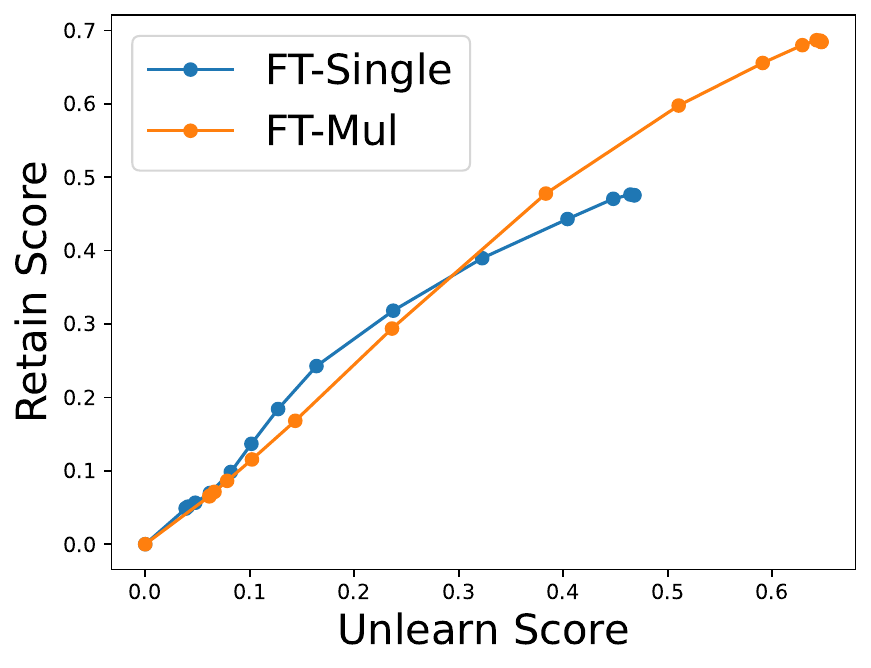}
    \caption{\texttt{UL-Exact}, GA}
\end{subfigure}
\hfill
\begin{subfigure}[t]{0.32\textwidth}
    \includegraphics[width=\linewidth]{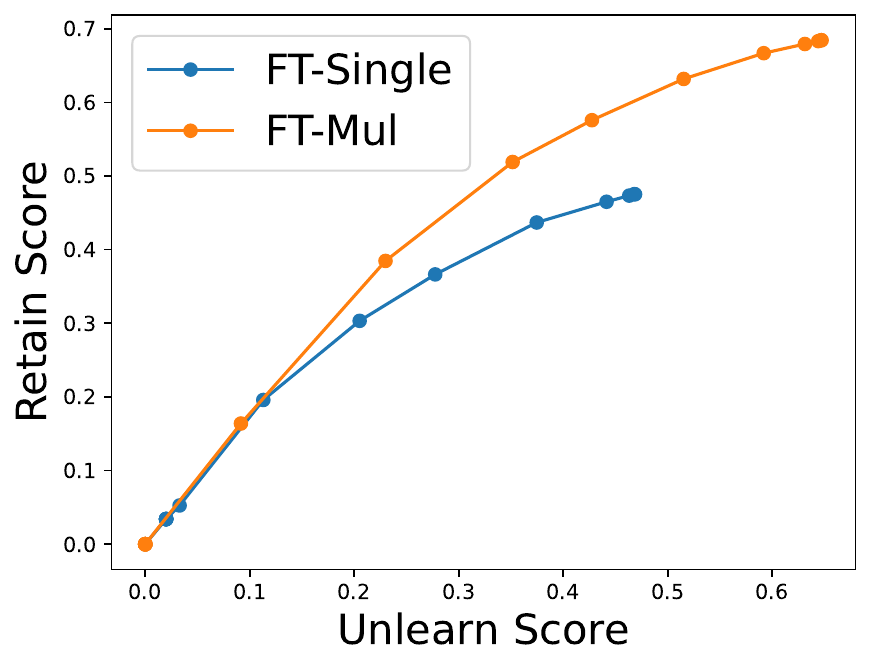}
    \caption{\texttt{UL-Single}, GA}\end{subfigure}
\hfill
\begin{subfigure}[t]{0.32\textwidth}
    \includegraphics[width=\linewidth]{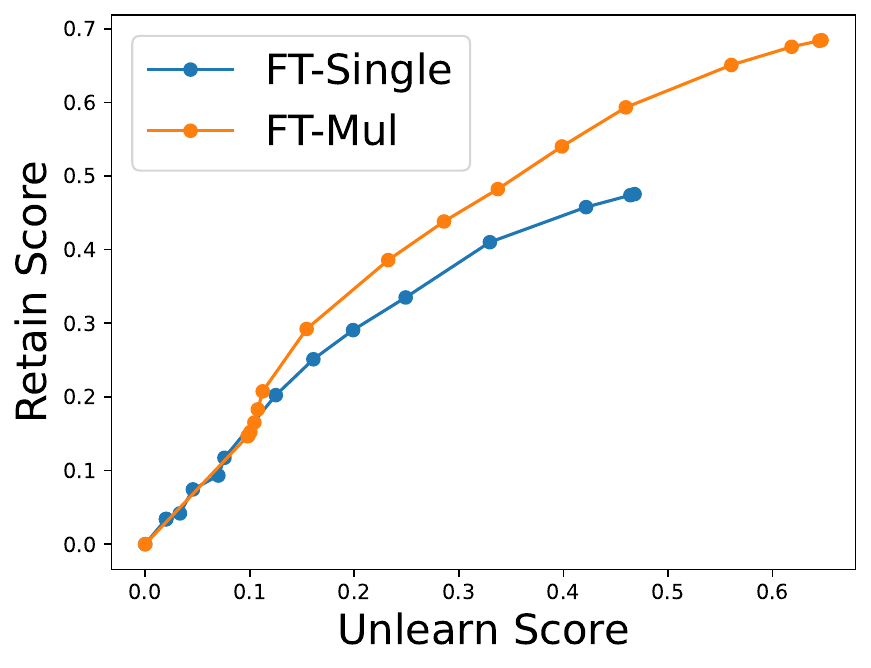}
    \caption{\texttt{UL-Mul}, GA}
\end{subfigure}

\begin{subfigure}[t]{0.32\textwidth}
    \includegraphics[width=\linewidth]{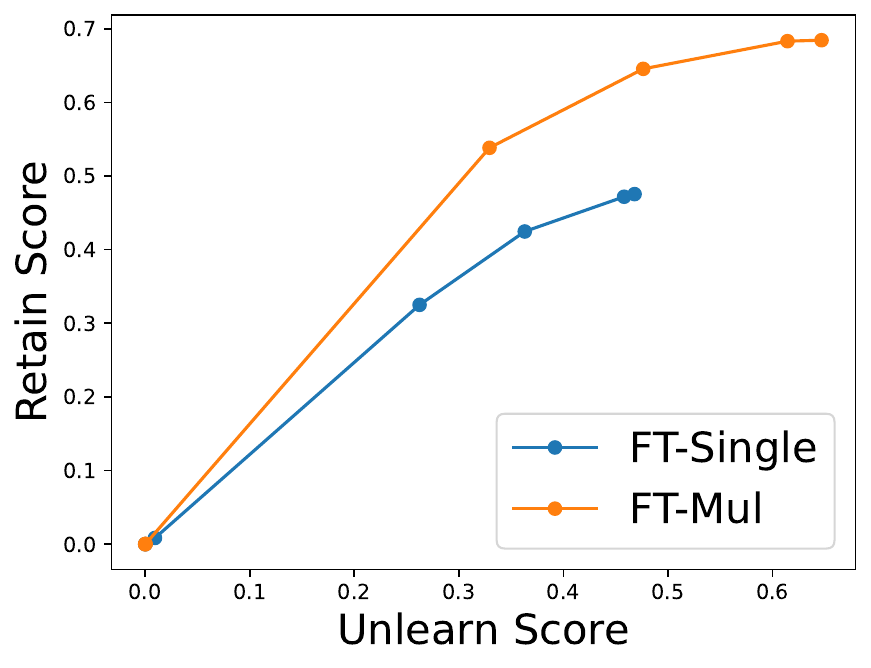}
    \caption{\texttt{UL-Exact}, TV}
\end{subfigure}
\hfill
\begin{subfigure}[t]{0.32\textwidth}
    \includegraphics[width=\linewidth]{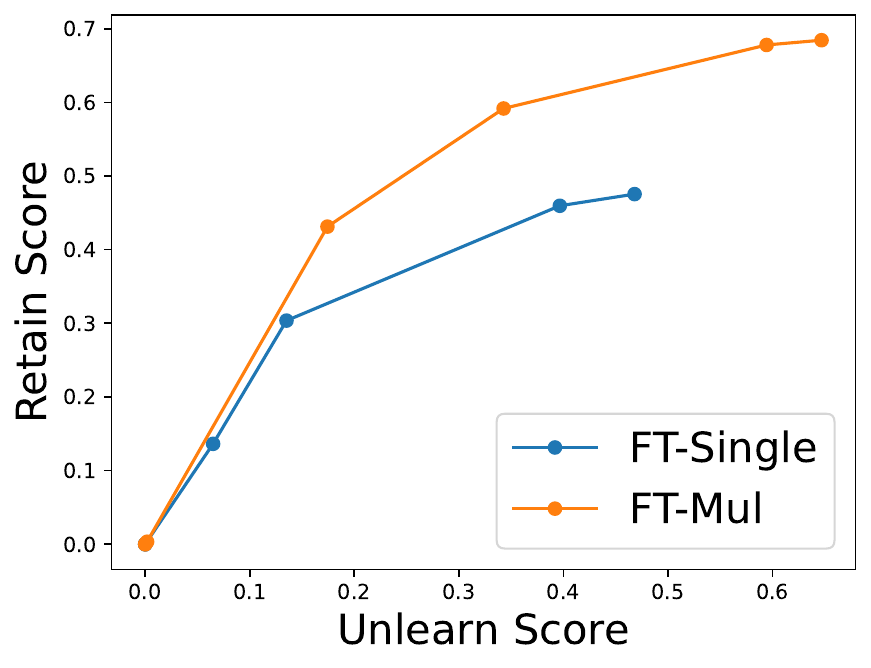}
    \caption{\texttt{UL-Single}, TV}\end{subfigure}
\hfill
\begin{subfigure}[t]{0.32\textwidth}
    \includegraphics[width=\linewidth]{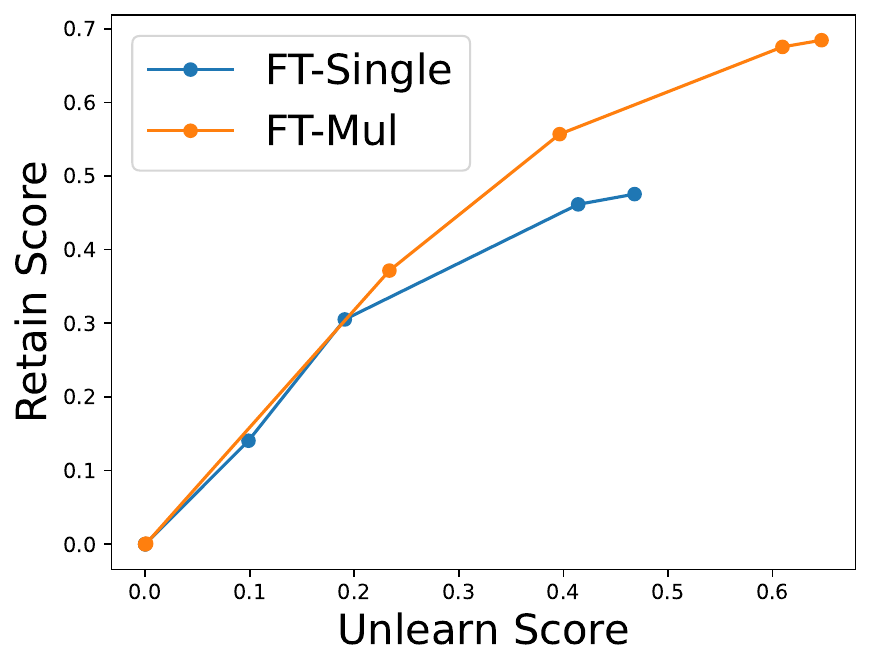}
    \caption{\texttt{UL-Mul}, TV}
\end{subfigure}

\caption{Vanilla \textbf{extraction} trade-off curves for three choices of unlearning data and two unlearning algorithms on \textbf{Eval-DU+} and \textbf{Llama2-7B}, when comparing \texttt{FT-Mul} and \texttt{FT-Single}}
\label{fig:extraction_curves_eval_du_llama2-7b}
\end{figure}

\begin{figure}[!t]
\centering
\begin{subfigure}[t]{0.32\textwidth}
    \includegraphics[width=\linewidth]{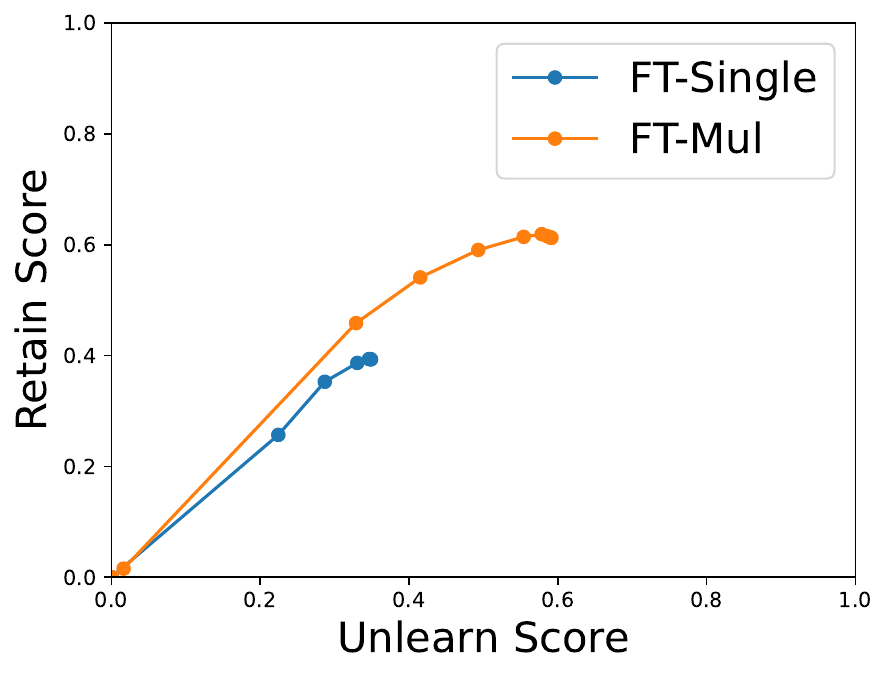}
    \caption{\texttt{UL-Exact}, GA}
\end{subfigure}
\hfill
\begin{subfigure}[t]{0.32\textwidth}
    \includegraphics[width=\linewidth]{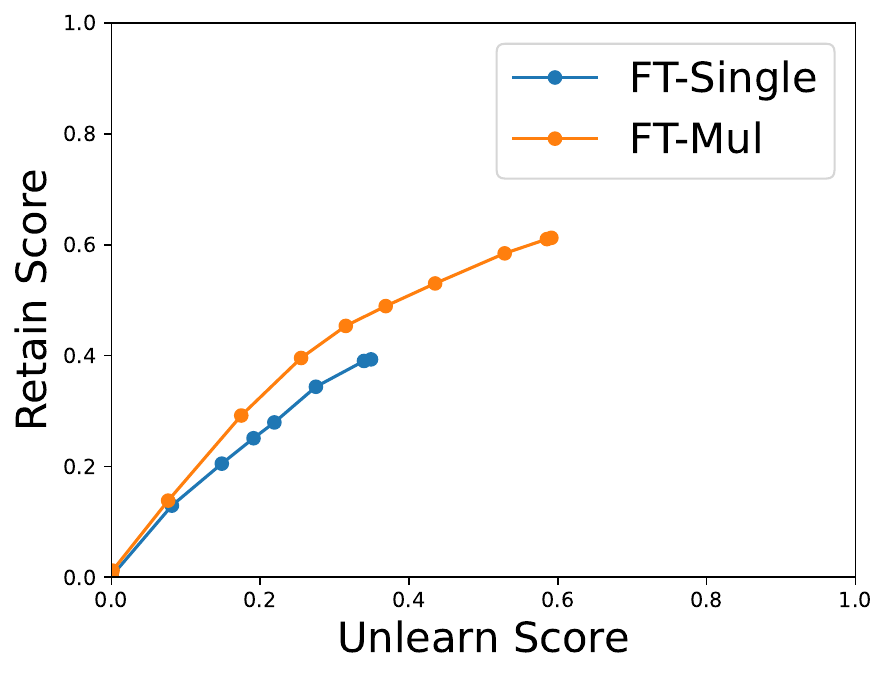}
    \caption{\texttt{UL-Single}, GA}\end{subfigure}
\hfill
\begin{subfigure}[t]{0.32\textwidth}
    \includegraphics[width=\linewidth]{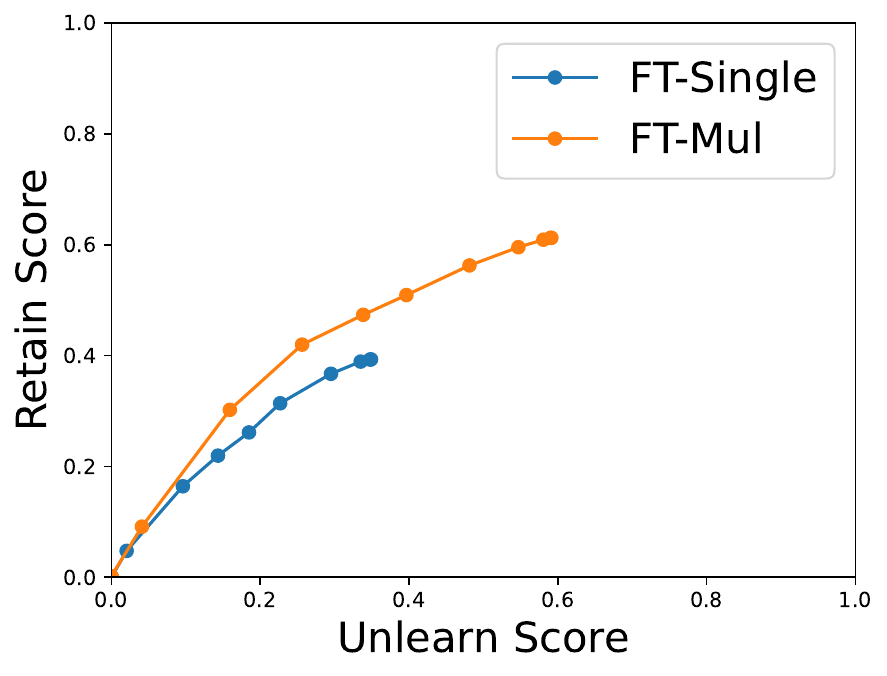}
    \caption{\texttt{UL-Mul}, GA}
\end{subfigure}

\begin{subfigure}[t]{0.32\textwidth}
    \includegraphics[width=\linewidth]{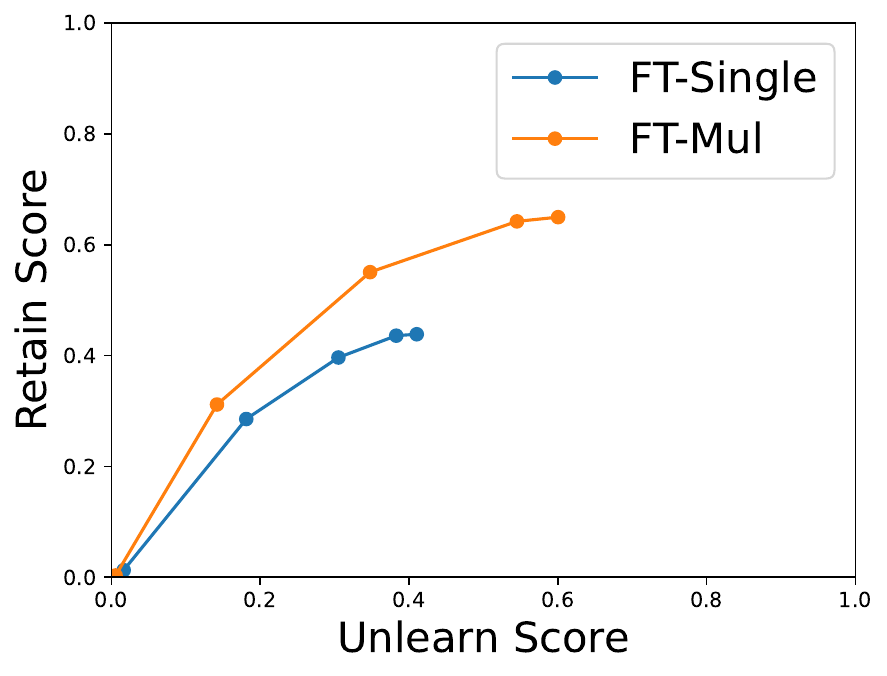}
    \caption{\texttt{UL-Exact}, TV}
\end{subfigure}
\hfill
\begin{subfigure}[t]{0.32\textwidth}
    \includegraphics[width=\linewidth]{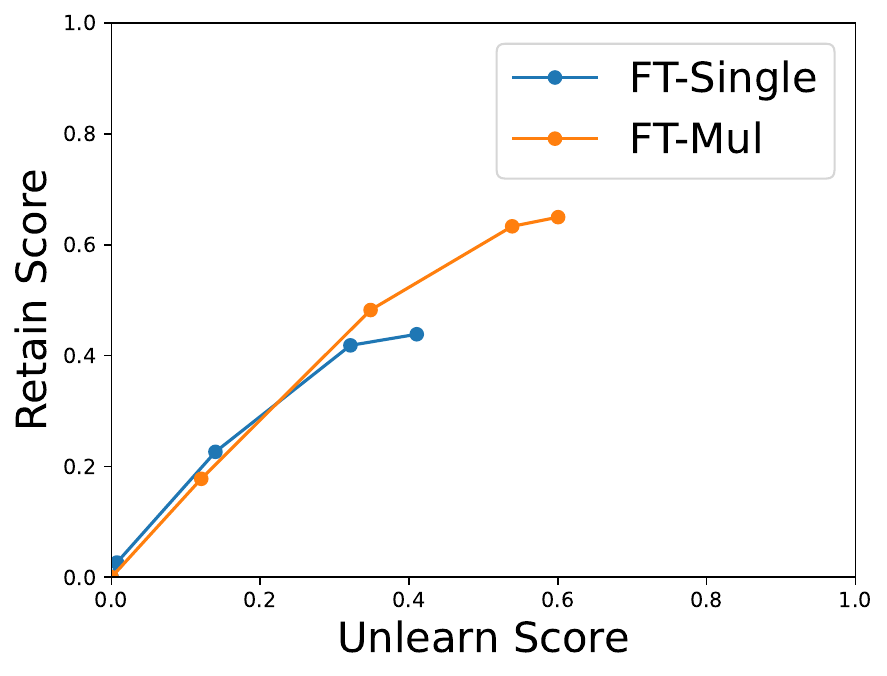}
    \caption{\texttt{UL-Single}, TV}\end{subfigure}
\hfill
\begin{subfigure}[t]{0.32\textwidth}
    \includegraphics[width=\linewidth]{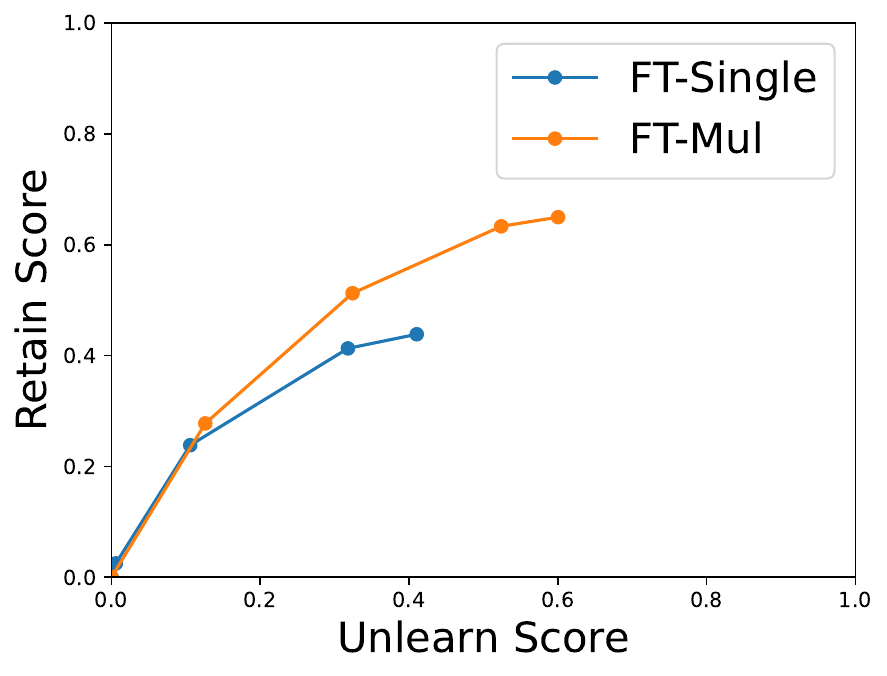}
    \caption{\texttt{UL-Mul}, TV}
\end{subfigure}

\caption{Vanilla \textbf{extraction} trade-off curves for three choices of unlearning data and two unlearning algorithms on \textbf{Eval-DU+} and \textbf{Llama3-8B}, when comparing \texttt{FT-Mul} and \texttt{FT-Single}}
\label{fig:extraction_curves_eval_du_llama3-8b}
\end{figure}

\begin{figure}[!t]
\centering
\begin{subfigure}[t]{0.32\textwidth}
    \includegraphics[width=\linewidth]{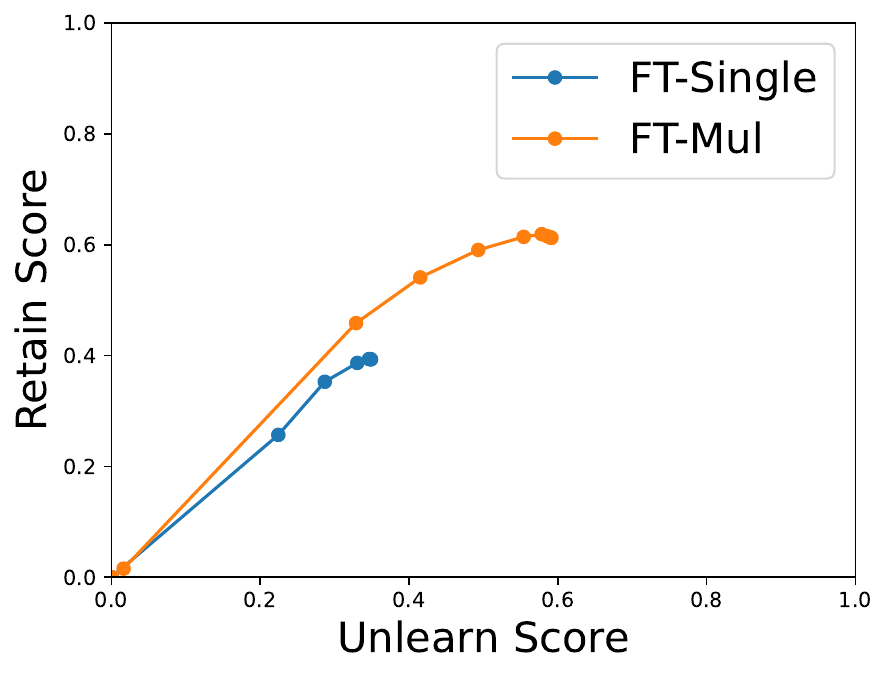}
    \caption{\texttt{UL-Exact}, GA}
\end{subfigure}
\hfill
\begin{subfigure}[t]{0.32\textwidth}
    \includegraphics[width=\linewidth]{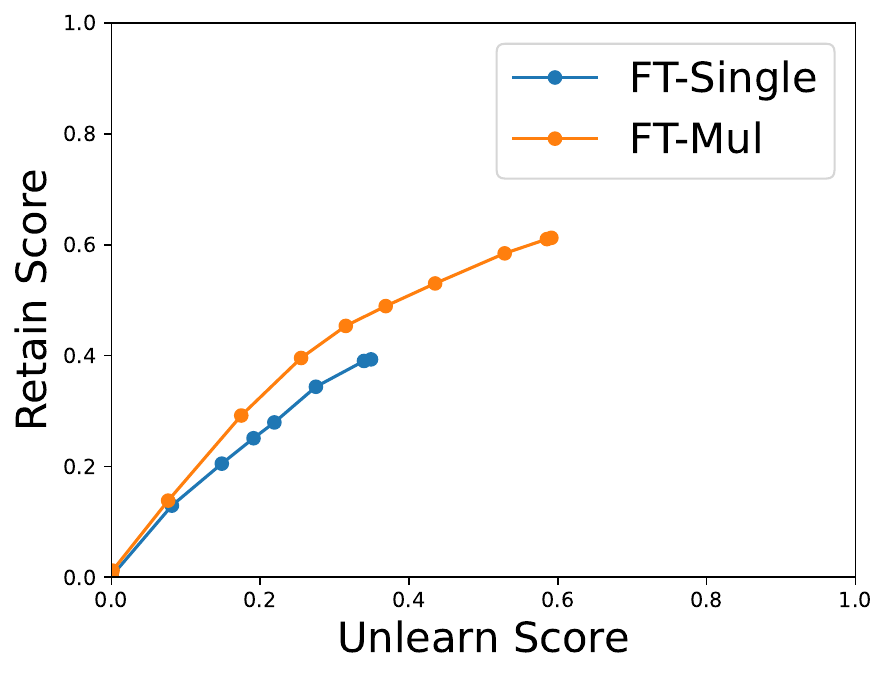}
    \caption{\texttt{UL-Single}, GA}\end{subfigure}
\hfill
\begin{subfigure}[t]{0.32\textwidth}
    \includegraphics[width=\linewidth]{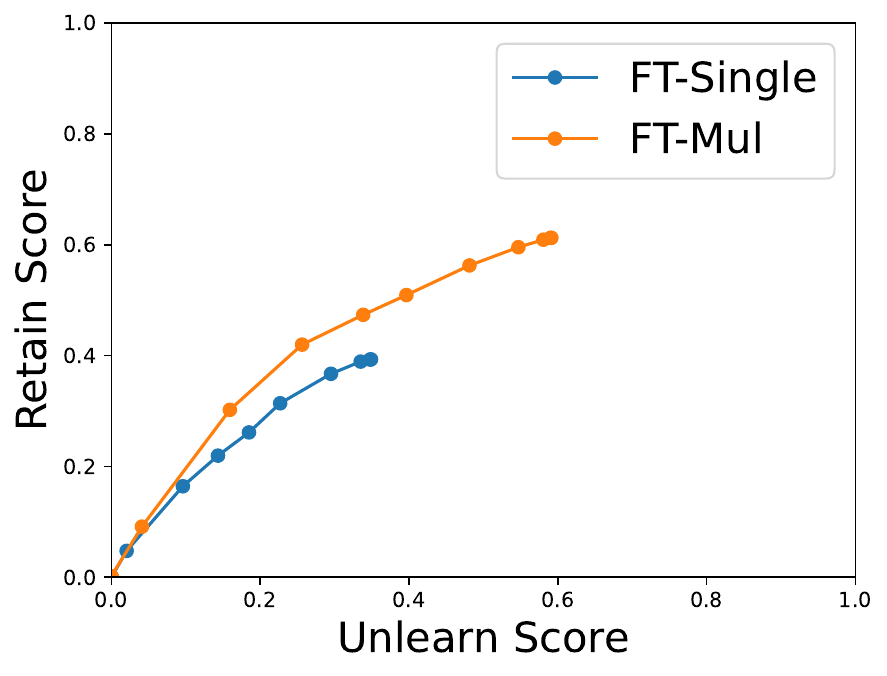}
    \caption{\texttt{UL-Mul}, GA}
\end{subfigure}

\begin{subfigure}[t]{0.32\textwidth}
    \includegraphics[width=\linewidth]{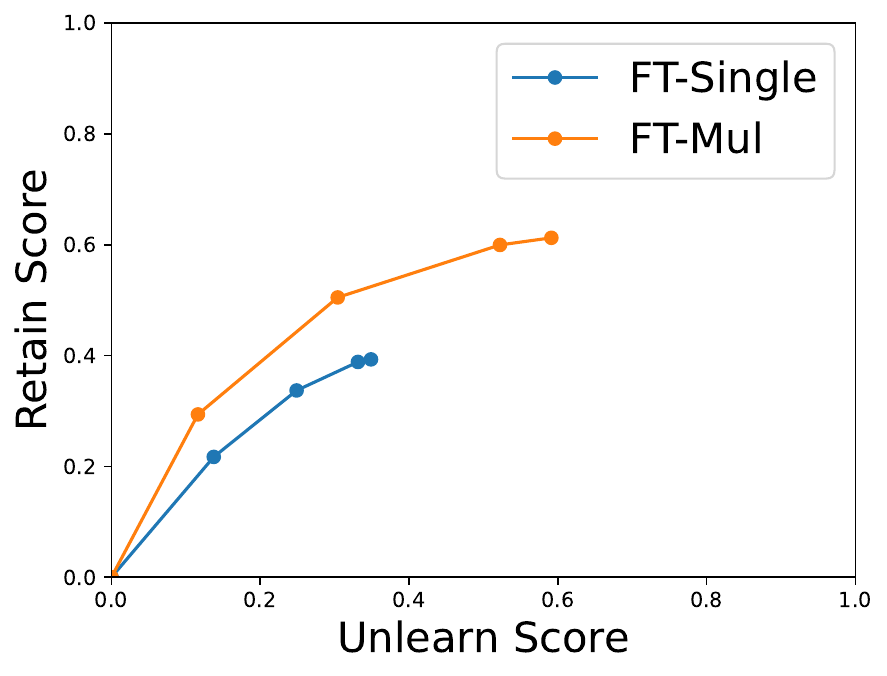}
    \caption{\texttt{UL-Exact}, TV}
\end{subfigure}
\hfill
\begin{subfigure}[t]{0.32\textwidth}
    \includegraphics[width=\linewidth]{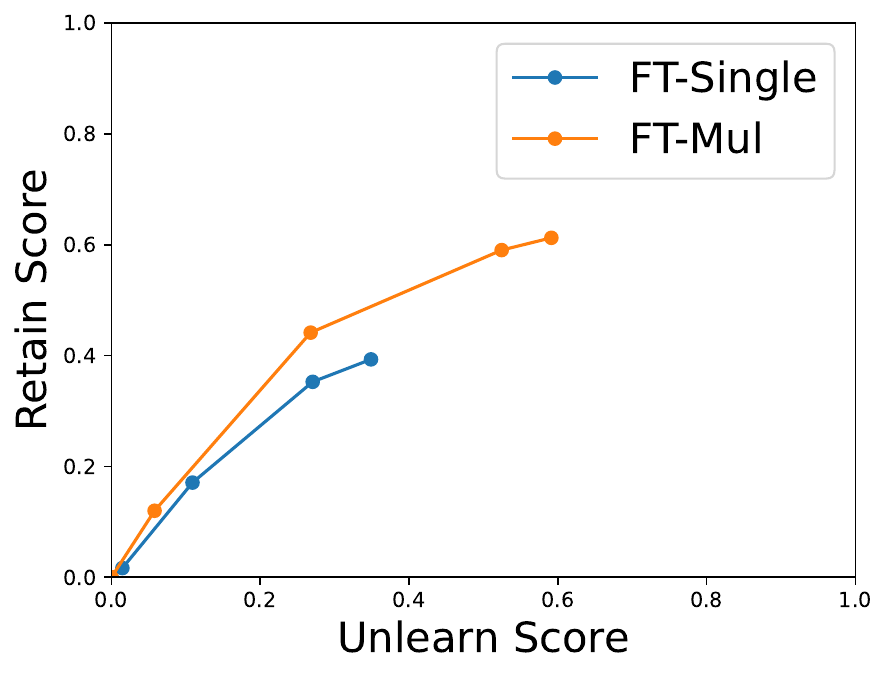}
    \caption{\texttt{UL-Single}, TV}\end{subfigure}
\hfill
\begin{subfigure}[t]{0.32\textwidth}
    \includegraphics[width=\linewidth]{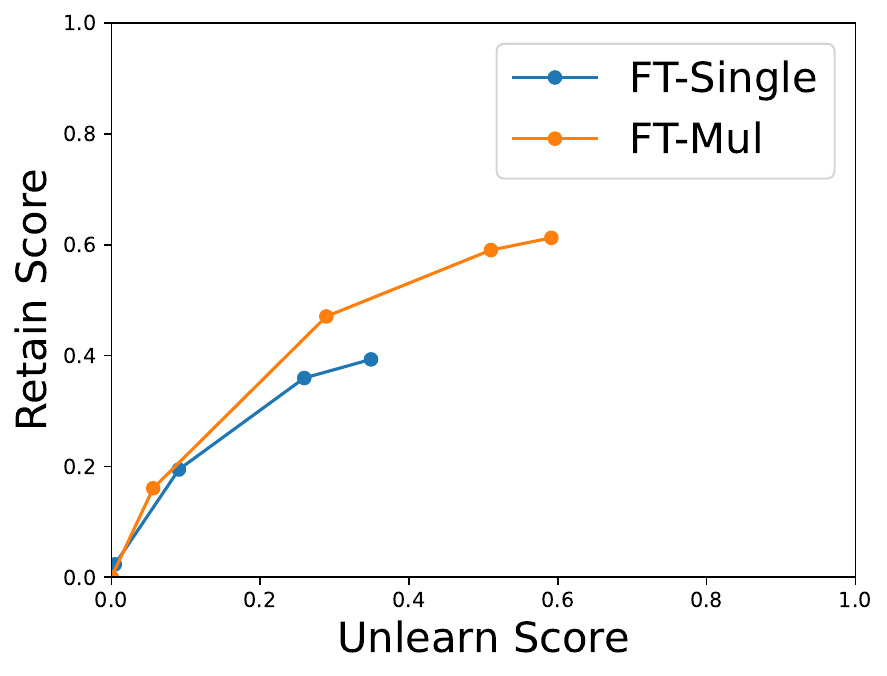}
    \caption{\texttt{UL-Mul}, TV}
\end{subfigure}

\caption{Vanilla \textbf{extraction} trade-off curves for three choices of unlearning data and two unlearning algorithms on \textbf{Eval-DU+} and \textbf{Gemma2-2B}, when comparing \texttt{FT-Mul} and \texttt{FT-Single}}
\label{fig:extraction_curves_eval_du_gemma2-2b}
\end{figure}

\begin{figure}[!t]
\centering
\begin{subfigure}[t]{0.32\textwidth}
    \includegraphics[width=\linewidth]{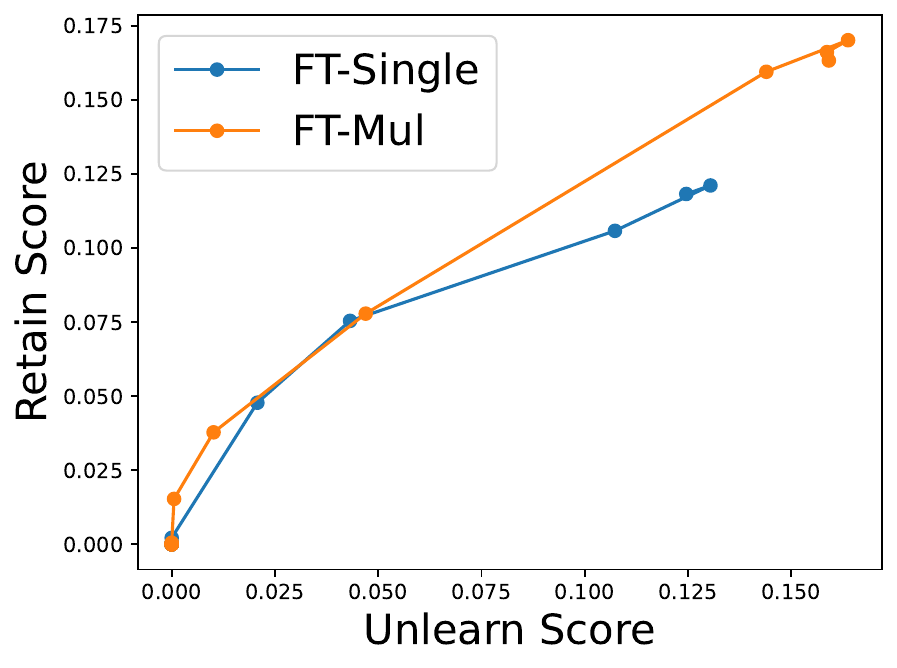}
    \caption{\texttt{UL-Exact}, GA}
\end{subfigure}
\hfill
\begin{subfigure}[t]{0.32\textwidth}
    \includegraphics[width=\linewidth]{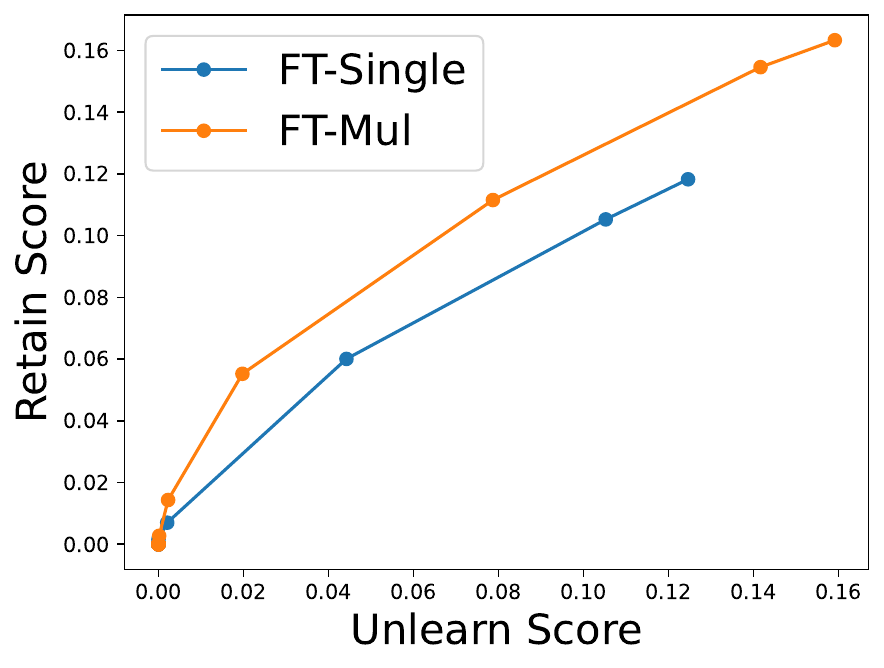}
    \caption{\texttt{UL-Single}, GA}\end{subfigure}
\hfill
\begin{subfigure}[t]{0.32\textwidth}
    \includegraphics[width=\linewidth]{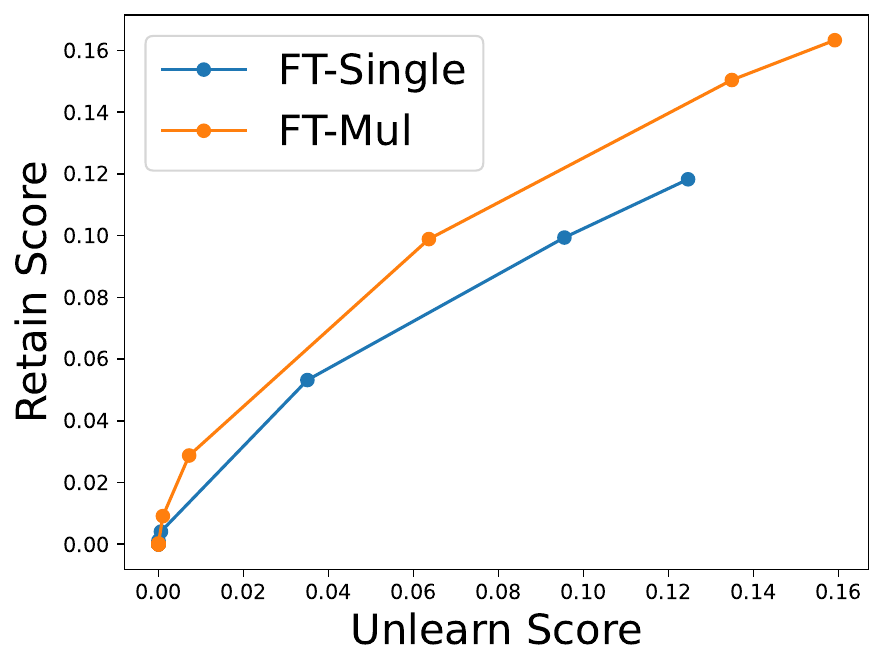}
    \caption{\texttt{UL-Mul}, GA}
\end{subfigure}

\begin{subfigure}[t]{0.32\textwidth}
    \includegraphics[width=\linewidth]{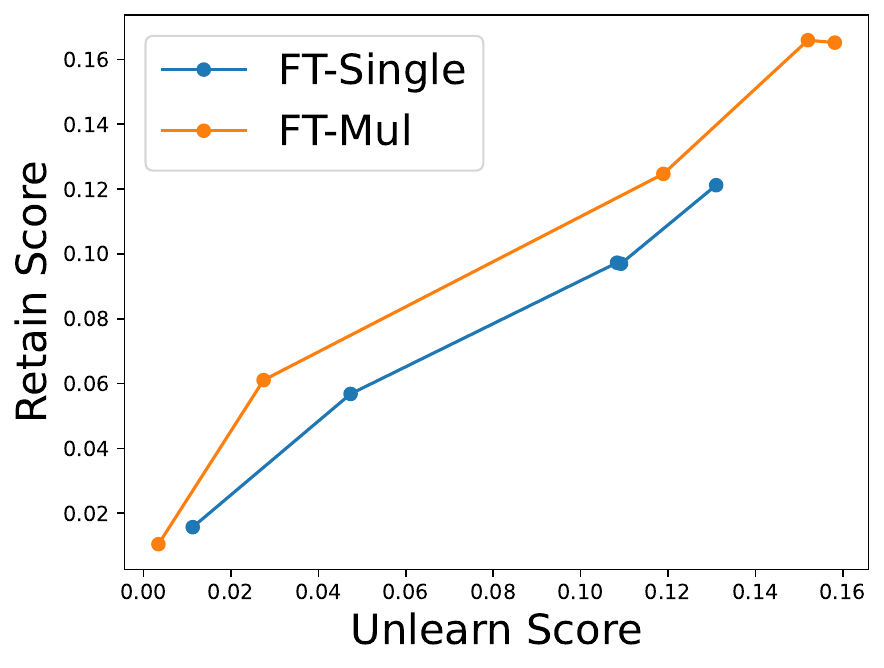}
    \caption{\texttt{UL-Exact}, TV}
\end{subfigure}
\hfill
\begin{subfigure}[t]{0.32\textwidth}
    \includegraphics[width=\linewidth]{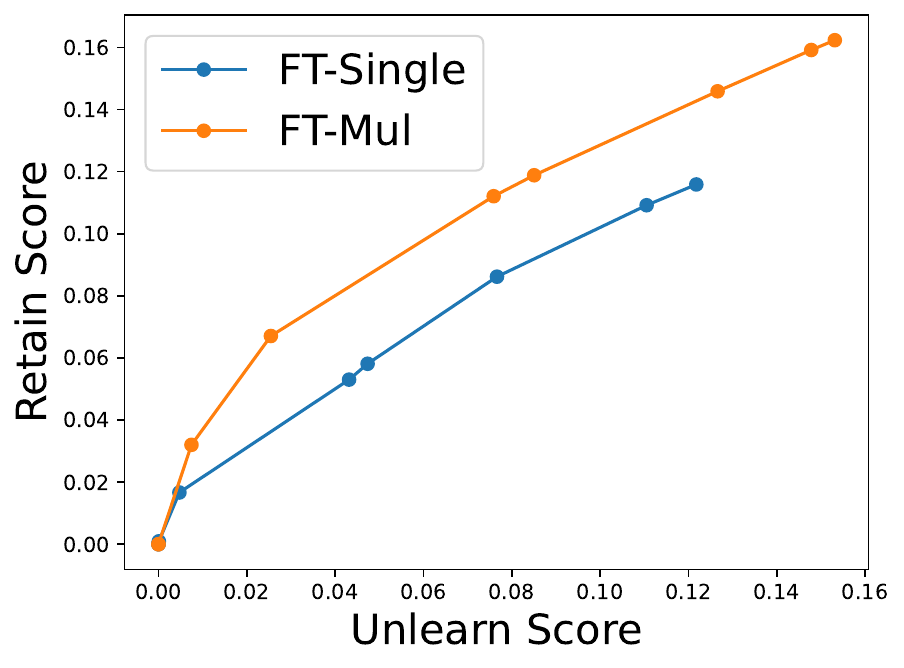}
    \caption{\texttt{UL-Single}, TV}\end{subfigure}
\hfill
\begin{subfigure}[t]{0.32\textwidth}
    \includegraphics[width=\linewidth]{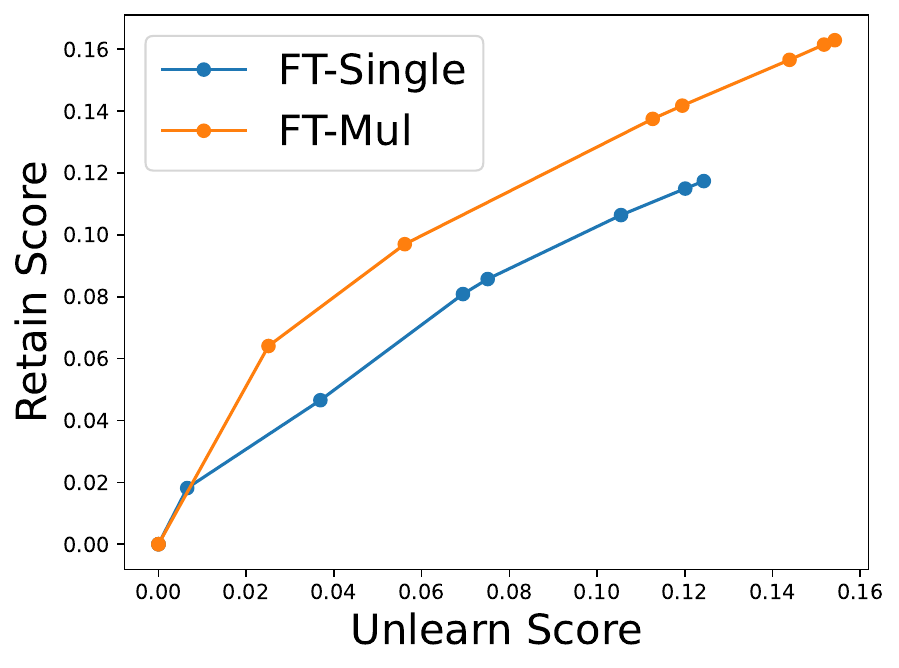}
    \caption{\texttt{UL-Mul}, TV}
\end{subfigure}

\caption{Vanilla \textbf{extraction} trade-off curves for three choices of unlearning data and two unlearning algorithms on \textbf{TOFU+} and \textbf{Llama2-7B}, when comparing \texttt{FT-Mul} and \texttt{FT-Single}}
\label{fig:extraction_curves_tofu_llama2-7b}
\end{figure}

\begin{figure}[!t]
\centering
\begin{subfigure}[t]{0.32\textwidth}
    \includegraphics[width=\linewidth]{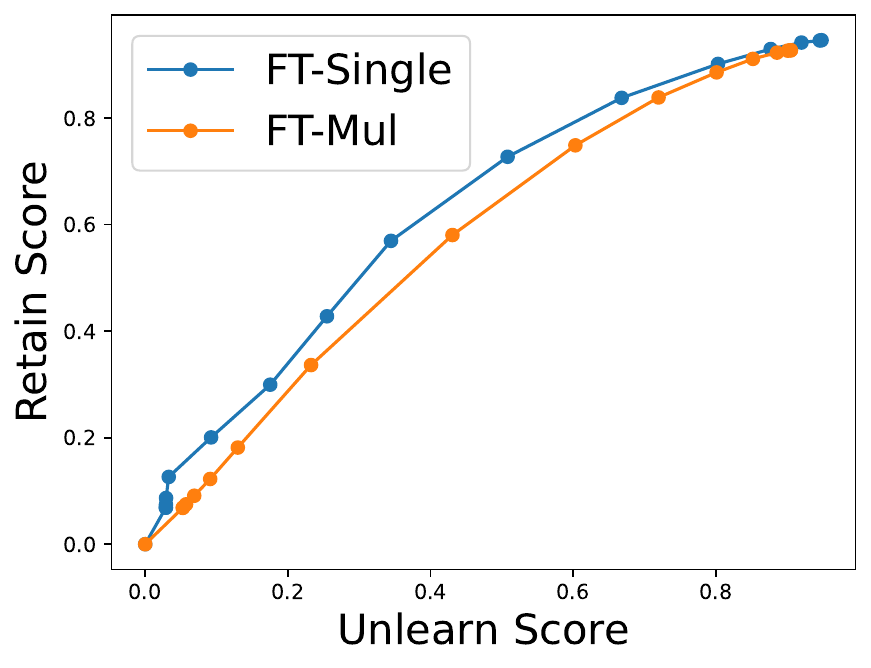}
    \caption{\texttt{UL-Exact}, GA}
\end{subfigure}
\hfill
\begin{subfigure}[t]{0.32\textwidth}
    \includegraphics[width=\linewidth]{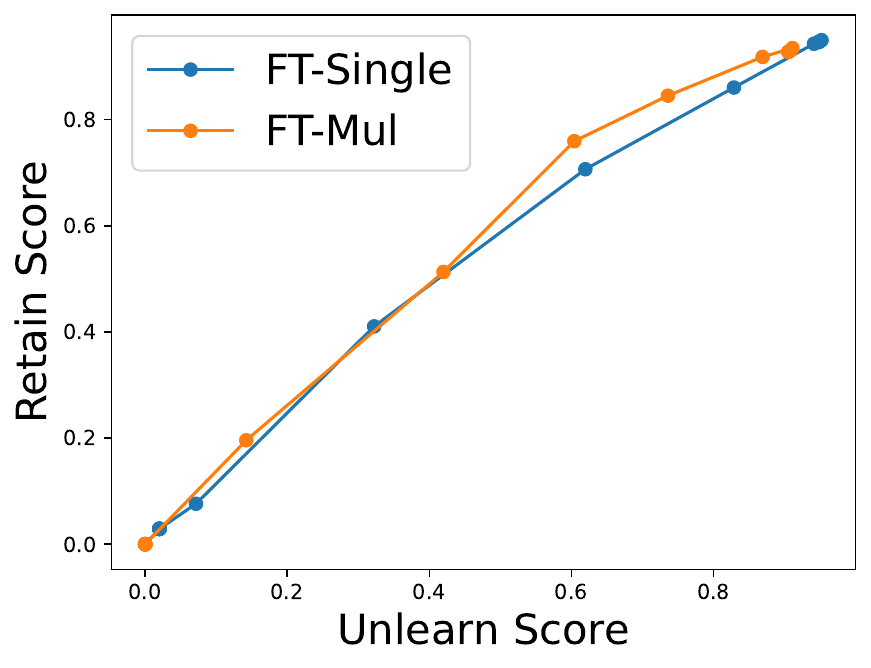}
    \caption{\texttt{UL-Single}, GA}\end{subfigure}
\hfill
\begin{subfigure}[t]{0.32\textwidth}
    \includegraphics[width=\linewidth]{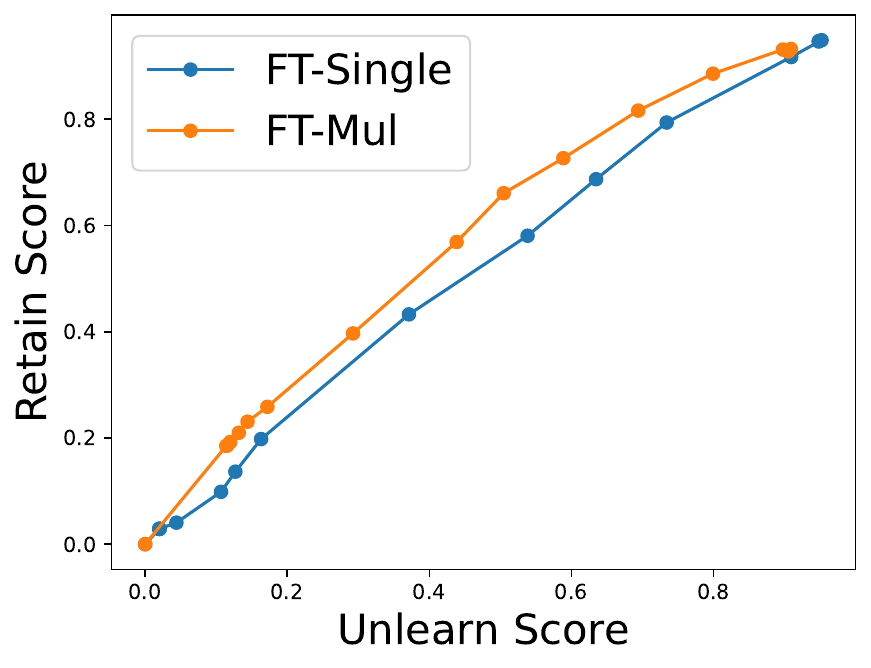}
    \caption{\texttt{UL-Mul}, GA}
\end{subfigure}

\begin{subfigure}[t]{0.32\textwidth}
    \includegraphics[width=\linewidth]{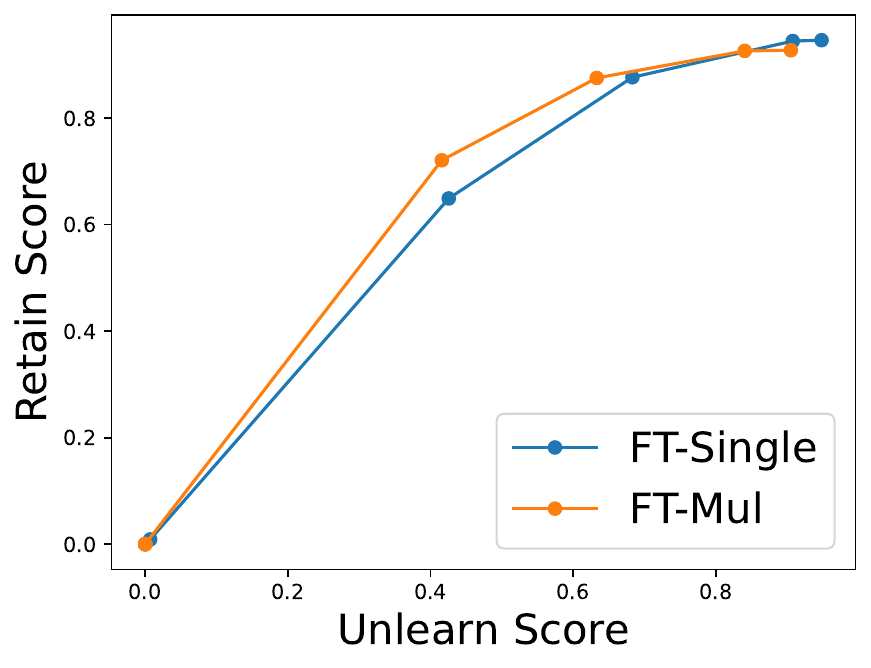}
    \caption{\texttt{UL-Exact}, TV}
\end{subfigure}
\hfill
\begin{subfigure}[t]{0.32\textwidth}
    \includegraphics[width=\linewidth]{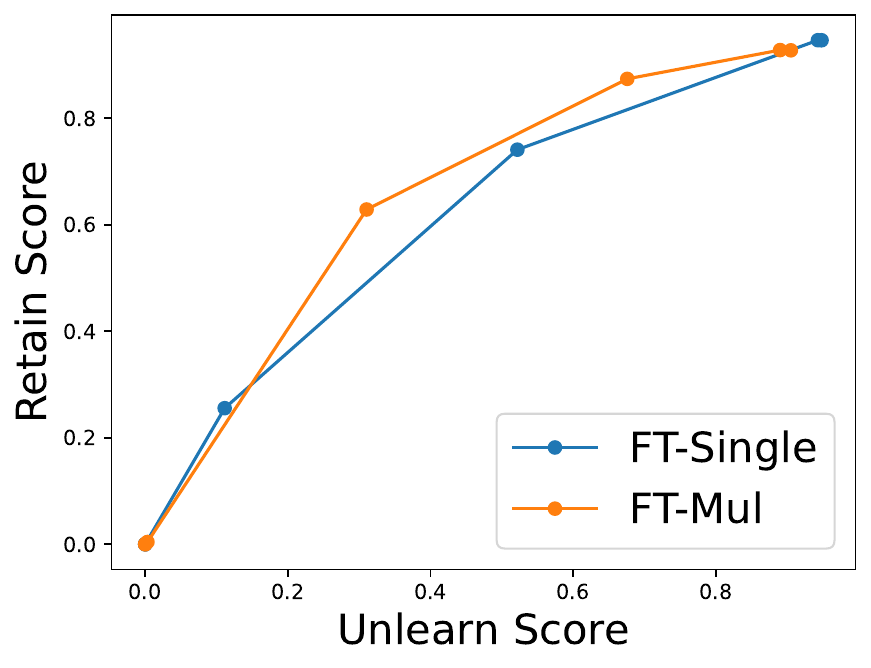}
    \caption{\texttt{UL-Single}, TV}\end{subfigure}
\hfill
\begin{subfigure}[t]{0.32\textwidth}
    \includegraphics[width=\linewidth]{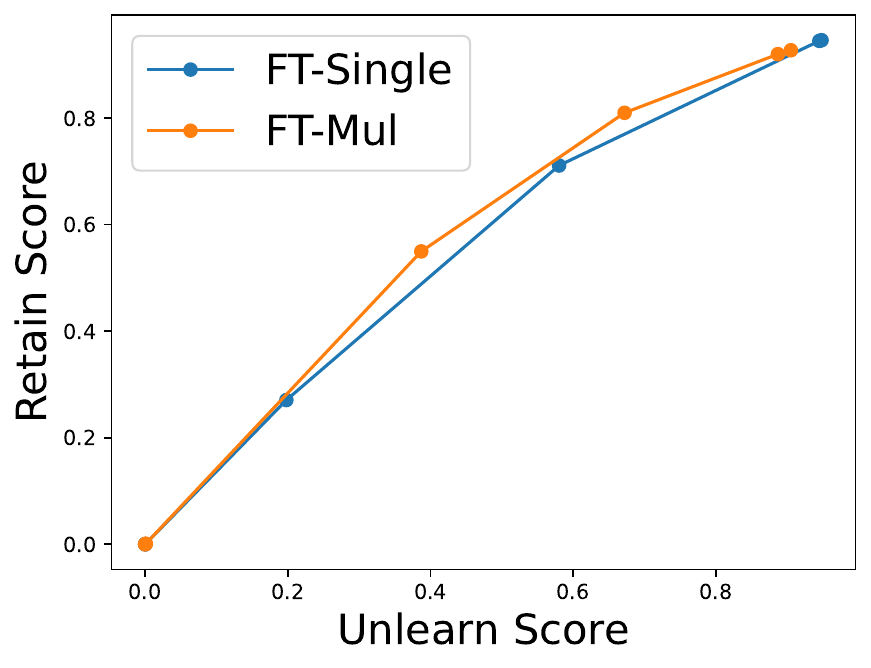}
    \caption{\texttt{UL-Mul}, TV}
\end{subfigure}

\caption{Vanilla \textbf{memorization} trade-off curves for three choices of unlearning data and two unlearning algorithms on \textbf{Eval-DU+} and \textbf{Llama2-7B}, when comparing \texttt{FT-Mul} and \texttt{FT-Single}}
\label{fig:memorization_curves_eval_du_llama2-7b}
\end{figure}

\begin{figure}[!t]
\centering
\begin{subfigure}[t]{0.32\textwidth}
    \includegraphics[width=\linewidth]{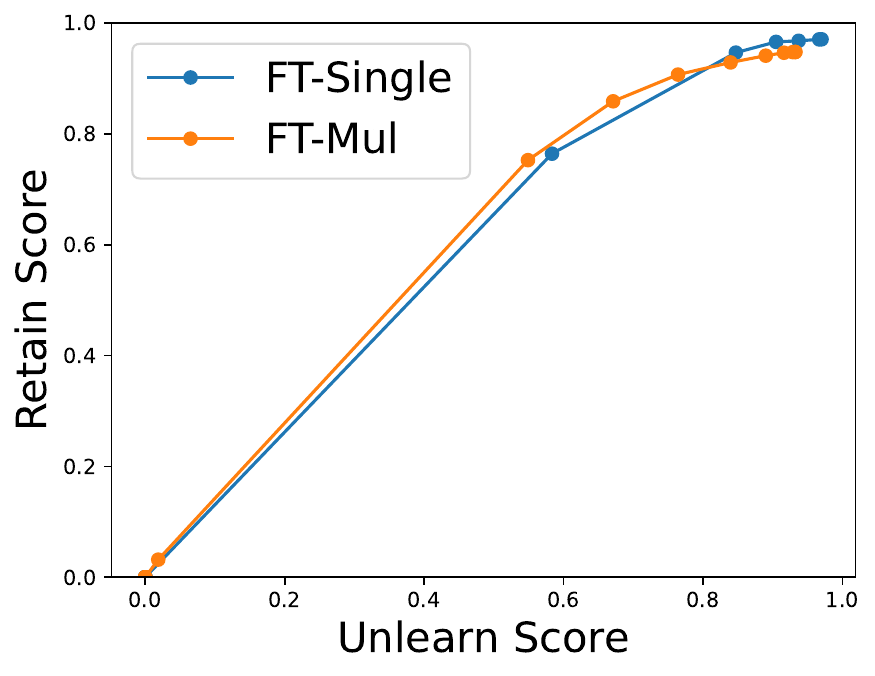}
    \caption{\texttt{UL-Exact}, GA}
\end{subfigure}
\hfill
\begin{subfigure}[t]{0.32\textwidth}
    \includegraphics[width=\linewidth]{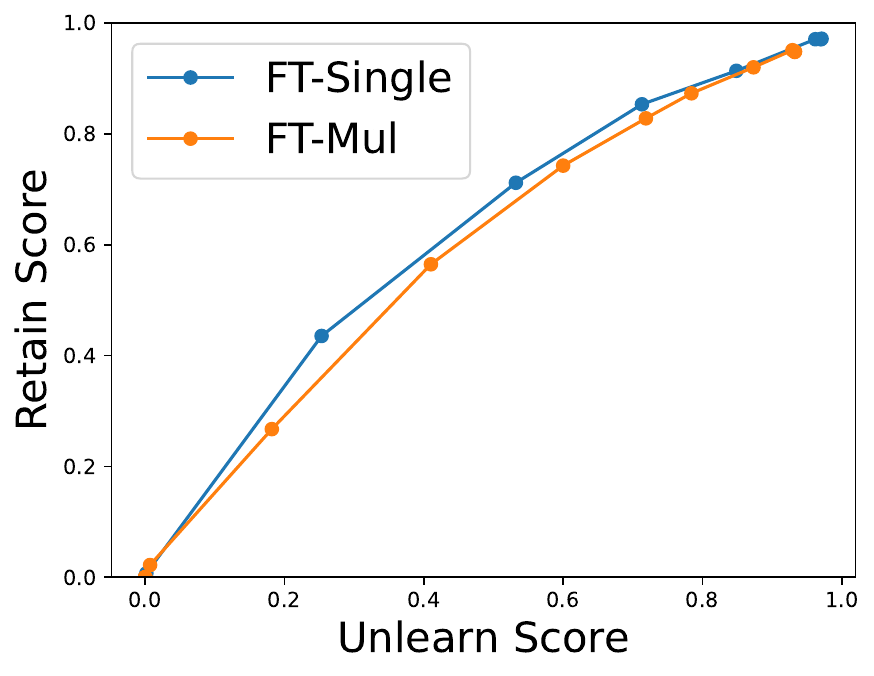}
    \caption{\texttt{UL-Single}, GA}\end{subfigure}
\hfill
\begin{subfigure}[t]{0.32\textwidth}
    \includegraphics[width=\linewidth]{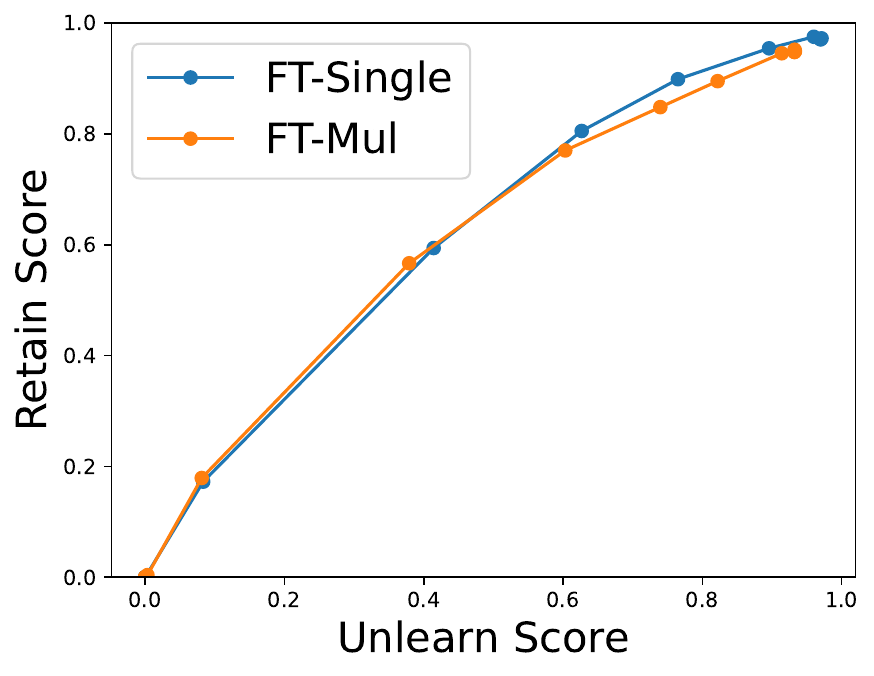}
    \caption{\texttt{UL-Mul}, GA}
\end{subfigure}

\begin{subfigure}[t]{0.32\textwidth}
    \includegraphics[width=\linewidth]{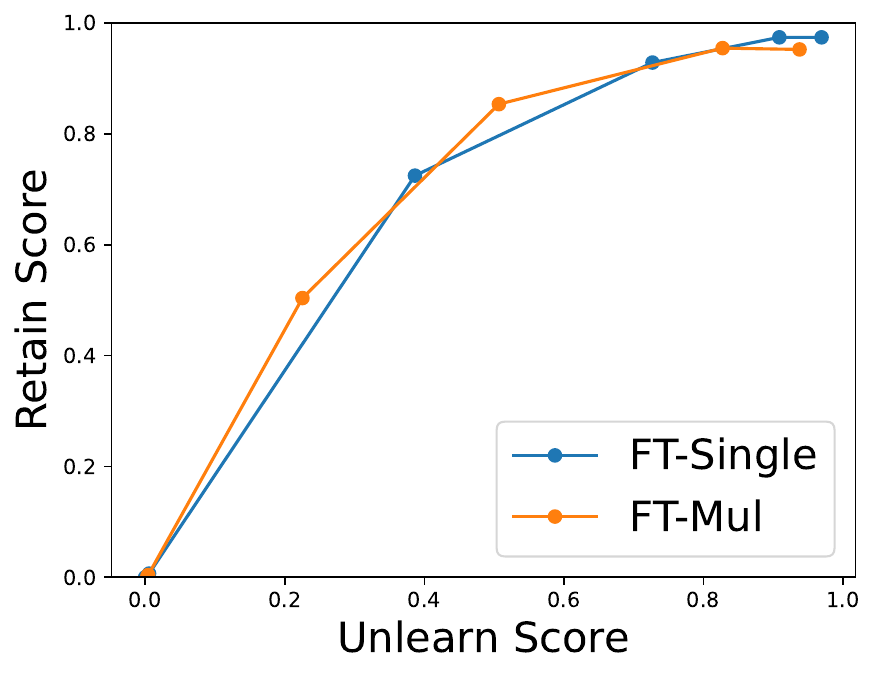}
    \caption{\texttt{UL-Exact}, TV}
\end{subfigure}
\hfill
\begin{subfigure}[t]{0.32\textwidth}
    \includegraphics[width=\linewidth]{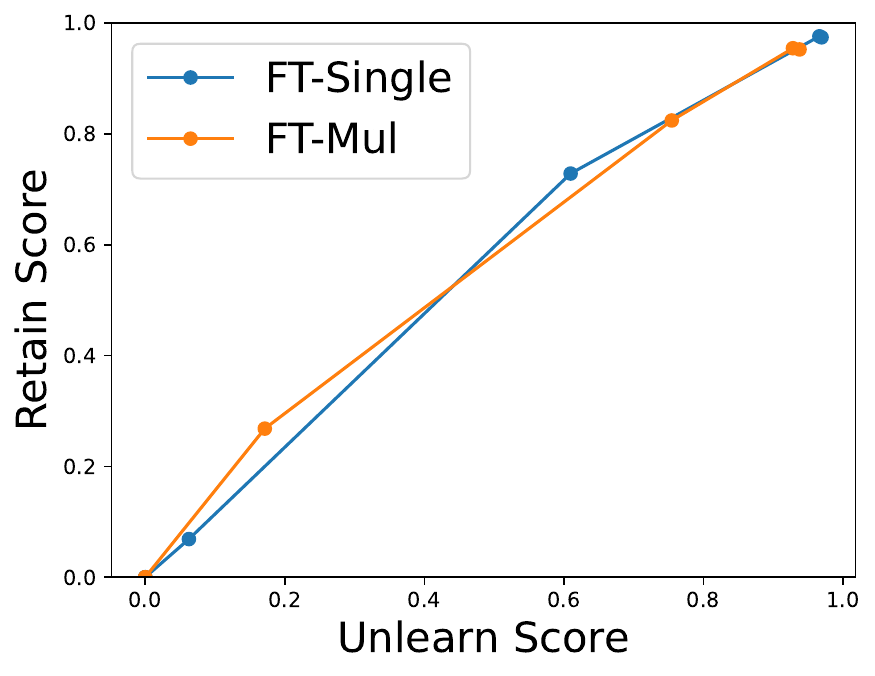}
    \caption{\texttt{UL-Single}, TV}\end{subfigure}
\hfill
\begin{subfigure}[t]{0.32\textwidth}
    \includegraphics[width=\linewidth]{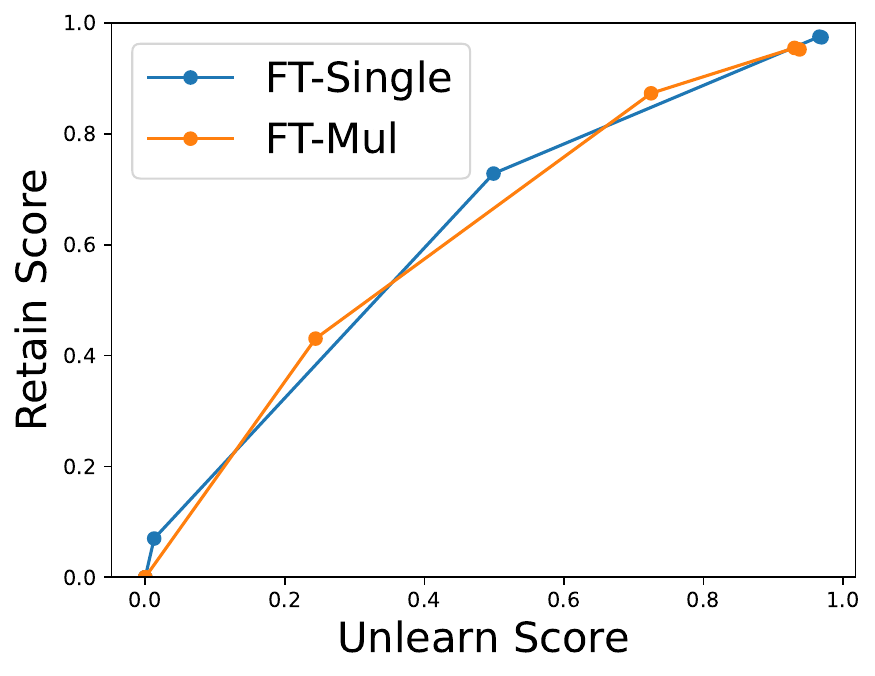}
    \caption{\texttt{UL-Mul}, TV}
\end{subfigure}

\caption{Vanilla \textbf{memorization} trade-off curves for three choices of unlearning data and two unlearning algorithms on \textbf{Eval-DU+} and \textbf{Llama3-8B}, when comparing \texttt{FT-Mul} and \texttt{FT-Single}.}
\label{fig:memorization_curves_eval_du_llama3-8b}
\end{figure}

\begin{figure}[!t]
\centering
\begin{subfigure}[t]{0.32\textwidth}
    \includegraphics[width=\linewidth]{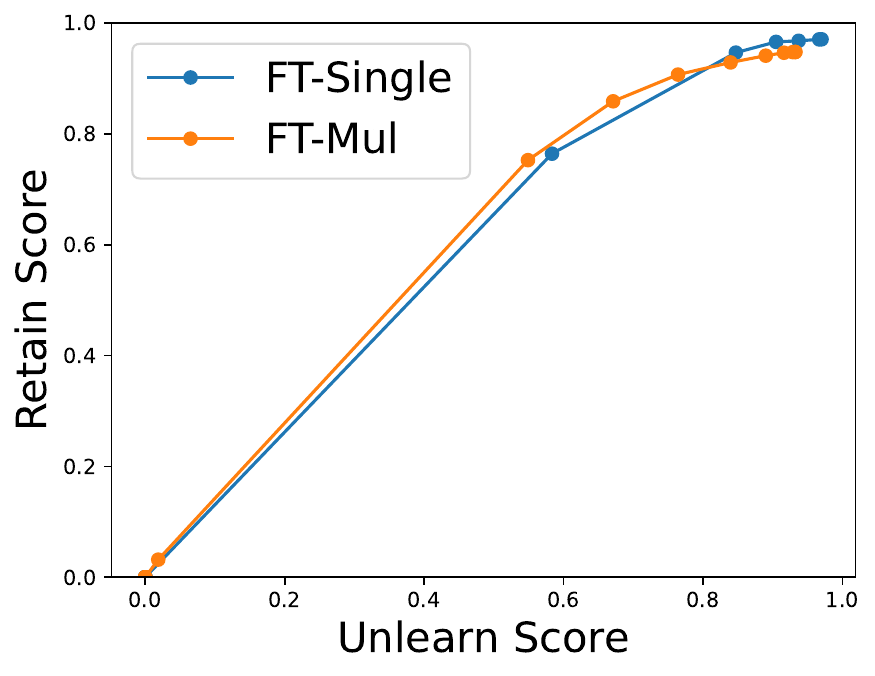}
    \caption{\texttt{UL-Exact}, GA}
\end{subfigure}
\hfill
\begin{subfigure}[t]{0.32\textwidth}
    \includegraphics[width=\linewidth]{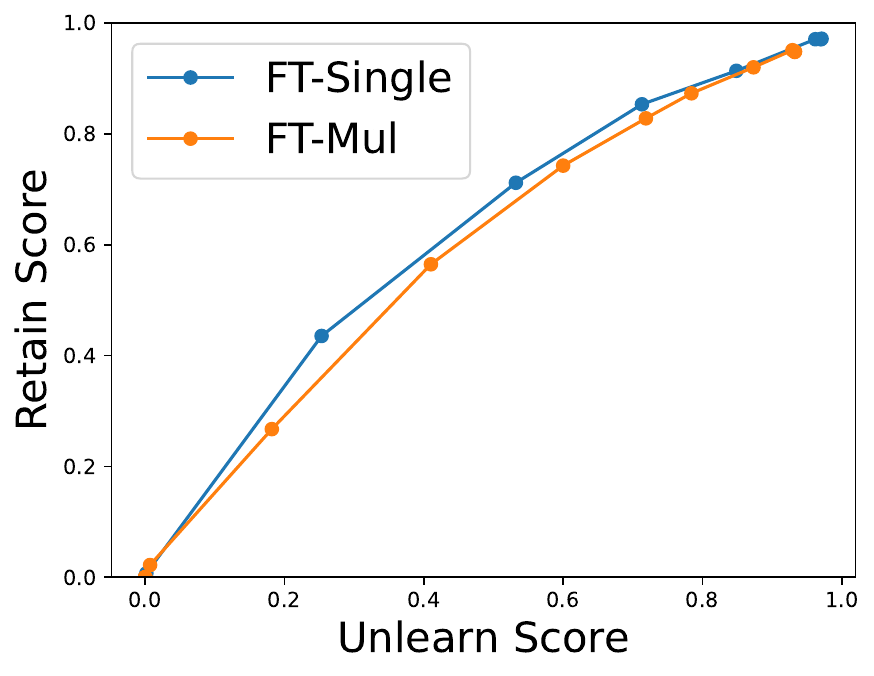}
    \caption{\texttt{UL-Single}, GA}\end{subfigure}
\hfill
\begin{subfigure}[t]{0.32\textwidth}
    \includegraphics[width=\linewidth]{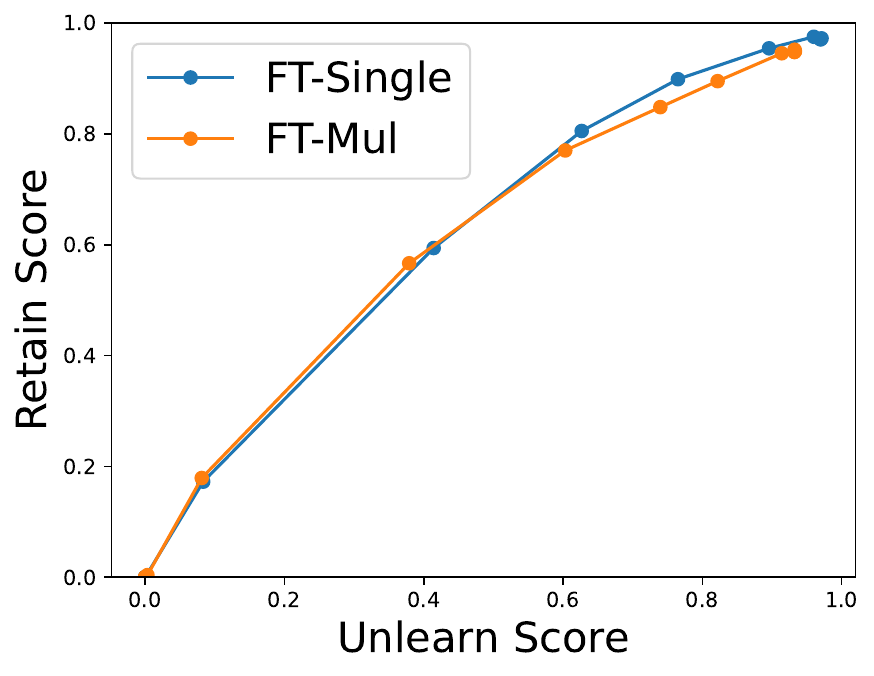}
    \caption{\texttt{UL-Mul}, GA}
\end{subfigure}

\begin{subfigure}[t]{0.32\textwidth}
    \includegraphics[width=\linewidth]{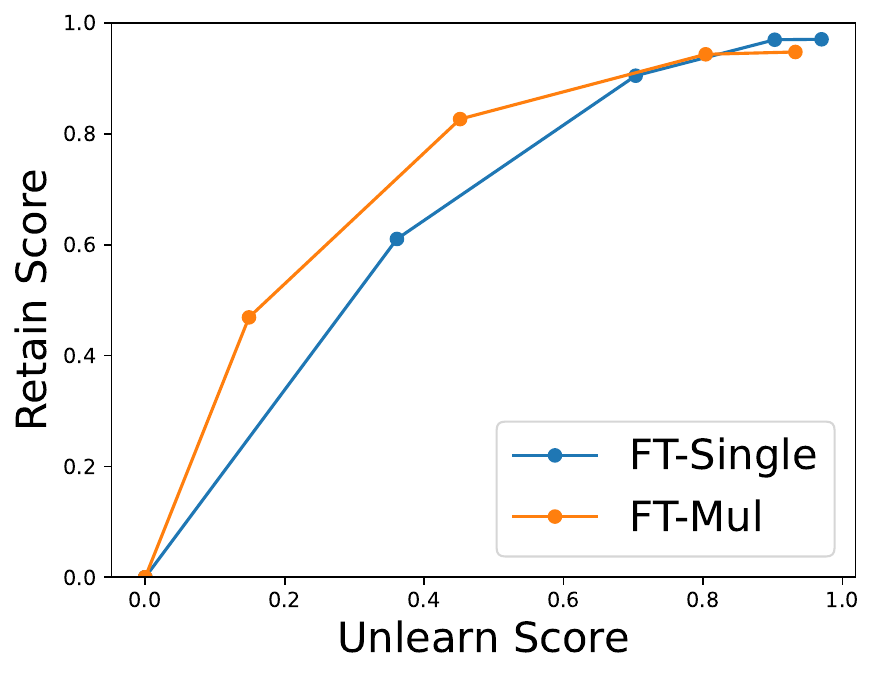}
    \caption{\texttt{UL-Exact}, TV}
\end{subfigure}
\hfill
\begin{subfigure}[t]{0.32\textwidth}
    \includegraphics[width=\linewidth]{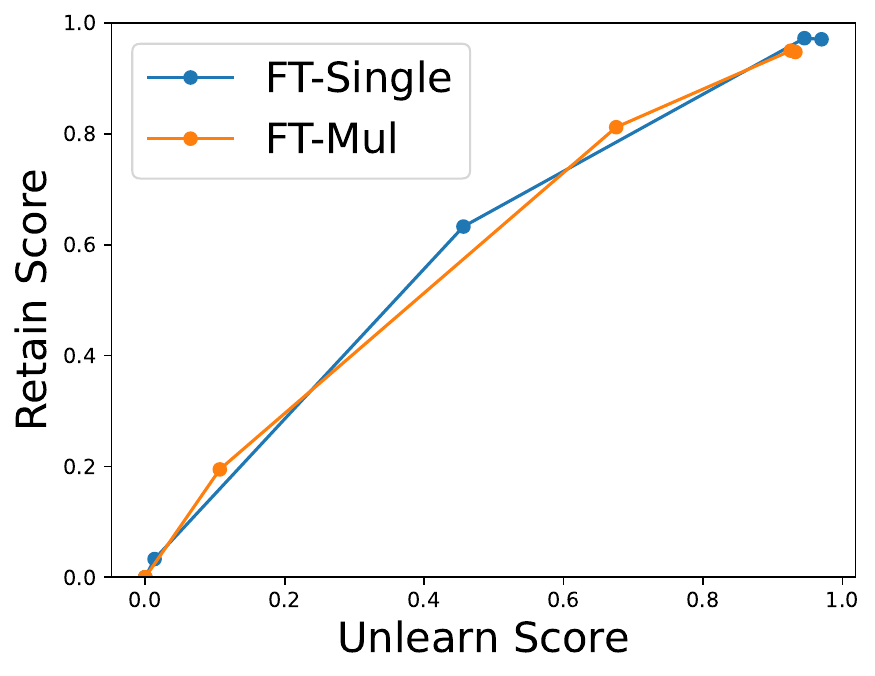}
    \caption{\texttt{UL-Single}, TV}\end{subfigure}
\hfill
\begin{subfigure}[t]{0.32\textwidth}
    \includegraphics[width=\linewidth]{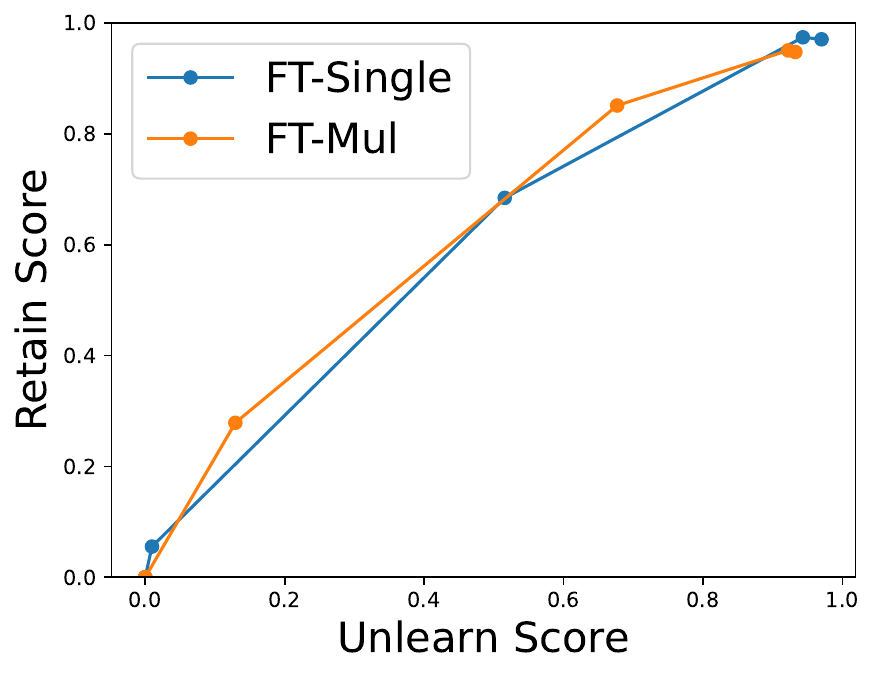}
    \caption{\texttt{UL-Mul}, TV}
\end{subfigure}

\caption{Vanilla \textbf{memorization} trade-off curves for three choices of unlearning data and two unlearning algorithms on \textbf{Eval-DU+} and \textbf{Gemma2-2b}, when comparing \texttt{FT-Mul} and \texttt{FT-Single}.}
\label{fig:memorization_curves_eval_du_gemma2-2b}
\end{figure}

\begin{figure}[!t]
\centering
\begin{subfigure}[t]{0.32\textwidth}
    \includegraphics[width=\linewidth]{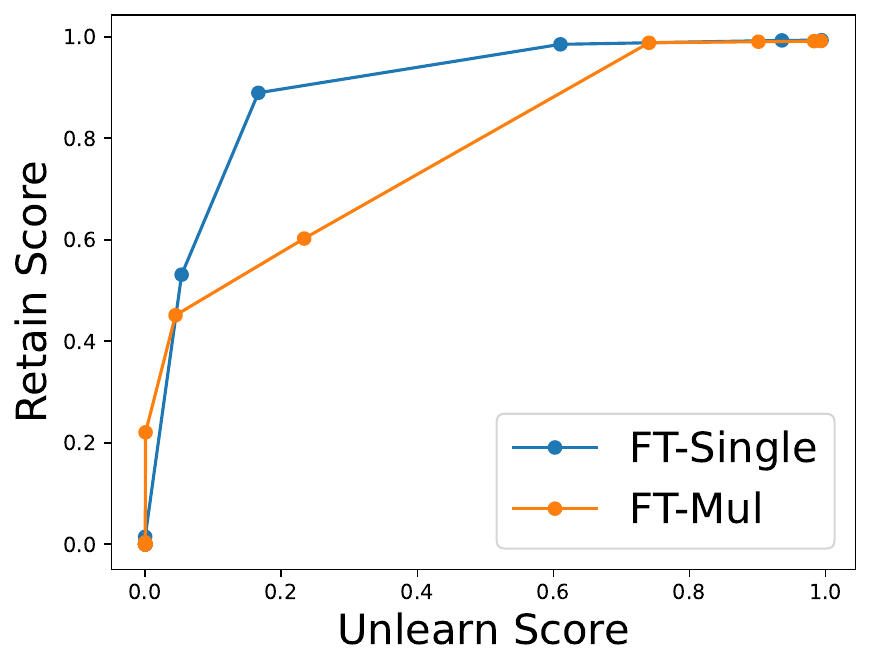}
    \caption{\texttt{UL-Exact}, GA}
\end{subfigure}
\hfill
\begin{subfigure}[t]{0.32\textwidth}
    \includegraphics[width=\linewidth]{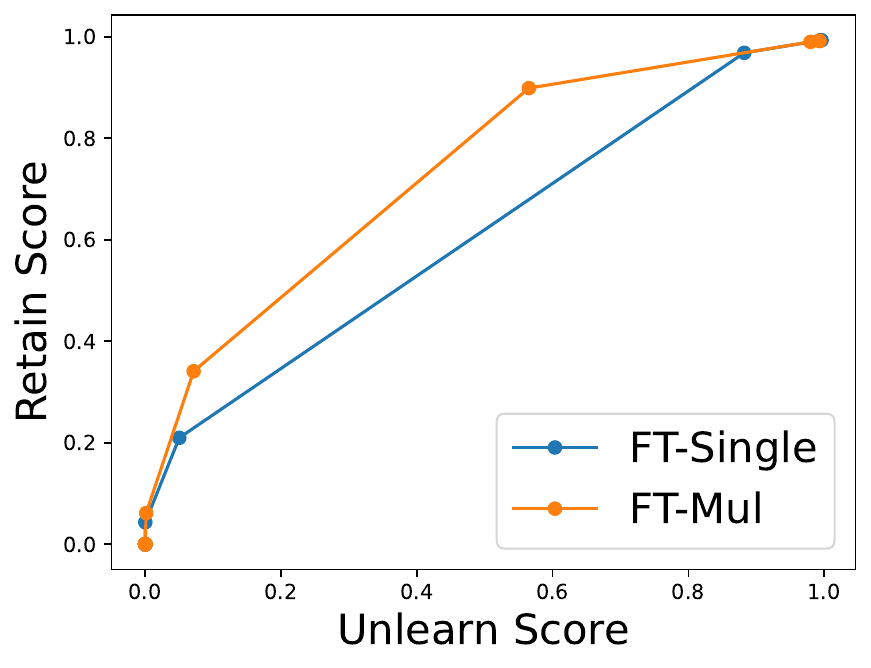}
    \caption{\texttt{UL-Single}, GA}\end{subfigure}
\hfill
\begin{subfigure}[t]{0.32\textwidth}
    \includegraphics[width=\linewidth]{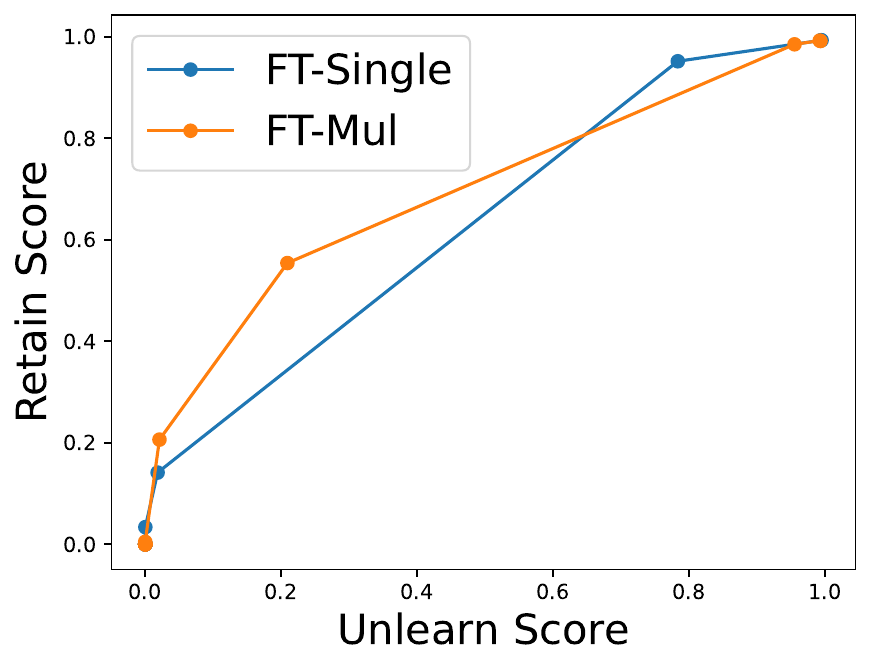}
    \caption{\texttt{UL-Mul}, GA}
\end{subfigure}

\begin{subfigure}[t]{0.32\textwidth}
    \includegraphics[width=\linewidth]{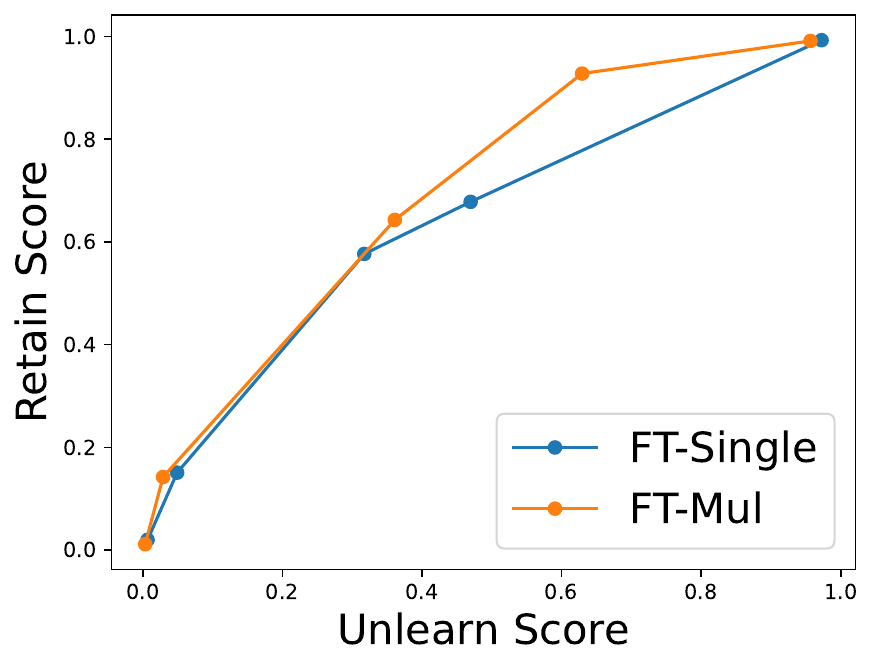}
    \caption{\texttt{UL-Exact}, TV}
\end{subfigure}
\hfill
\begin{subfigure}[t]{0.32\textwidth}
    \includegraphics[width=\linewidth]{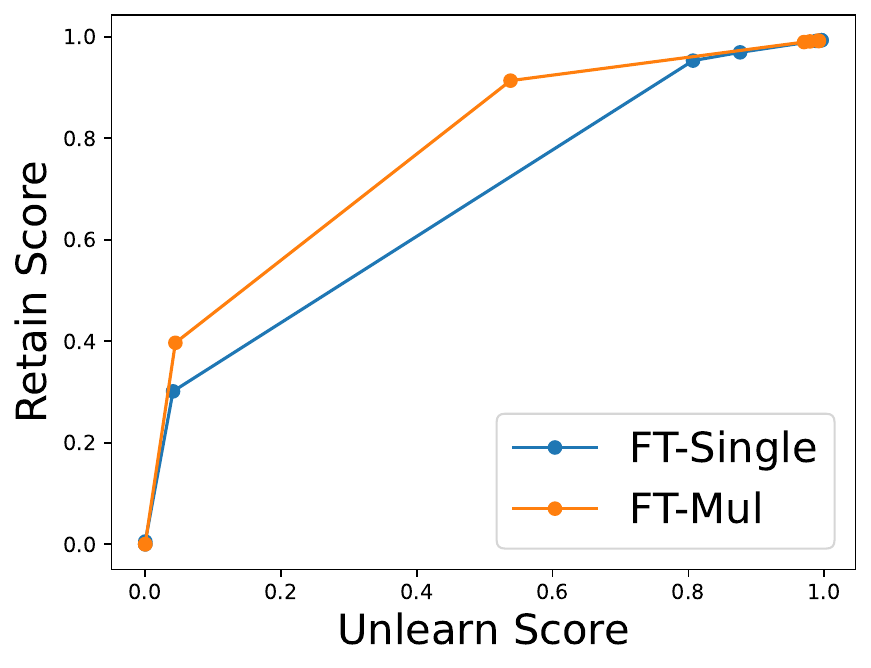}
    \caption{\texttt{UL-Single}, TV}\end{subfigure}
\hfill
\begin{subfigure}[t]{0.32\textwidth}
    \includegraphics[width=\linewidth]{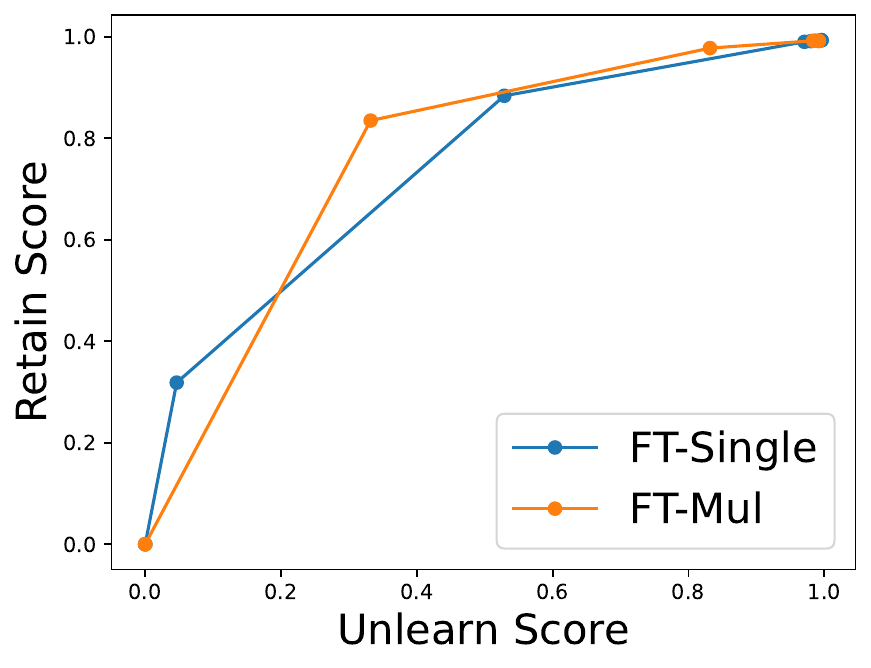}
    \caption{\texttt{UL-Mul}, TV}
\end{subfigure}

\caption{Vanilla \textbf{memorization} trade-off curves for three choices of unlearning data and two unlearning algorithms on \textbf{TOFU+} and \textbf{Llama2-7B}, when comparing \texttt{FT-Mul} and \texttt{FT-Single}.}
\label{fig:memorization_curves_tofu_llama2-7b}
\end{figure}

For the results in Section~\ref{sec:result_paraphrase}, the extraction trade-off plots for (Llama2-7B, Eval-DU+), and (Llama2-7B, TOFU+) are in Figure~\ref{fig:extraction_curves_eval_du_llama2-7b}, ~\ref{fig:extraction_curves_eval_du_llama3-8b}, ~\ref{fig:extraction_curves_eval_du_gemma2-2b},~\ref{fig:extraction_curves_tofu_llama2-7b} respectively; 
the memorization trade-off plots for (Llama2-7B, Eval-DU+), and (Llama2-7B, TOFU+) are in Figure~\ref{fig:memorization_curves_eval_du_llama2-7b},~\ref{fig:memorization_curves_eval_du_llama3-8b},~\ref{fig:memorization_curves_eval_du_gemma2-2b},~\ref{fig:memorization_curves_tofu_llama2-7b} respectively.

\begin{figure}[!t]
\centering
\begin{subfigure}[t]{0.32\textwidth}
    \includegraphics[width=\linewidth]{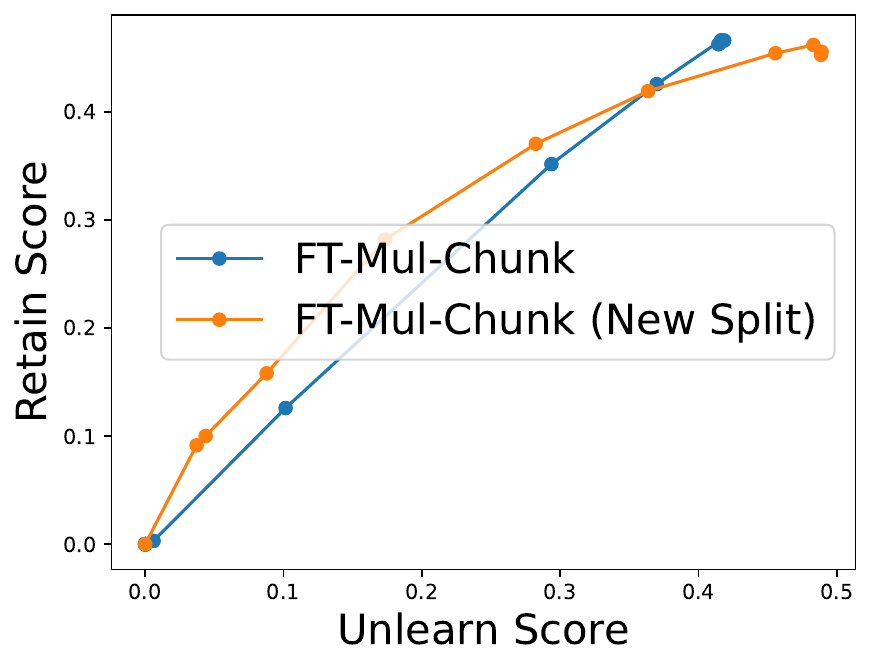}
    \caption{\texttt{UL-Exact}, GA}
\end{subfigure}
\hfill
\begin{subfigure}[t]{0.32\textwidth}
    \includegraphics[width=\linewidth]{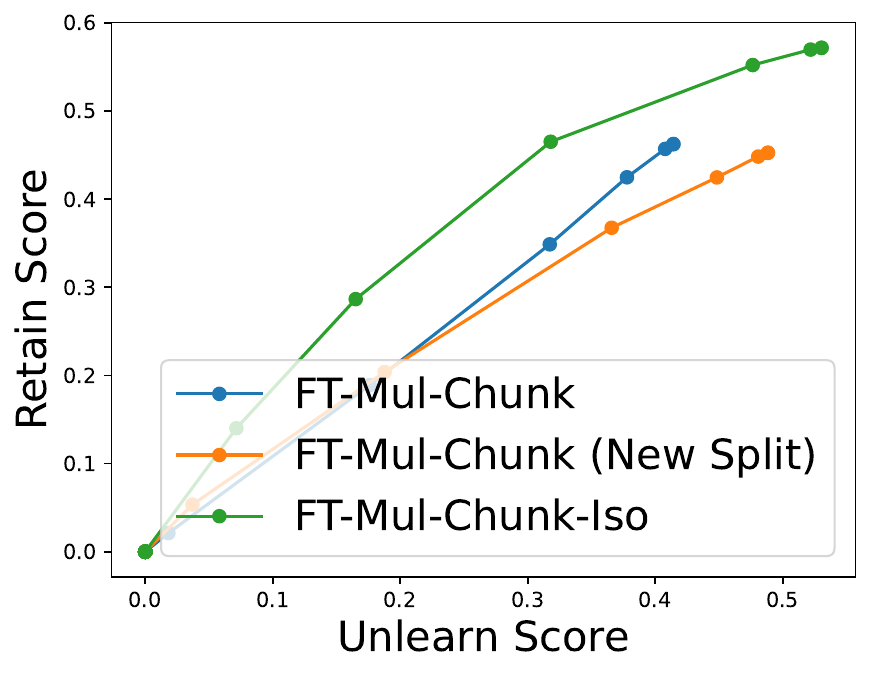}
    \caption{\texttt{UL-Single}, GA}\end{subfigure}
\hfill
\begin{subfigure}[t]{0.32\textwidth}
    \includegraphics[width=\linewidth]{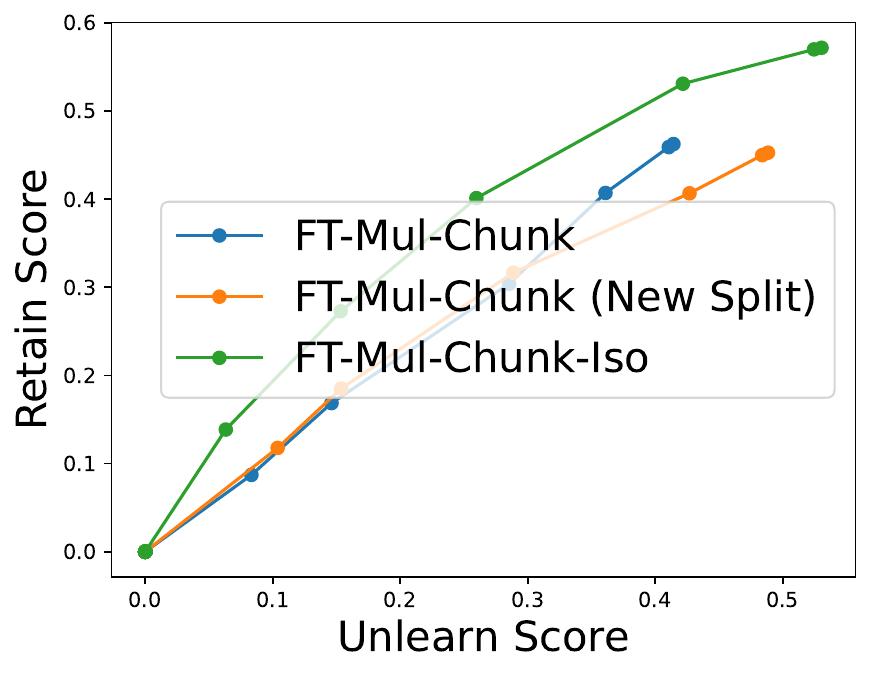}
    \caption{\texttt{UL-Mul}, GA}
\end{subfigure}

\begin{subfigure}[t]{0.32\textwidth}
    \includegraphics[width=\linewidth]{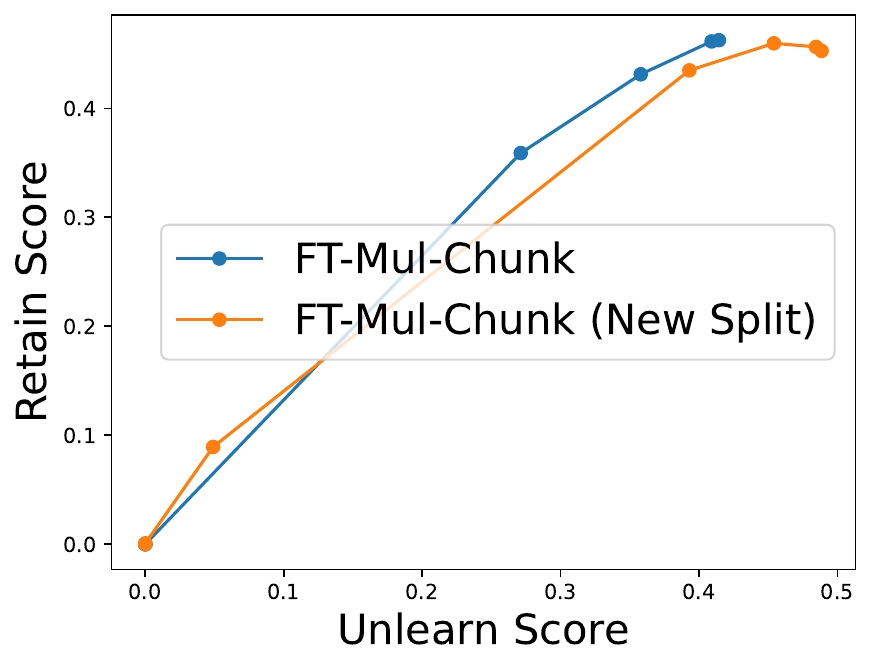}
    \caption{\texttt{UL-Exact}, TV}
\end{subfigure}
\hfill
\begin{subfigure}[t]{0.32\textwidth}
    \includegraphics[width=\linewidth]{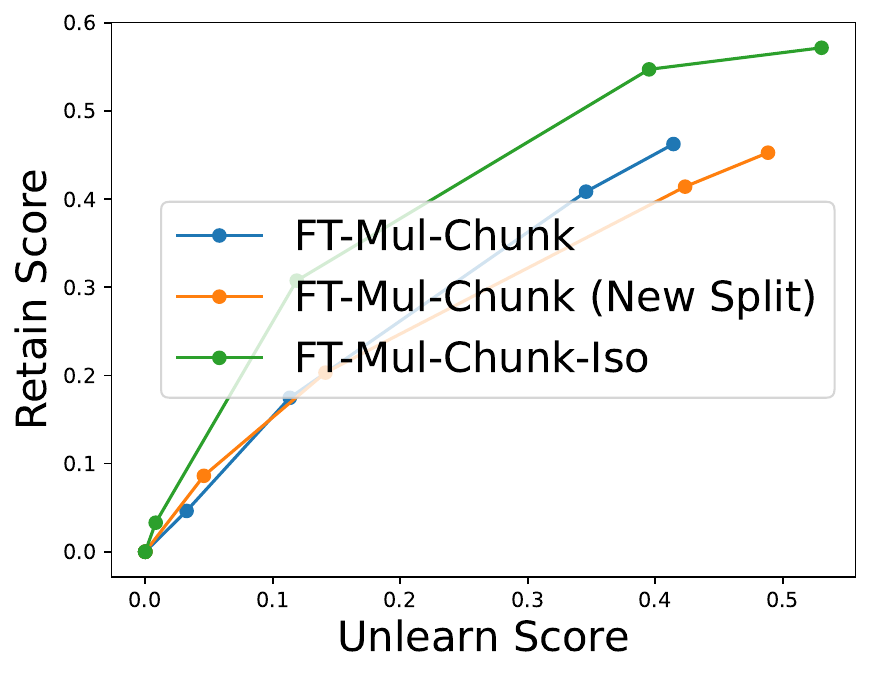}
    \caption{\texttt{UL-Single}, TV}\end{subfigure}
\hfill
\begin{subfigure}[t]{0.32\textwidth}
    \includegraphics[width=\linewidth]{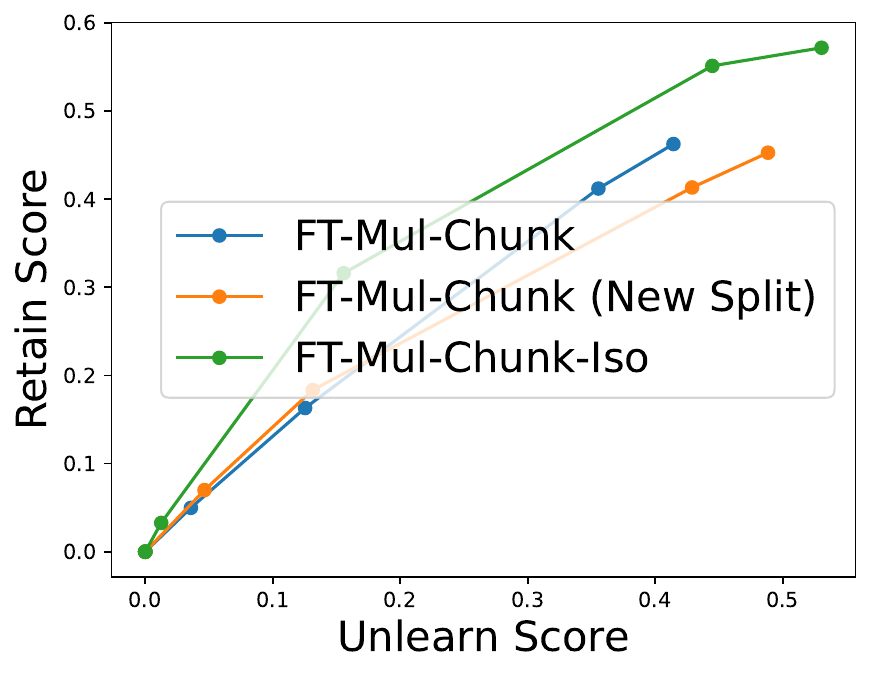}
    \caption{\texttt{UL-Mul}, TV}
\end{subfigure}
\caption{Vanilla \textbf{extraction} trade-off curves for three choices of unlearning data and two unlearning algorithms on \textbf{Eval-DU+} and \textbf{Llama2-7B}, when the model is fine-tuned from any text chunks.}
\label{fig:extraction_curves_eval_du_text_chunk_llama2-7b}
\end{figure}

\begin{figure}[!t]
\centering
\begin{subfigure}[t]{0.32\textwidth}
    \includegraphics[width=\linewidth]{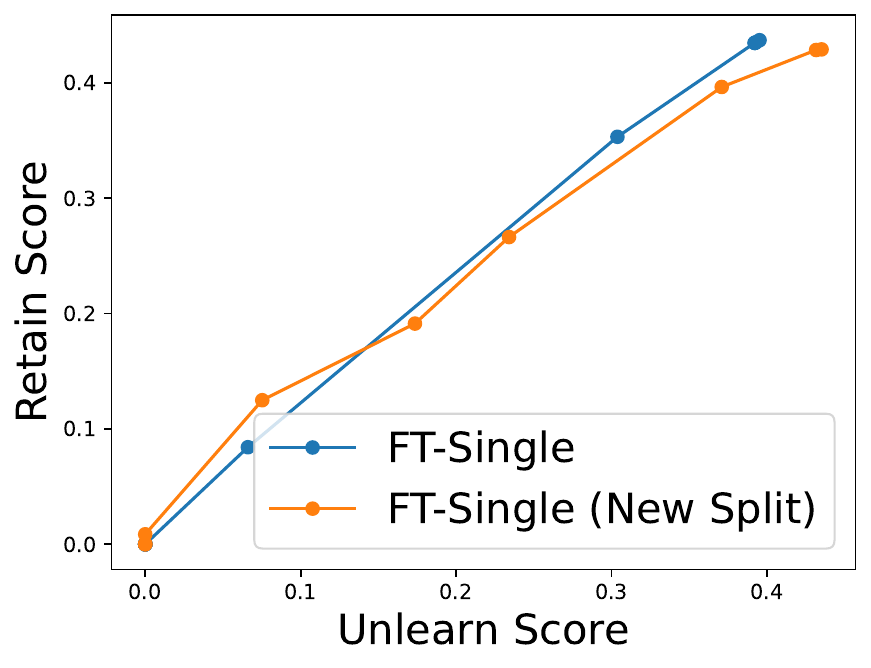}
    \caption{\texttt{UL-Exact}, GA}
\end{subfigure}
\hfill
\begin{subfigure}[t]{0.32\textwidth}
    \includegraphics[width=\linewidth]{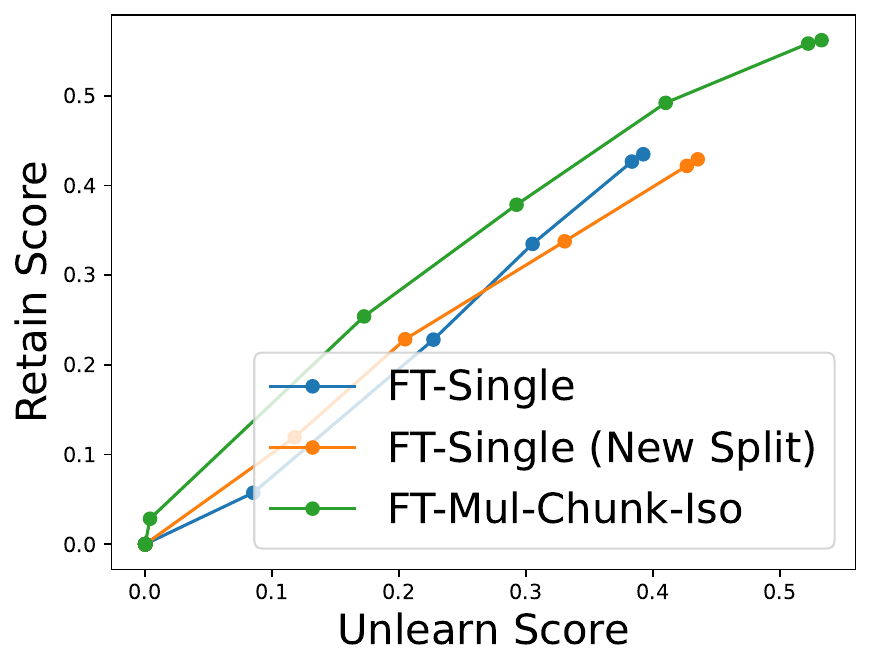}
    \caption{\texttt{UL-Single}, GA}\end{subfigure}
\hfill
\begin{subfigure}[t]{0.32\textwidth}
    \includegraphics[width=\linewidth]{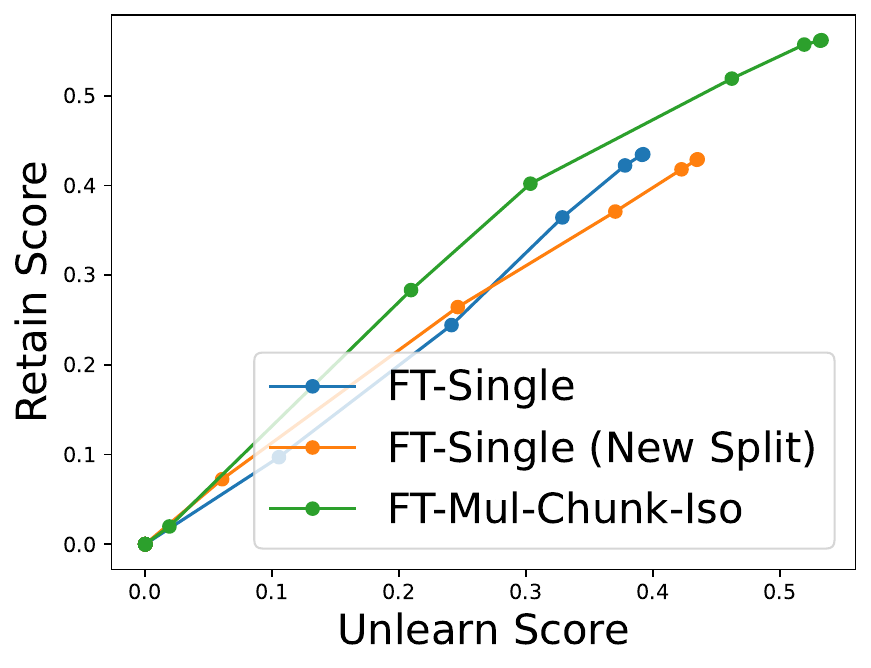}
    \caption{\texttt{UL-Mul}, GA}
\end{subfigure}

\begin{subfigure}[t]{0.32\textwidth}
    \includegraphics[width=\linewidth]{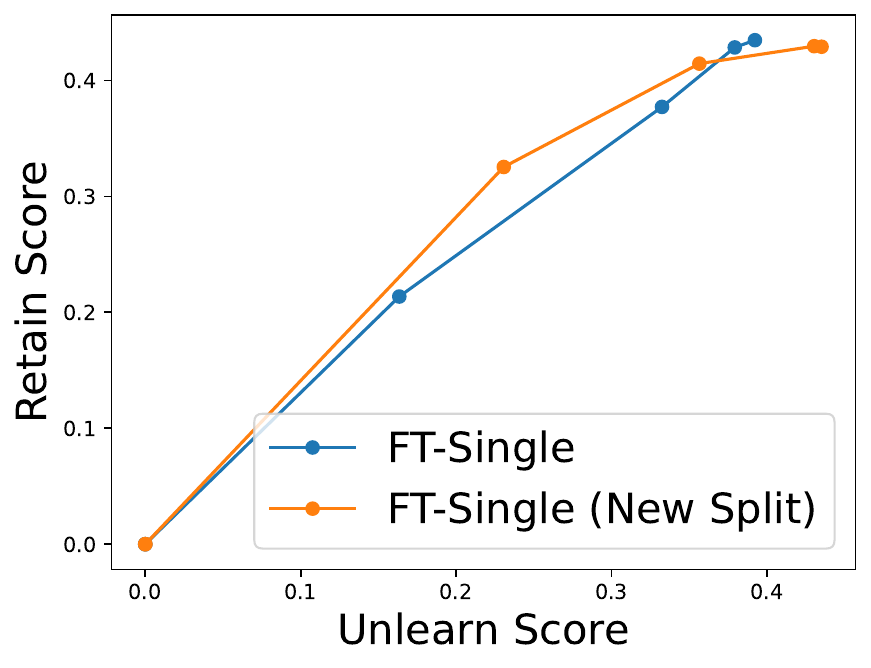}
    \caption{\texttt{UL-Exact}, TV}
\end{subfigure}
\hfill
\begin{subfigure}[t]{0.32\textwidth}
    \includegraphics[width=\linewidth]{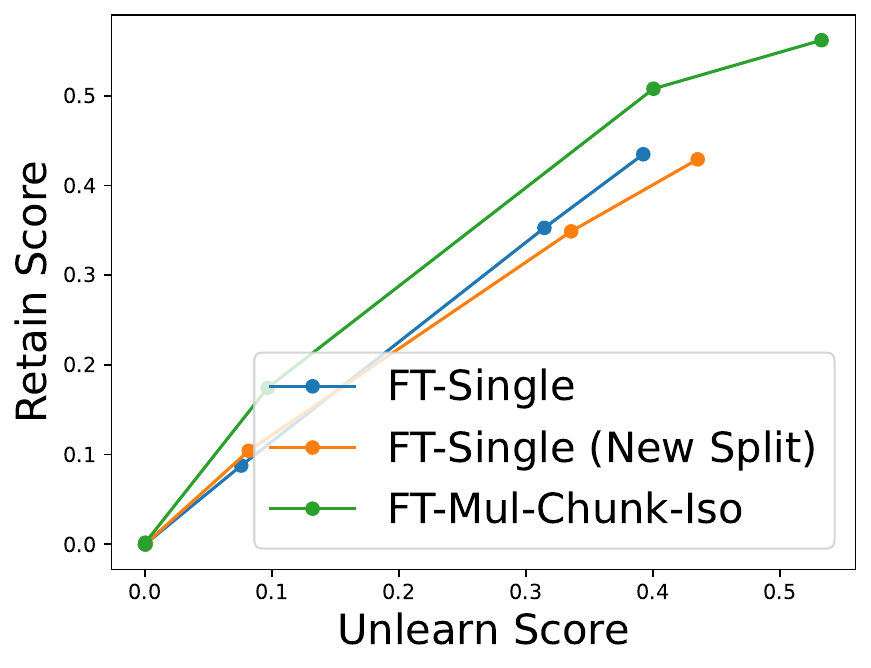}
    \caption{\texttt{UL-Single}, TV}\end{subfigure}
\hfill
\begin{subfigure}[t]{0.32\textwidth}
    \includegraphics[width=\linewidth]{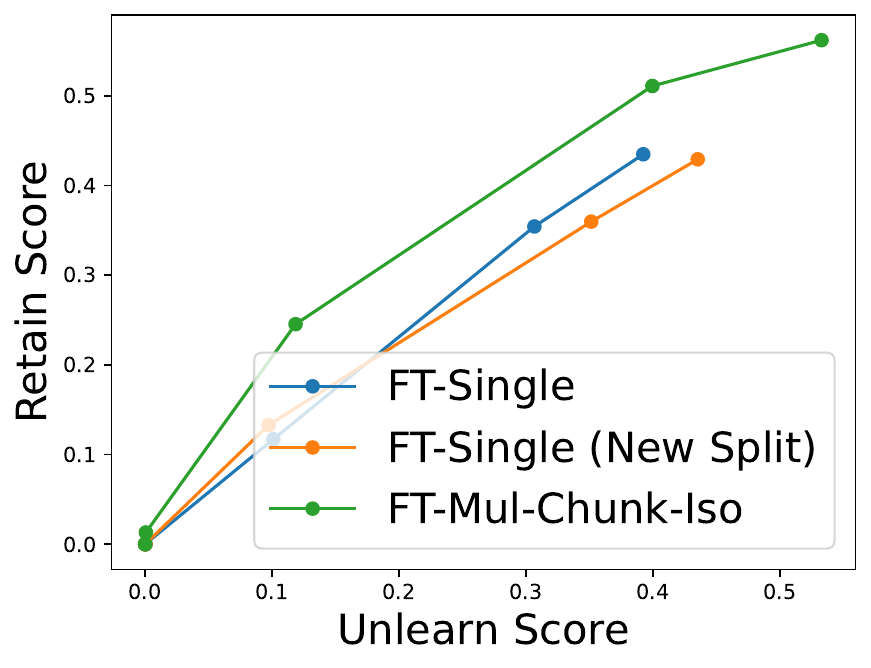}
    \caption{\texttt{UL-Mul}, TV}
\end{subfigure}
\caption{Vanilla \textbf{extraction} trade-off curves for three choices of unlearning data and two unlearning algorithms on \textbf{Eval-DU+} and \textbf{Llama3-8B}, when the model is fine-tuned from any text chunks.}
\label{fig:extraction_curves_eval_du_text_chunk_llama3-8b}
\end{figure}

\begin{figure}[!t]
\centering
\begin{subfigure}[t]{0.32\textwidth}
    \includegraphics[width=\linewidth]{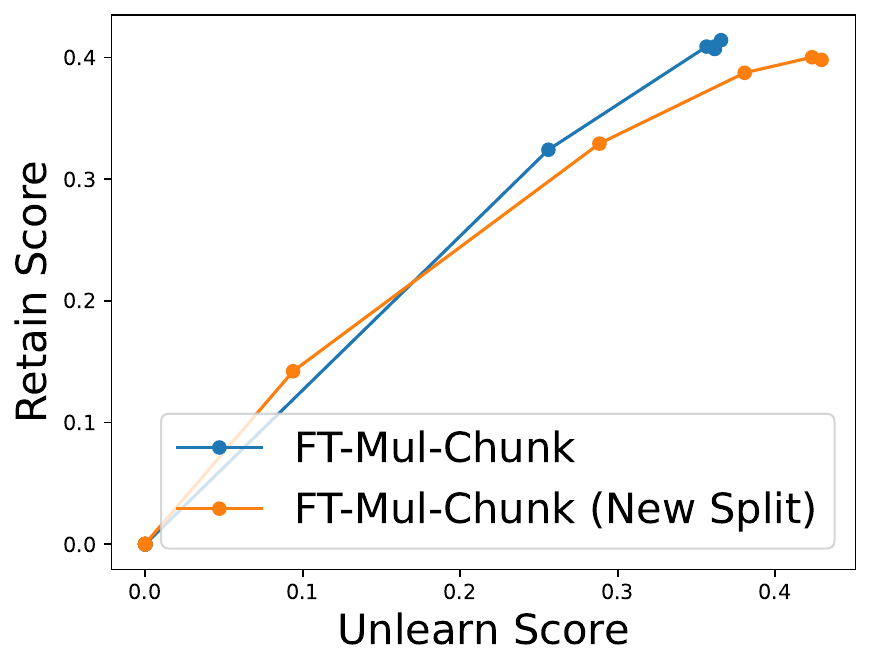}
    \caption{\texttt{UL-Exact}, GA}
\end{subfigure}
\hfill
\begin{subfigure}[t]{0.32\textwidth}
    \includegraphics[width=\linewidth]{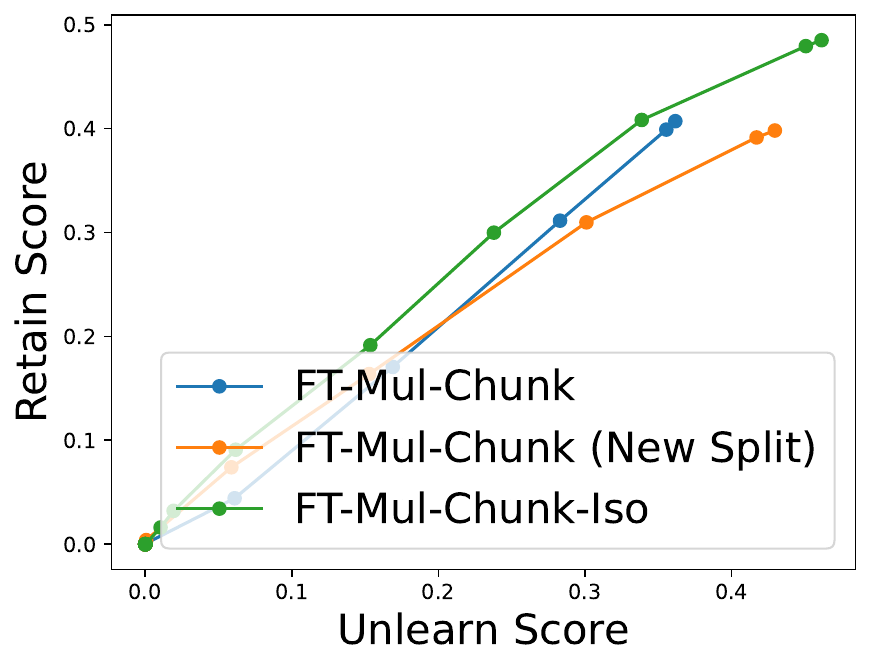}
    \caption{\texttt{UL-Single}, GA}\end{subfigure}
\hfill
\begin{subfigure}[t]{0.32\textwidth}
    \includegraphics[width=\linewidth]{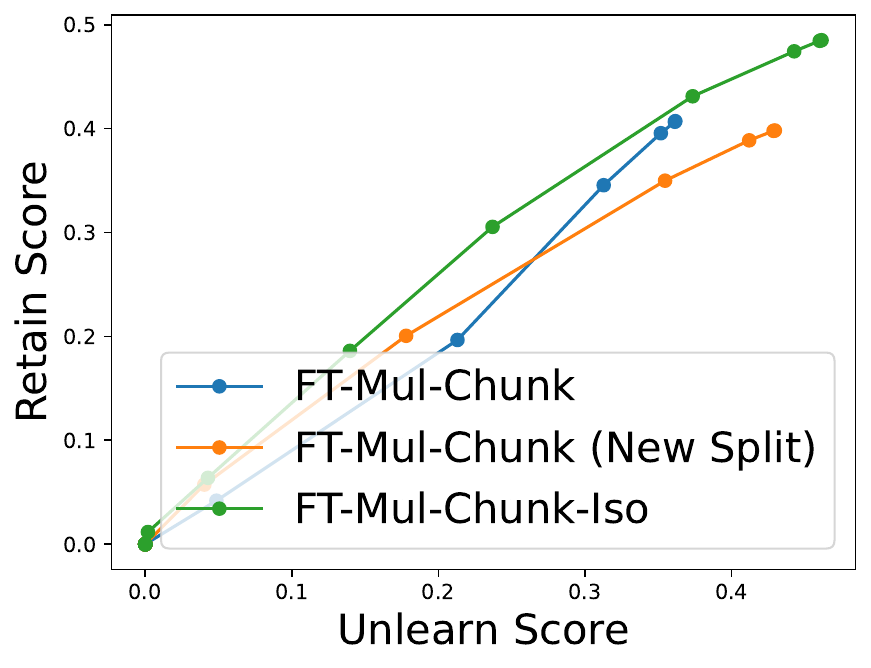}
    \caption{\texttt{UL-Mul}, GA}
\end{subfigure}

\begin{subfigure}[t]{0.32\textwidth}
    \includegraphics[width=\linewidth]{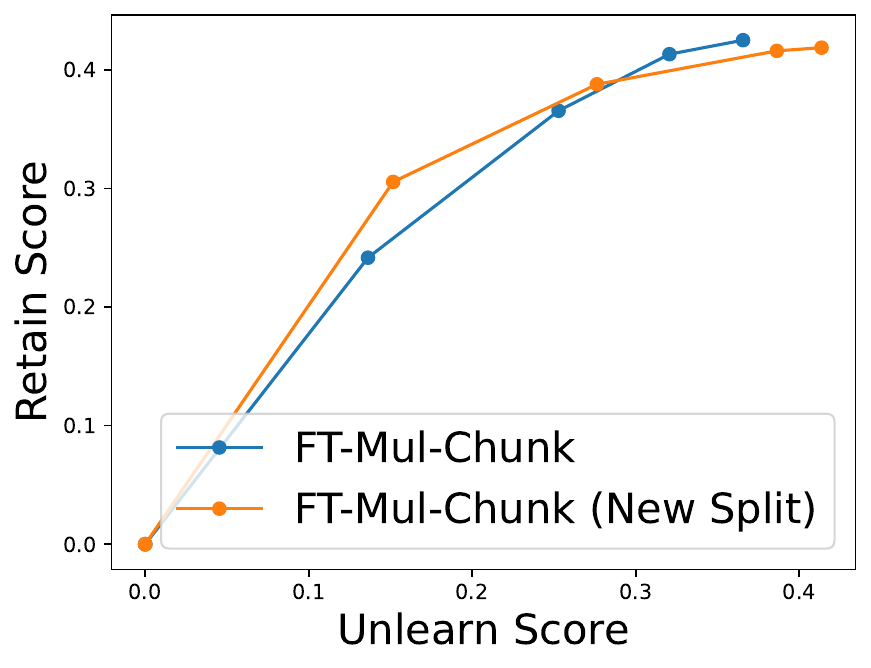}
    \caption{\texttt{UL-Exact}, TV}
\end{subfigure}
\hfill
\begin{subfigure}[t]{0.32\textwidth}
    \includegraphics[width=\linewidth]{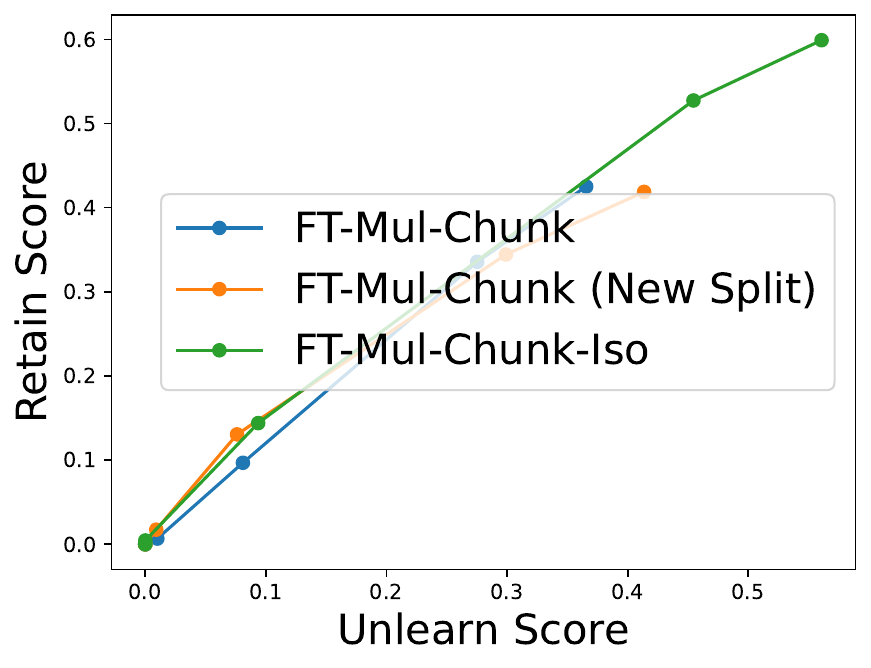}
    \caption{\texttt{UL-Single}, TV}\end{subfigure}
\hfill
\begin{subfigure}[t]{0.32\textwidth}
    \includegraphics[width=\linewidth]{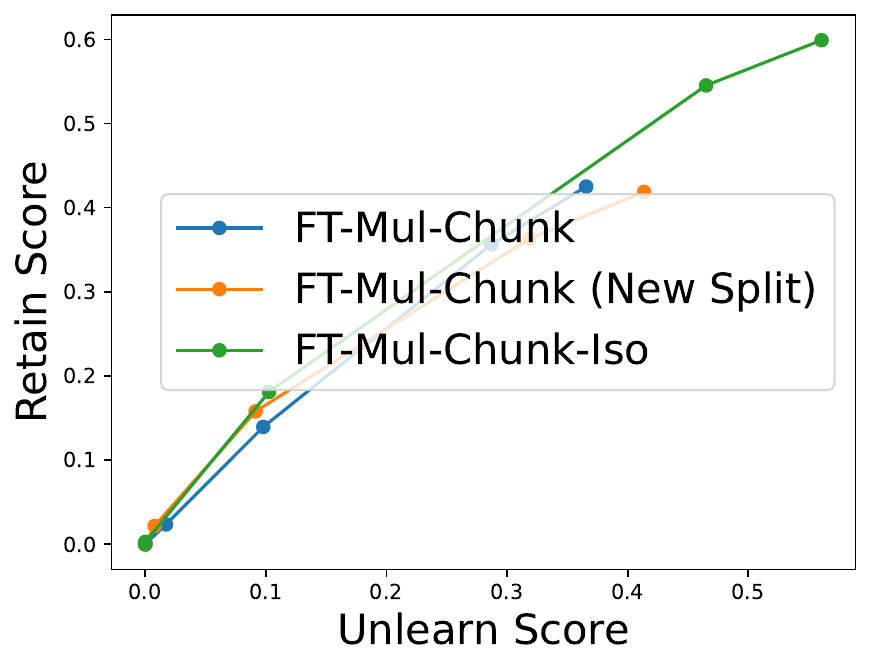}
    \caption{\texttt{UL-Mul}, TV}
\end{subfigure}
\caption{Vanilla \textbf{extraction} trade-off curves for three choices of unlearning data and two unlearning algorithms on \textbf{Eval-DU+} and \textbf{Gemma2-2b}, when the model is fine-tuned from any text chunks.}
\label{fig:extraction_curves_eval_du_text_chunk_gemma2-2b}
\end{figure}

\begin{figure}[!t]
\centering
\begin{subfigure}[t]{0.32\textwidth}
    \includegraphics[width=\linewidth]{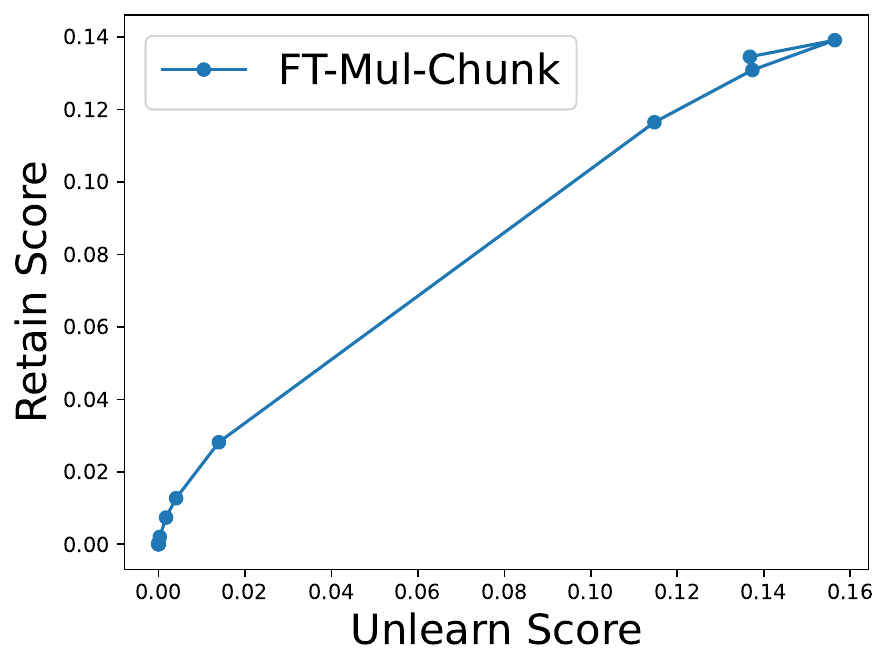}
    \caption{\texttt{UL-Exact}, GA}
\end{subfigure}
\hfill
\begin{subfigure}[t]{0.32\textwidth}
    \includegraphics[width=\linewidth]{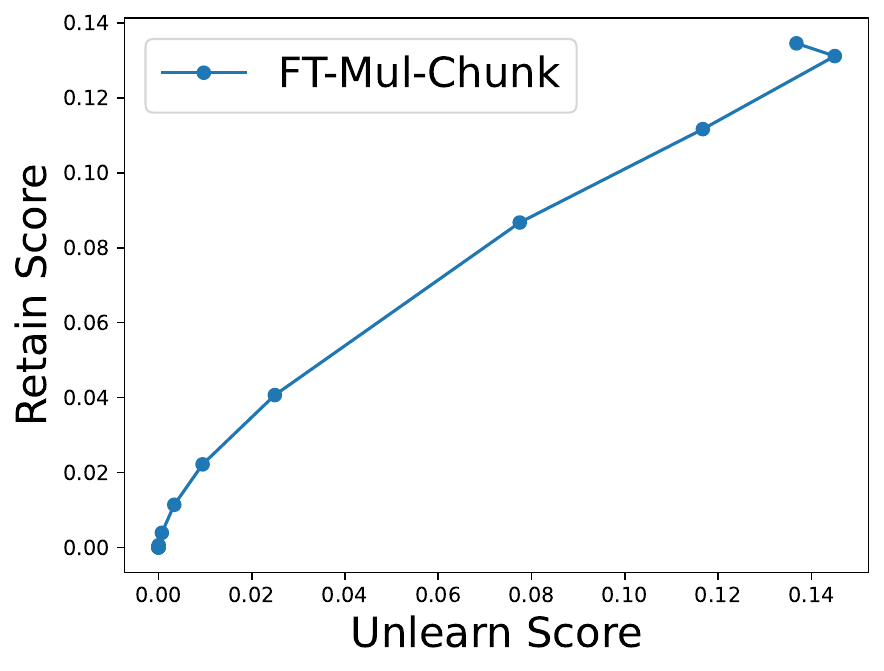}
    \caption{\texttt{UL-Single}, GA}\end{subfigure}
\hfill
\begin{subfigure}[t]{0.32\textwidth}
    \includegraphics[width=\linewidth]{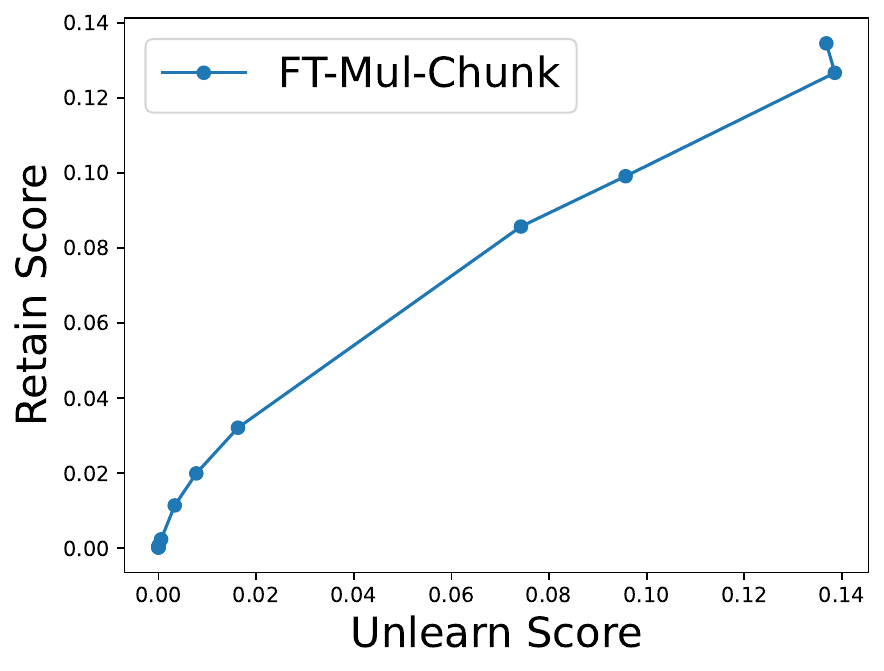}
    \caption{\texttt{UL-Mul}, GA}
\end{subfigure}

\begin{subfigure}[t]{0.32\textwidth}
    \includegraphics[width=\linewidth]{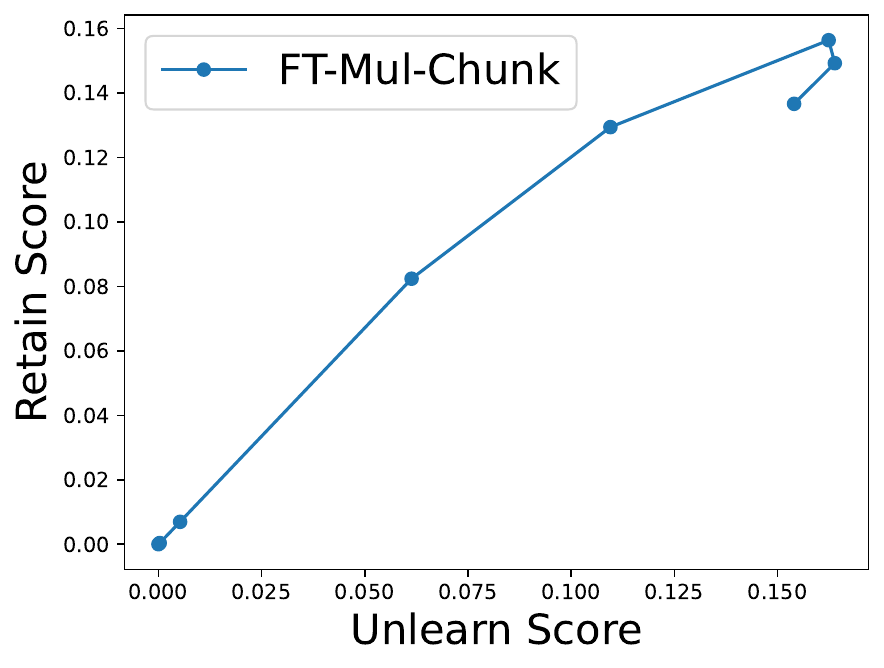}
    \caption{\texttt{UL-Exact}, TV}
\end{subfigure}
\hfill
\begin{subfigure}[t]{0.32\textwidth}
    \includegraphics[width=\linewidth]{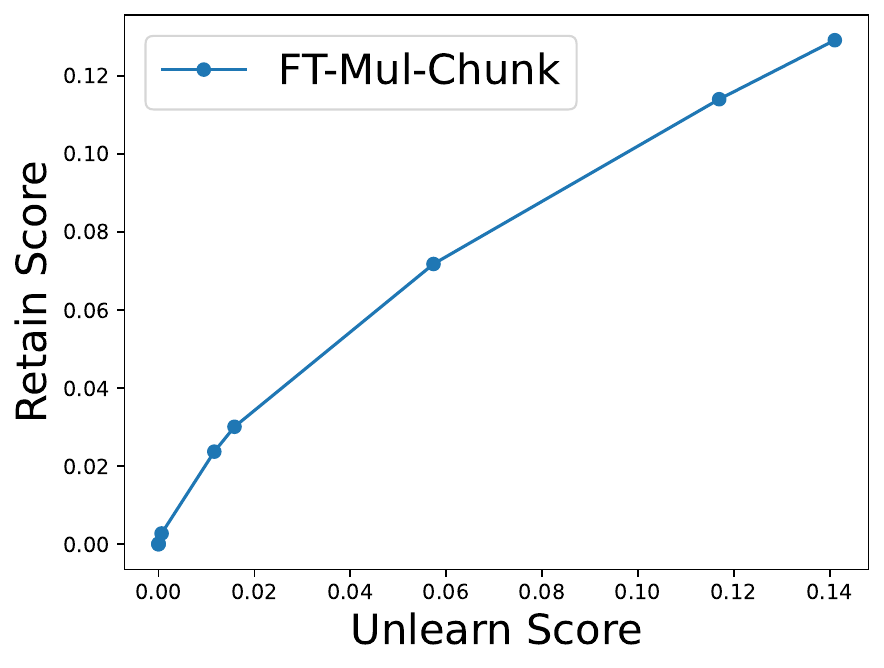}
    \caption{\texttt{UL-Single}, TV}\end{subfigure}
\hfill
\begin{subfigure}[t]{0.32\textwidth}
    \includegraphics[width=\linewidth]{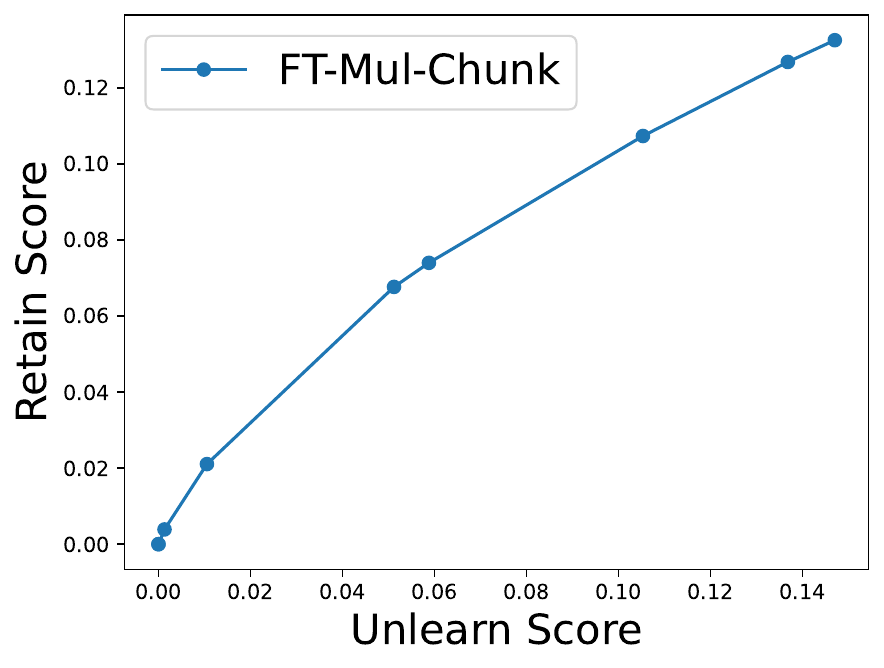}
    \caption{\texttt{UL-Mul}, TV}
\end{subfigure}

\caption{Vanilla \textbf{extraction} trade-off curves for three choices of unlearning data and two unlearning algorithms on \textbf{TOFU+} and \textbf{Llama2-7B}, when the model is fine-tuned from any text chunks.}
\label{fig:extraction_curves_tofu_text_chunk_llama2-7b}
\end{figure}

For the results in Section~\ref{sec:result_chunk}, the extraction trade-off plots for (Llama2-7B, Eval-DU+), and (Llama2-7B, TOFU+) are in Figure~\ref{fig:extraction_curves_eval_du_text_chunk_llama2-7b},~\ref{fig:extraction_curves_eval_du_text_chunk_llama3-8b},~\ref{fig:extraction_curves_eval_du_text_chunk_gemma2-2b},~\ref{fig:extraction_curves_tofu_text_chunk_llama2-7b} respectively.

\section{Implementation Details in Experiments}
\label{sec:app_exp_setup}
\paragraph{Compute resources in the experiment.} All experiments are conducted by NVIDIA RTX 6000 Ada GPU. Each run of the fine-tuning and the unlearning is run on two GPUs. The fine-tuning will take 6-12 hours, and each run of the unlearning process as well as the evaluation will take will take 4-8 hours; the time varies on different models.

\paragraph{Fine-tuning details.} The batch sizes are $16$ for all models fine-tuned on Eval-DU+ and $32$ for the model fine-tuned on TOFU+. In addition, we pick the learning rate $\eta\in \{2\cdot 10^{-5}, 10^{-5}, 2\cdot 10^{-6}\}$ and the number of epochs $N\in \{1, \cdots, 8\}$ to ensure a good fit on the fine-tuning set while having a good test performance. The final selection of the two parameters are presented in Table~\ref{tab:ft_hyper}.

\begin{table}[!h]
\caption{Hyperparameter values of the fine-tuning on different models and datasets: the learning rate $\eta$ and the number of epochs $N$ }
\label{tab:ft_hyper}
\resizebox{\textwidth}{!}{%
	\begin{tabular}{c|cc|cc|cc|cc}
	\toprule
	& \multicolumn{2}{c|}{Llama2-7B, Eval-DU+} & \multicolumn{2}{c|}{Llama3-8B, Eval-DU+} & \multicolumn{2}{c|}{Gemma2-2B, Eval-DU+} & \multicolumn{2}{c}{Llama2-7B, TOFU+}\\
	\midrule
	& $\eta$ & $N$ & $\eta$ & $N$ & $\eta$ & $N$ & $\eta$ & $N$ \\
	\midrule
	\texttt{FT-Single} & $10^{-5}$ & $5$ & $10^{-5}$ & $8$  & $10^{-5}$ & $8$   &  $10^{-5}$ & $5$  \\
	\texttt{FT-Mul} & $10^{-5}$ & $5$  & $10^{-5}$ & $8$ &  $10^{-5}$ & $8$  &  $10^{-5}$ & $5$  \\
	\texttt{FT-Mul-Chunk} & $10^{-5}$ & $4$ & $10^{-5}$ & $8$  &  $10^{-5}$ & $8$  &  $10^{-5}$ & $4$\\
	\bottomrule	
	\end{tabular}
}
\end{table}

\paragraph{Unlearning details.} First of all, \texttt{UL-Mul} has 3 paraphrased descriptions for the same target knowledge. In addition, each unlearning algorithm has its own hyperparameters: gradient ascent (GA) has a list of step numbers $t$ to control the trade-off and the learning rate $\eta_{ga}$ (the batch sizes are fixed as $8$ for Eval-DU+ and $16$ for TOFU+), task vector (TV) has a list of scaling parameter values $\alpha$ to control the trade-off, as well as the number of epoch $T_{tv}$ and the learning rate $\eta_{tv}$ to train the reinforced model (the batch sizes are fixed as $16$ for Eval-DU+ and $32$ for TOFU+). The values are picked to best present the trade-off. Their values given different fine-tuning data choices and unlearning data choices are presented as below:
\begin{table}[!h]
\caption{Hyperparameter values of gradient ascent (GA) when the unlearning data choice is \texttt{UL-Exact}.}
\label{tab:unlearn_hyper}
\resizebox{\textwidth}{!}{%
	\begin{tabular}{c|cc|cc|cc|cc}
	\toprule
	& \multicolumn{2}{c|}{Llama2-7B, Eval-DU+} & \multicolumn{2}{c|}{Llama3-8B, Eval-DU+} & \multicolumn{2}{c|}{Gemma2-2B, Eval-DU+} & \multicolumn{2}{c}{Llama2-7B, TOFU+}\\
	\midrule
	&  List of $t$ & $\eta_{ga}$& List of $t$ & $\eta_{ga}$& List of $t$ & $\eta_{ga}$& List of $t$ & $\eta_{ga}$ \\
	\midrule
	\texttt{FT-Single} & $\{0, 5, 10, \cdots, 75\}$ & $3\times 10^{-6}$ &  $\{0, 5, 10, \cdots, 75\}$ & $3\times 10^{-6}$ & $\{0, 5, 10, \cdots, 75\}$ &  $3\times 10^{-6}$ &  $\{0, 5, 10, \cdots, 75\}$ & $3\times 10^{-6}$  \\
	\texttt{FT-Mul} & $\{0, 5, 10, \cdots, 75\}$ & $3\times 10^{-6}$  &  $\{0, 5, 10, \cdots, 75\}$ & $3\times 10^{-6}$ &  $\{0, 5, 10, \cdots, 75\}$ & $3\times 10^{-6}$  &  $\{0, 5, 10, \cdots, 75\}$ & $3\times 10^{-6}$  \\
	\texttt{FT-Mul-Chunk} & $\{0, 5, 10, \cdots, 75\}$ & $3\times 10^{-6}$ & $\{0, 5, 10, \cdots, 75\}$ & $3\times 10^{-6}$  & $\{0, 5, 10, \cdots, 75\}$  & $3\times 10^{-6}$ &  $\{0, 5, 10, \cdots, 75\}$ & $10^{-5}$\\
	\bottomrule	
	\end{tabular}
}
\vspace{2ex}
\caption{Hyperparameter values of gradient ascent (GA) when the unlearning data choice is \texttt{UL-Single}.}
\resizebox{\textwidth}{!}{%
	\begin{tabular}{c|cc|cc|cc|cc}
	\toprule
	& \multicolumn{2}{c|}{Llama2-7B, Eval-DU+} & \multicolumn{2}{c|}{Llama3-8B, Eval-DU+} & \multicolumn{2}{c|}{Gemma2-2B, Eval-DU+} & \multicolumn{2}{c}{Llama2-7B, TOFU+}\\
	\midrule
	&  List of $t$ & $\eta_{ga}$& List of $t$ & $\eta_{ga}$& List of $t$ & $\eta_{ga}$& List of $t$ & $\eta_{ga}$ \\
	\midrule
	\texttt{FT-Single} & $\{0, 5, 10, \cdots, 75\}$ & $3\times 10^{-6}$ &  $\{0, 5, 10, \cdots, 75\}$ & $3\times 10^{-6}$  & $\{0, 5, 10, \cdots, 75\}$ & $3\times 10^{-6}$  &  $\{0, 5, 10, \cdots, 75\}$ & $3\times 10^{-6}$  \\
	\texttt{FT-Mul} & $\{0, 5, 10, \cdots, 75\}$ & $3\times 10^{-6}$  &  $\{0, 5, 10, \cdots, 75\}$ & $3\times 10^{-6}$   &  $\{0, 5, 10, \cdots, 75\}$ & $3\times 10^{-6}$   &  $\{0, 5, 10, \cdots, 75\}$ & $3\times 10^{-6}$  \\
	\texttt{FT-Mul-Chunk} & $\{0, 5, 10, \cdots, 75\}$ & $3\times 10^{-6}$ & $\{0, 5, 10, \cdots, 75\}$ & $3\times 10^{-6}$   &  $\{0, 5, 10, \cdots, 75\}$ & $3\times 10^{-6}$   &  $\{0, 5, 10, \cdots, 75\}$ & $10^{-6}$\\
	\bottomrule	
	\end{tabular}
}
\vspace{2ex}
\caption{Hyperparameter values of gradient ascent (GA) when the unlearning data choice is \texttt{UL-Mul}.}
\resizebox{\textwidth}{!}{%
	\begin{tabular}{c|cc|cc|cc|cc}
	\toprule
	& \multicolumn{2}{c|}{Llama2-7B, Eval-DU+} & \multicolumn{2}{c|}{Llama3-8B, Eval-DU+} & \multicolumn{2}{c|}{Gemma2-2B, Eval-DU+} & \multicolumn{2}{c}{Llama2-7B, TOFU+}\\
	\midrule
	&  List of $t$ & $\eta_{ga}$& List of $t$ & $\eta_{ga}$& List of $t$ & $\eta_{ga}$& List of $t$ & $\eta_{ga}$ \\
	\midrule
	\texttt{FT-Single} & $\{0, 5, 10, \cdots, 75\}$ & $3\times 10^{-6}$ &  $\{0, 5, 10, \cdots, 75\}$ & $3\times 10^{-6}$   &  $\{0, 5, 10, \cdots, 75\}$ & $3\times 10^{-6}$   &  $\{0, 5, 10, \cdots, 75\}$ & $3\times 10^{-6}$  \\
	\texttt{FT-Mul} & $\{0, 5, 10, \cdots, 75\}$ & $3\times 10^{-6}$  & $\{0, 5, 10, \cdots, 75\}$ & $3\times 10^{-6}$  &  $\{0, 5, 10, \cdots, 75\}$ & $3\times 10^{-6}$ &  $\{0, 5, 10, \cdots, 75\}$ & $3\times 10^{-6}$  \\
	\texttt{FT-Mul-Chunk} & $\{0, 5, 10, \cdots, 75\}$ & $3\times 10^{-6}$ & $\{0, 5, 10, \cdots, 75\}$ & $3\times 10^{-6}$   &  $\{0, 5, 10, \cdots, 75\}$ & $3\times 10^{-6}$  &  $\{0, 5, 10, \cdots, 75\}$ & $10^{-6}$\\
	\bottomrule	
	\end{tabular}
}
\vspace{2ex}
\caption{Hyperparameter values of task vector (TV) when the unlearning data choice is \texttt{UL-Exact}.}
\resizebox{\textwidth}{!}{%
	\begin{tabular}{c|ccc|ccc|ccc|ccc}
	\toprule
	& \multicolumn{3}{c|}{Llama2-7B, Eval-DU+} & \multicolumn{3}{c|}{Llama3-8B, Eval-DU+} & \multicolumn{3}{c|}{Gemma2-2B, Eval-DU+} & \multicolumn{3}{c}{Llama2-7B, TOFU+}\\
	\midrule
	& List of $\alpha$ & $N_{tv}$ & $\eta_{tv}$ & List of $\alpha$ & $N_{tv}$ & $\eta_{tv}$& List of $\alpha$ & $N_{tv}$ & $\eta_{tv}$& List of $\alpha$ & $N_{tv}$ & $\eta_{tv}$ \\
	\midrule
	\texttt{FT-Single} & $\{0, 0.2, 0.5, 1.0, 5.0, 10.0\}$ & $20$ & $10^{-5}$ &  $\{0, 0.2, 0.5, 1.0, 5.0, 10.0\}$ & $20$ & $10^{-5}$   &  $\{0, 0.2, 0.5, 1.0, 5.0, 10.0\}$ & $20$ & $10^{-5}$   &  $\{0, 0.05, 0.1, 0.2, 0.3, 0.4, 0.5, 1.0, 5.0, 10.0\}$ & $20$ & $10^{-5}$    \\
	\texttt{FT-Mul} & $\{0, 0.2, 0.5, 1.0, 5.0, 10.0\}$ & $20$  & $10^{-5}$ & $\{0, 0.2, 0.5, 1.0, 5.0, 10.0\}$ & $20$ & $10^{-5}$  & $\{0, 0.2, 0.5, 1.0, 5.0, 10.0\}$ & $20$ & $10^{-5}$   &  $\{0, 0.05, 0.1, 0.2, 0.3, 0.4, 0.5, 1.0, 5.0, 10.0\}$ & $20$ & $10^{-5}$ \\
	\texttt{FT-Mul-Chunk} & $\{0, 0.2, 0.5, 1.0, 5.0, 10.0 \}$ & $20$ & $10^{-5}$ & $\{0, 0.2, 0.5, 1.0, 5.0, 10.0\}$ & $20$ & $10^{-5}$   &  $\{0, 0.2, 0.5, 1.0, 5.0, 10.0\}$ & $20$ & $10^{-5}$   &  $\{0, 0.2, 0.5, 1.0, 5.0, 10.0, 20.0, 30.0, 50.0\}$ & $400$ & $10^{-5}$\\
	\bottomrule	
	\end{tabular}
}
\vspace{2ex}
\caption{Hyperparameter values of task vector (TV) when the unlearning data choice is \texttt{UL-Single}.}
\resizebox{\textwidth}{!}{%
	\begin{tabular}{c|ccc|ccc|ccc|ccc}
	\toprule
	& \multicolumn{3}{c|}{Llama2-7B, Eval-DU+} & \multicolumn{3}{c|}{Llama3-8B, Eval-DU+} & \multicolumn{3}{c|}{Gemma2-2B, Eval-DU+} & \multicolumn{3}{c}{Llama2-7B, TOFU+}\\
	\midrule
	& List of $\alpha$ & $N_{tv}$ & $\eta_{tv}$ & List of $\alpha$ & $N_{tv}$ & $\eta_{tv}$& List of $\alpha$ & $N_{tv}$ & $\eta_{tv}$& List of $\alpha$ & $N_{tv}$ & $\eta_{tv}$ \\
	\midrule
	\texttt{FT-Single} & $\{0, 0.2, 0.5, 1.0, 5.0, 10.0\}$ & $20$ & $10^{-5}$ &  $\{0, 0.2, 0.5, 1.0, 5.0, 10.0\}$ & $20$ & $10^{-5}$   &  $\{0, 0.2, 0.5, 1.0, 5.0, 10.0\}$ & $20$ & $10^{-5}$   &  $\{0, 0.05, 0.1, 0.2, 0.3, 0.4, 0.5, 1.0, 5.0, 10.0\}$ & $20$ & $10^{-5}$    \\
	\texttt{FT-Mul} & $\{0, 0.2, 0.5, 1.0, 5.0, 10.0\}$ & $20$  & $10^{-5}$ & $\{0, 0.2, 0.5, 1.0, 5.0, 10.0\}$ & $20$ & $10^{-5}$  & $\{0, 0.2, 0.5, 1.0, 5.0, 10.0\}$ & $20$ & $10^{-5}$   &  $\{0, 0.05, 0.1, 0.2, 0.3, 0.4, 0.5, 1.0, 5.0, 10.0\}$ & $20$ & $10^{-5}$ \\
	\texttt{FT-Mul-Chunk} & $\{0, 0.2, 0.5, 1.0, 5.0, 10.0 \}$ & $20$ & $10^{-5}$ & $\{0, 0.2, 0.5, 1.0, 5.0, 10.0\}$ & $20$ & $10^{-5}$   &  $\{0, 0.2, 0.5, 1.0, 5.0, 10.0\}$ & $20$ & $10^{-5}$   &  $\{0, 0.2, 0.5, 1.0, 5.0, 10.0, 20.0, 30.0, 50.0\}$ & $400$ & $10^{-5}$\\
	\bottomrule	
	\end{tabular}
}
\vspace{2ex}
\caption{Hyperparameter values of task vector (TV) when the unlearning data choice is \texttt{UL-Mul}.}
\resizebox{\textwidth}{!}{%
	\begin{tabular}{c|ccc|ccc|ccc|ccc}
	\toprule
	& \multicolumn{3}{c|}{Llama2-7B, Eval-DU+} & \multicolumn{3}{c|}{Llama3-8B, Eval-DU+} & \multicolumn{3}{c|}{Gemma2-2B, Eval-DU+} & \multicolumn{3}{c}{Llama2-7B, TOFU+}\\
	\midrule
	& List of $\alpha$ & $N_{tv}$ & $\eta_{tv}$ & List of $\alpha$ & $N_{tv}$ & $\eta_{tv}$& List of $\alpha$ & $N_{tv}$ & $\eta_{tv}$& List of $\alpha$ & $N_{tv}$ & $\eta_{tv}$ \\
	\midrule
	\texttt{FT-Single} & $\{0, 0.2, 0.5, 1.0, 5.0, 10.0\}$ & $20$ & $10^{-5}$ &  $\{0, 0.2, 0.5, 1.0, 5.0, 10.0\}$ & $20$ & $10^{-5}$   &  $\{0, 0.2, 0.5, 1.0, 5.0, 10.0\}$ & $20$ & $10^{-5}$   &  $\{0, 0.05, 0.1, 0.2, 0.3, 0.4, 0.5, 1.0, 5.0, 10.0\}$ & $20$ & $10^{-5}$    \\
	\texttt{FT-Mul} & $\{0, 0.2, 0.5, 1.0, 5.0, 10.0\}$ & $20$  & $10^{-5}$ & $\{0, 0.2, 0.5, 1.0, 5.0, 10.0\}$ & $20$ & $10^{-5}$  & $\{0, 0.2, 0.5, 1.0, 5.0, 10.0\}$ & $20$ & $10^{-5}$   &  $\{0, 0.05, 0.1, 0.2, 0.3, 0.4, 0.5, 1.0, 5.0, 10.0\}$ & $20$ & $10^{-5}$ \\
	\texttt{FT-Mul-Chunk} & $\{0, 0.2, 0.5, 1.0, 5.0, 10.0 \}$ & $20$ & $10^{-5}$ & $\{0, 0.2, 0.5, 1.0, 5.0, 10.0\}$ & $20$ & $10^{-5}$   &  $\{0, 0.2, 0.5, 1.0, 5.0, 10.0\}$ & $20$ & $10^{-5}$   &  $\{0, 0.2, 0.5, 1.0, 5.0, 10.0, 20.0, 30.0, 50.0\}$ & $400$ & $10^{-5}$\\
	\bottomrule	
	\end{tabular}
}
\end{table}

